\documentclass{article}

\usepackage[final,nonatbib]{neurips_2022}

\usepackage{ragged2e}
\usepackage{microtype}
\usepackage{graphicx}
\usepackage{subfigure}
\usepackage{booktabs} % for professional tables
\usepackage{amsmath,amssymb,amsfonts}
\usepackage{amsthm}
\usepackage{mathrsfs}
\usepackage{algorithmic}
\usepackage{float}
\usepackage{graphicx}
\usepackage{textcomp}
\usepackage{xcolor}
\usepackage{colortbl,booktabs}
\usepackage[OT1]{fontenc} 
\definecolor{mydarkblue}{rgb}{0,0.08,0.45}
\usepackage[colorlinks, citecolor=mydarkblue]{hyperref}
\usepackage{times}  %Required
\usepackage{helvet}  %Required
\usepackage{courier}  %Required
\usepackage{bm}
\usepackage{bbm}
\usepackage[square,numbers]{natbib}
\newtheorem{Theorem}{\textbf{Problem}}

\newtheorem{assumption}{\textbf{Assumption}}
\newtheorem{theorem*}{Theorem}

\newtheorem{lemma}{Lemma}
\newtheorem{Definition}{Definition}
\newtheorem{Proposition}{Proposition}

\newtheorem{Condition}{\textbf{Condition}}
\DeclareMathOperator*{\argmin}{arg\,min}
\DeclareMathOperator*{\argmax}{arg\,max}
\usepackage{subfigure}
\usepackage{graphbox}
\usepackage{thm-restate}
\usepackage{setspace}
\usepackage{titletoc}
\makeatletter
  \def\ttl@Hy@steplink#1{%
    \Hy@MakeCurrentHrefAuto{#1*}%
    \edef\ttl@Hy@saveanchor{%
      \noexpand\Hy@raisedlink{%
        \noexpand\hyper@anchorstart{\@currentHref}%
        \noexpand\hyper@anchorend
        \def\noexpand\ttl@Hy@SavedCurrentHref{\@currentHref}%
        \noexpand\ttl@Hy@PatchSaveWrite
      }%
    }%
  }%
  \def\ttl@Hy@PatchSaveWrite{%
    \begingroup
      \toks@\expandafter{\ttl@savewrite}%
      \edef\x{\endgroup
        \def\noexpand\ttl@savewrite{%
          \let\noexpand\@currentHref
              \noexpand\ttl@Hy@SavedCurrentHref
          \the\toks@
        }%
      }%
    \x
  }%
  \def\ttl@Hy@refstepcounter#1{%
    \let\ttl@b\Hy@raisedlink
    \def\Hy@raisedlink##1{%
      \def\ttl@Hy@saveanchor{\Hy@raisedlink{##1}}%
    }%
    \refstepcounter{#1}%
    \let\Hy@raisedlink\ttl@b
  }%
\def\ttl@gobblecontents#1#2#3#4{\ignorespaces}%
\makeatother
\newcommand\DoToC{%
  \startcontents
  \printcontents{}{1}{
  \textbf{\Large Table of Contents of Appendix}
  \vskip2pt\hrule\vskip2pt
  \protect\setlength{\parskip}{0pt}\protect\onehalfspacing
  }
  \vskip2pt\hrule\vskip2pt
}

\title{Is Out-of-Distribution Detection Learnable?}

\author{
 Zhen~Fang\textsuperscript{1},~
 Yixuan~Li\textsuperscript{2},~
  Jie Lu\textsuperscript{1}\thanks{Corresponding author},~
 Jiahua~Dong\textsuperscript{3,4},~
 Bo~Han\textsuperscript{5},~
 Feng~Liu\textsuperscript{1,6$*$}\\ 
 \textsuperscript{1}Australian Artificial Intelligence Institute, University of Technology Sydney.\\
  \textsuperscript{2}Department of Computer Sciences, University of Wisconsin-Madison.\\
 \textsuperscript{3}State Key Laboratory of Robotics,  Shenyang Institute of Automation, \\ Chinese Academy of Sciences. 
 \textsuperscript{4}ETH Zurich, Switzerland. \\
 \textsuperscript{5}Department of Computer Science, Hong Kong Baptist University. \\
  \textsuperscript{6}School of Mathematics and Statistics, University of Melbourne. \\
 {\tt\small \{zhen.fang,jie.lu\}@uts.edu.au, sharonli@cs.wisc.edu,}\\  {\tt\small dongjiahua1995@gmail.com,bhanml@comp.hkbu.edu.hk,feng.liu1@unimelb.edu.au}
}

\begin{document}

\maketitle

\begin{abstract}
Supervised learning aims to train a classifier under the assumption that training and test data are from the same distribution. 
To ease the above assumption, researchers have studied a more realistic setting: \emph{out-of-distribution} (OOD) detection, where test data may come from classes that are unknown during training (\textit{i.e.}, OOD data).
Due to the unavailability and diversity of OOD data,
good generalization ability is crucial for effective OOD detection algorithms. 
To study the generalization of OOD detection, in this paper, we investigate the \emph{probably approximately correct} (PAC) learning theory of OOD detection, which is proposed by researchers as an \emph{open problem}. First, we find a necessary condition for the learnability of OOD detection. Then, using this condition, we prove several impossibility theorems for the learnability of OOD detection under some scenarios. Although the impossibility theorems are frustrating, we find that some conditions of these impossibility theorems may not hold in some practical scenarios.
Based on this observation, 
we next give several necessary and sufficient conditions to characterize the learnability of OOD detection in some practical scenarios. 
Lastly, we also offer theoretical supports for several representative OOD detection works based on our OOD theory.
\end{abstract}

\section{Introduction}
\label{sec:intro}

The success of supervised learning is established on an implicit assumption
that training and test data share a same distribution, \textit{i.e.}, \textit{in-distribution} (ID)  \cite{Dosovitskiy2021animage,Huang2017densely,DBLP:conf/cvpr/HsuSJK20,Yang2021generalized}. However, test data distribution in many real-world scenarios may violate the assumption and, instead, contain \emph{out-of-distribution} (OOD) data whose labels have not been seen during the training process \cite{openmax16cvpr,chen2021robustifying}. To mitigate the risk of OOD data, researchers have 
considered a more practical learning scenario: OOD detection which determines whether an input is ID/OOD, while classifying the ID data into respective classes. 
OOD detection has shown great potential to ensure the reliable deployment of machine learning models in the real world. 
A rich line of algorithms have been developed to {empirically} address the OOD detection problem \cite{chen2021robustifying, Hendrycks2017abaseline,liang2018enhancing,lee2018asimple,zong2018deep, Pidhorskyi2018generative,Nalisnick2019do,Hendrycks2019deep, ren2019likelihood, lin2021mood, salehi2021unified,sun2021react,Huang2021on, Fort2021exploring, ming2022spurious}. However, very few works study theory of OOD detection,  which hinders the rigorous path forward for the field. This paper aims to bridge the gap.

In this paper, we provide a theoretical framework to  understand the learnability of the OOD detection problem. We investigate the probably approximately correct (PAC) learning theory of OOD detection, which is posed as an open problem to date. Unlike the classical PAC learning theory in a supervised setting, our problem setting is fundamentally challenging due to the \emph{absence of OOD data} in training. In many real-world scenarios, OOD data can be diverse and priori-unknown. Given this, we study whether there exists an algorithm that can be used to detect various OOD data instead of merely some specified OOD data. Such is the significance of studying the learning theory for OOD detection \citep{Yang2021generalized}. This motivates our question: \textit{is OOD detection PAC learnable? {i.e.}, is there the PAC learning  theory to guarantee the generalization ability of OOD detection?} 
% however, very few works discuss OOD detection in theory. Researchers \citep{Yang2021generalized} appeal to focus on developing theory to provide in-depth insights that guide algorithmic development with rigorous guarantees.
%Normally, OOD data are unknown during the training process and can be various in the real world. 

% Although \citep{Fang2021learning} has studied the \textit{almost agnostic PAC learnability} for OOD detection,
% % \citep{Fang2021learning} proves a almost convergent generalization bound, 
% the assumption made in \citep{Fang2021learning} is very strong and unpractical. \citep{Fang2021learning} realizes this drawback and raises an open question:
To investigate the learning theory, we mainly focus on two basic spaces: domain space and hypothesis space. The domain space is a space consisting of some distributions, and the hypothesis space is a space consisting of some classifiers. Existing agnostic PAC theories in supervised learning \cite{shalev2014understanding,mohri2018foundations} are distribution-free, \textit{i.e.}, the domain space consists of all domains. Yet, in Theorem \ref{T4}, we shows that the learning theory of OOD detection is not distribution-free. In fact, we discover that OOD detection is learnable only if the domain space and the hypothesis space satisfy some special conditions, \textit{e.g.,} Conditions \ref{C1} and \ref{Con2}. Notably, 
% it is \textit{very difficult} to explore the essential conditions to ensure the learnability of OOD detection from the scratch. 
there are many conditions and theorems in existing learning theories and many OOD detection algorithms in the literature. Thus, it is very difficult to analyze the relation between these theories and algorithms, and explore useful conditions to ensure the learnability of OOD detection, especially when we have to explore them \textit{from the scratch}. Thus, the main aim of our paper is to study these essential conditions. From these
essential conditions, we can know \textit{when} OOD detection can be successful in practical scenarios.
We restate our question and goal in following:

$~~~~~~~~~~~~~~~~~$\fbox{
\parbox{0.745\textwidth}{
\textit{Given  hypothesis spaces and several representative domain spaces, what are the {conditions} to ensure the learnability of OOD detection? If possible, we hope that these conditions are necessary and sufficient in some scenarios.}
}
}

\textbf{Main Results.} We investigate the learnability of OOD detection starting from the largest space---the total space, and give a {necessary condition} (Condition \ref{C1}) for the learnability. However, we find that the overlap between ID and OOD data may result in that the necessary condition does not hold.
Therefore, we give an impossibility theorem to demonstrate that OOD detection fails in the total space (Theorem~\ref{T4}). Next, we study OOD detection in the separate space, where there are no overlaps between the ID and OOD data. Unfortunately, there still exists impossibility theorem (Theorem~\ref{T12}), which demonstrates that OOD detection is not learnable in the separate space under some conditions. 

Although the impossibility theorems obtained in the separate space are frustrating, we find that some conditions of these impossibility theorems may not hold in some practical scenarios.
Based on this observation, we give several {necessary and sufficient} conditions to characterize the learnability of OOD detection in the separate space (Theorems \ref{T13} and \ref{T24}). Especially, when our model is based on \emph{fully-connected neural network} (FCNN),  OOD detection is learnable in the separate space if and only if the feature space is finite.
Furthermore, we investigate the learnability of OOD detection in other more practical domain spaces, \textit{e.g.}, the finite-ID-distribution space (Theorem \ref{T-SET}) and the density-based space (Theorem \ref{T-SET2}). By studying the finite-ID-distribution space, we discover a compatibility condition (Condition \ref{Con2}) that is a necessary and sufficient condition for this space. Next, we further investigate the compatibility condition in the density-based space, and find that such condition is also the necessary and sufficient condition in some practical scenarios (Theorem \ref{T24.3}).

% \vspace{-0.2cm}
\textbf{Implications and Impacts of Theory.} Our study is not of purely theoretical interest; it has also practical impacts.
First, when we design OOD detection algorithms, we normally only have finite ID datasets, 
% \textit{i.e.}, 
corresponding to
the finite-ID-distribution space. In this case, Theorem \ref{T-SET} gives the necessary and sufficient condition to the success of OOD detection.
Second, our theory provides theoretical support (Theorems \ref{T24} and \ref{T24.3}) for several representative OOD detection works \cite{Hendrycks2017abaseline,liang2018enhancing,liu2020energy}. 
Third, our theory shows that OOD detection is learnable in image-based scenarios when ID images have clearly different semantic labels and styles  (\textit{far-OOD}) from OOD images. 
Fourth, we should not expect a universally working algorithm. 
It is necessary to design different algorithms in different scenarios.
\vspace{-0.5em}
\section{Learning Setups}\label{S3}
\vspace{-0.5em}
We start by introducing the necessary concepts and notations for our theoretical framework.
%Here, we introduce OOD detection and necessary concepts.
Given a feature space $\mathcal{X}\subset \mathbb{R}^d$ and a label space $\mathcal{Y}:=\{1,...,K\}$, we have an ID joint distribution $D_{X_{\rm I}Y_{\rm I}}$ over $\mathcal{X}\times \mathcal{Y}$, where $X_{\rm I}\in \mathcal{X}$ and $Y_{\rm I}\in \mathcal{Y}$ are random variables. We also have an OOD joint distribution $D_{X_{\rm O}Y_{\rm O}}$, where $X_{\rm O}$ is a random variable from $\mathcal{X}$, but $Y_{\rm O}$ is a random variable whose outputs do not belong to $\mathcal{Y}$. During testing, we will meet a mixture of ID and OOD joint distributions: $D_{XY}:=(1-\pi^{\rm out})D_{X_{\rm I}Y_{\rm I}}+\pi^{\rm out}D_{X_{\rm O}Y_{\rm O}}$, and can only observe the marginal distribution $D_X:=(1-\pi^{\rm out})D_{X_{\rm I}}+\pi^{\rm out}D_{X_{\rm O}}$, where the constant $\pi^{\rm out}\in [0,1)$ is an unknown class-prior probability.

% there is an unknown class-prior probability $\pi^{\rm out}\in [0,1)$ to mix marginal distributions $D_{X_{\rm I}}$ and $D_{X_{\rm O}}$. We set $D:=(1-\pi^{\rm out})D_{X_{\rm I}}+\pi^{\rm out}D_{X_{\rm O}}$. For  simplicity, we also call the distribution $D_{XY}:=(1-\pi^{\rm out})D_{X_{\rm I}Y_{\rm I}}+\pi^{\rm out}D_{X_{\rm O}Y_{\rm O}}$ \textit{domain}.

%$\mathcal{Y}$ and $\mathcal{Y}^{\rm out}$ are the ID and OOD label spaces, respectively. $D_{XY|Y\in \mathcal{Y}}$ is the ID distribution, and $D_{X_{\rm O}Y_{\rm O}}$ is the OOD distribution. $D_{Y\in }>0$ ensures that the ID distribution can be observed. 

\begin{Theorem}[OOD Detection \citep{Yang2021generalized}]\label{P1}
Given an ID joint distribution $D_{X_{\rm I}Y_{\rm I}}$ and a training data $S:=\{(\mathbf{x}^1,{y}^1),...,(\mathbf{x}^n,{y}^n)\}$ drawn {independent and identically distributed}  from $D_{X_{\rm I}Y_{\rm I}}$,
the aim of OOD detection is to train a classifier $f$ by using the training data $S$ such that, for any test data $\mathbf{x}$ drawn from the mixed marginal distribution $D_X$:
1) if $\mathbf{x}$ is an observation from $D_{X_{\rm I}}$, $f$ can classify $
\mathbf{x}$ into correct ID classes; and
2) if $\mathbf{x}$ is an observation from $D_{X_{\rm O}}$, $f$ can detect $
\mathbf{x}$ as OOD data.
\end{Theorem}

According to the survey \citep{Yang2021generalized}, when $K>1$, OOD detection is also known as the open-set recognition or open-set learning \cite{Fang2021learning,Chen2021Adversarial}; and when $K=1$, OOD detection reduces to one-class novelty detection and  semantic anomaly detection \cite{DBLP:conf/icml/RuffGDSVBMK18,DBLP:conf/icml/GoyalRJS020,DBLP:conf/pkdd/DeeckeVRMK18}.

\textbf{OOD Label and Domain Space.} Based on Problem \ref{P1}, we know it is not necessary to classify OOD data
into the correct OOD classes. Without loss of generality,
let all OOD data be allocated to one big OOD class, \textit{i.e.}, $Y_{\rm O}=K+1$ \cite{Fang2021learning,Fang2020Open}. 
 To investigate the PAC learnability of OOD detection, we define a domain space $\mathscr{D}_{XY}$, which is a set consisting of some joint distributions $D_{XY}$ mixed by some ID joint distributions and some OOD joint distributions.
In this paper, the joint distribution $D_{XY}$ mixed by ID joint distribution $D_{X_{\rm I}Y_{\rm I}}$ and OOD joint distribution $D_{X_{\rm O}Y_{\rm O}}$ is called \textbf{\textit{domain}}.

\textbf{Hypothesis Spaces and Scoring Function Spaces.} A hypothesis space $\mathcal{H}$ is a subset of function space, \textit{i.e.},
$
  \mathcal{H}\subset  \{ { h}: \mathcal{X}\rightarrow \mathcal{Y}\cup \{K+1\}\}.
$ We  set $\mathcal{H}^{\rm in}\subset  \{ { h}: \mathcal{X}\rightarrow \mathcal{Y}\}$ to the ID hypothesis space. We also define $\mathcal{H}^{\rm b}\subset  \{ { h}: \mathcal{X}\rightarrow \{1,2\}\}$ as the hypothesis space for binary classification, where $1$ represents the ID data, and $2$ represents the OOD data. The function $h$ is called the hypothesis function. A scoring function space is a subset of function space, \textit{i.e.}, $\mathcal{F}_{l}\subset  \{ \mathbf{f}: \mathcal{X}\rightarrow \mathbb{R}^{l}\}$, where $l$ is the output's dimension of the vector-valued function $\mathbf{f}$. The function $\mathbf{f}$ is called the scoring function.

\textbf{Loss and Risks.} Let $\mathcal{Y}_{\rm all}=\mathcal{Y}\cup\{K+1\}$. Given a loss function $\ell\footnote{Note that $\mathcal{Y}_{\rm all}\times \mathcal{Y}_{\rm all}$ is a finite set, therefore, $\ell$ is bounded.}:\mathcal{Y}_{\rm all}\times \mathcal{Y}_{\rm all}\rightarrow \mathbb{R}_{\geq 0}$ satisfying that $\ell(y_1,y_2)=0$ if and only if $y_1=y_2$, and any ${ h}\in \mathcal{H}$, then the \textit{risk} with respect to ${D}_{XY}$ is
\begin{equation}\label{risks}
\begin{split}
   & R_{D}({h}):= \mathbb{E}_{(\mathbf{x},y)\sim D_{XY}} \ell({ h}(\mathbf{x}),{y}).
\end{split}
 \end{equation} 
The $\alpha$-risk $R_{D}^{\alpha}({h}):= (1-\alpha)R^{\rm in}_D({h})+\alpha R^{\rm out}_D({h}), \forall \alpha\in [0,1]$, where the risks $R^{\rm in}_D({h})$, $R^{\rm out}_D({h})$ are
% {The \textit{partial risks} $R^{\rm in}_D({h})$, $R^{\rm out}_D({h})$ with respect to ID, OOD distributions}, and the \textit{$\alpha$-risk} $R_{D}^{\alpha}({h})$ are
\begin{equation*}
\begin{split}
  & R^{\rm in}_D({h}):= \mathbb{E}_{(\mathbf{x},y)\sim D_{X_{\rm I}Y_{\rm I}}} \ell({ h}(\mathbf{x}),{y}),
~~~~~~R^{\rm out}_D({h}):= \mathbb{E}_{\mathbf{x}\sim D_{X_{\rm O}}} \ell({ h}(\mathbf{x}),K+1).
\end{split}
\end{equation*}
% Then, the \textit{$\alpha$-risk} is 
%\begin{equation}\label{alpha-risks}
%\begin{split}
% R_{D}^{\alpha}({h}):= (1-\alpha)R^{\rm in}_D({h})+\alpha R^{\rm out}_D({h}), ~\forall \alpha\in [0,1].
%\end{split}
%\end{equation}
%Clearly, when $\alpha=\pi^{\rm out}$, then
%$
%  R_{D}({h})=  R_{D}^{\alpha}({h}).
%$
%Here $\pi^{\rm out}$ is called the class-prior probability and is unknown during the training process, since we cannot observe the OOD distribution. 

\textbf{Learnability.} We aim to select a hypothesis function $h \in \mathcal{H}$ with approximately minimal risk,
based on finite data. Generally, we expect the approximation to get better, with the increase in sample size. Algorithms achieving this are said to be consistent. Formally, we introduce the following definition: 
\begin{Definition}[Learnability of OOD Detection]\label{D0}
Given a domain space $\mathscr{D}_{XY}$ and a hypothesis space $\mathcal{H}\subset  \{ { h}: \mathcal{X}\rightarrow \mathcal{Y}_{\rm all}\}$, we say  OOD detection is \textbf{learnable} in  $\mathscr{D}_{XY}$ for $\mathcal{H}$, if there exists an algorithm $\mathbf{A}\footnote{Similar to \cite{shalev2010learnability}, in this paper, we regard an algorithm as a mapping from $\cup_{n=1}^{+\infty}(\mathcal{X}\times\mathcal{Y})^n$ to $\mathcal{H}$.}: \cup_{n=1}^{+\infty}(\mathcal{X}\times\mathcal{Y})^n\rightarrow \mathcal{H}$ and a monotonically decreasing
sequence $\epsilon_{\rm cons}(n)$, such that $\epsilon_{\rm cons}(n)\rightarrow 0$, as $n\rightarrow +\infty$, and for any domain $D_{XY}\in \mathscr{D}_{XY}$,
\begin{equation}\label{issue-definition1}
    \mathbb{E}_{S\sim D^n_{X_{\rm I}Y_{\rm I}}}\big[ R_D(\mathbf{A}(S))- \inf_{h\in \mathcal{H}}R_D(h)\big] \leq \epsilon_{\rm cons}(n),
\end{equation}
An algorithm $\mathbf{A}$ for which this holds is said to be consistent with respect to $\mathscr{D}_{XY}$.
\end{Definition}
 Definition \ref{D0} is a natural extension of agnostic PAC learnability of supervised learning \cite{shalev2010learnability}. If for any $D_{XY}\in \mathscr{D}_{XY}$, $\pi^{\rm out}=0$, then Definition \ref{D2} is the agnostic PAC learnability of supervised learning. Although the expression of Definition \ref{D0} is different from the normal definition of agnostic PAC learning in \cite{shalev2014understanding}, one can easily prove that they are equivalent when $\ell$ is bounded, see Appendix \ref{SB.3}.  

Since OOD data are unavailable, it is impossible to obtain information about the class-prior probability $\pi^{\rm out}$. Furthermore, in the real world, it is possible that $\pi^{\rm out}$
can be any value in $[0,1)$. Therefore, the imbalance issue between ID and OOD distributions, and the priori-unknown issue (\textit{i.e.}, $\pi^{\rm out}$ is unknown) are the core challenges. To ease these challenges, researchers use AUROC, AUPR and FPR95 to estimate the performance of OOD detection \cite{Huang2021on,Chen2020Learning,chen2020informative,chen2020robust,Bao2021Evidental,DBLP:journals/corr/abs-2203-05114}. It seems that there is a gap between Definition \ref{D0} and existing works. To eliminate this gap, we revise Eq. \eqref{issue-definition1} as follows:
\begin{equation}\label{issue-definition2}
\begin{split}
   & \mathbb{E}_{S\sim D^n_{X_{\rm I}Y_{\rm I}}}\big[ R_D^{\alpha}(\mathbf{A}(S))- \inf_{h\in \mathcal{H}}R_D^{\alpha}(h)\big] \leq \epsilon_{\rm cons}(n),~\forall \alpha\in[0,1].
    \end{split}
\end{equation}
If an algorithm $\mathbf{A}$ satisfies Eq. \eqref{issue-definition2}, then the imbalance issue and the prior-unknown issue disappear. That is, $\mathbf{A}$ can simultaneously classify the ID data and detect the OOD data well. Based on the above discussion, we define the strong learnability of OOD detection as follows:
\begin{Definition}[Strong Learnability of OOD Detection]\label{D2}
Given a domain space $\mathscr{D}_{XY}$ and a hypothesis space $\mathcal{H}\subset  \{ { h}: \mathcal{X}\rightarrow \mathcal{Y}_{\rm all}\}$, we say  OOD detection is \textbf{strongly learnable} in  $\mathscr{D}_{XY}$ for $\mathcal{H}$, if there exists an algorithm $\mathbf{A}: \cup_{n=1}^{+\infty}(\mathcal{X}\times\mathcal{Y})^n\rightarrow \mathcal{H}$ and a monotonically decreasing
sequence $\epsilon_{\rm cons}(n)$, such that $\epsilon_{\rm cons}(n)\rightarrow 0$, as $n\rightarrow +\infty$, and for any domain $D_{XY}\in \mathscr{D}_{XY}$,
\begin{equation*}
    \mathbb{E}_{S\sim D^n_{X_{\rm I}Y_{\rm I}}}\big[ R_D^{\alpha}(\mathbf{A}(S))- \inf_{h\in \mathcal{H}}R_D^{\alpha}(h)\big] \leq \epsilon_{\rm cons}(n), ~\forall \alpha \in[0,1].
\end{equation*}
%An algorithm $\mathbf{A}$ for which this holds is said to be strongly consistent with respect to $\mathscr{D}_{XY}$.
\end{Definition}

In Theorem \ref{T1}, we have shown that the strong learnability of OOD detection is equivalent to the learnability of OOD detection, if the domain space $\mathscr{D}_{XY}$ is a \textit{prior-unknown space} (see Definition \ref{D3}). In this paper, we mainly discuss the learnability in the prior-unknown space. Therefore, \textit{when we mention that OOD detection is learnable, we also mean that OOD detection is strongly learnable}.

\textbf{Goal of Theory.}
Note that the agnostic PAC learnability of supervised learning is distribution-free, \textit{i.e.}, the domain space $\mathscr{D}_{XY}$ consists of all domains. However, due to the absence of OOD data during the training process \citep{liang2018enhancing,ren2019likelihood,Fang2021learning}, it is obvious that the learnability of OOD detection is not distribution-free (\textit{i.e.}, Theorem \ref{T4}). In fact, we discover that the learnability of OOD detection is deeply correlated with the relationship between the domain space $\mathscr{D}_{XY}$ and the hypothesis space $\mathcal{H}$. That is, OOD detection is learnable only when the domain space $\mathscr{D}_{XY}$ and the hypothesis space $\mathcal{H}$ satisfy some special conditions, \textit{e.g.,} Condition \ref{C1} and Condition \ref{Con2}. We present our goal as follows:

$~~~~~~~~~~$\fbox{
\parbox{0.85\textwidth}{
\textit{\textbf{Goal:} given a hypothesis space $\mathcal{H}$ and several representative domain spaces $\mathscr{D}_{XY}$, what are the \textbf{conditions} to ensure the learnability of OOD detection? Furthermore, if possible, we hope that these conditions are \textbf{necessary and sufficient} in some scenarios.}
}
}

Therefore, compared to the agnostic PAC learnability of supervised learning, our theory doesn't focus on the distribution-free case, but focuses on discovering essential conditions to guarantee the learnability of OOD detection in several representative and practical domain spaces $\mathscr{D}_{XY}$. By these essential conditions, we can know \textit{when} OOD detection can be successful in real applications.
\vspace{-0.1em}
\section{Learning in Priori-unknown Spaces}
\vspace{-0.1em}
% Here, we introduce five representative domain spaces. 

We first investigate a special space, called {prior-unknown} space. In such space, Definition \ref{D0} and Definition \ref{D2} are equivalent. Furthermore, we also prove that if OOD detection is strongly learnable in a space $\mathscr{D}_{XY}$, then one can discover a larger domain space, which is prior-unknown, to ensure the learnability of OOD detection. These results imply that it is enough to consider our theory in the prior-unknown spaces. The prior-unknown space is introduced as follows:

\begin{Definition}\label{D3}
Given a domain space $\mathscr{D}_{XY}$, we say $\mathscr{D}_{XY}$ is a priori-unknown space, if for any domain $D_{XY}\in \mathscr{D}_{XY}$ and any $\alpha \in [0,1)$, we have $D_{XY}^{\alpha}:=(1-\alpha)D_{X_{\rm I}Y_{\rm I}}+\alpha D_{X_{\rm O}Y_{\rm O}}\in \mathscr{D}_{XY}$.
\end{Definition}

%Theorem \ref{T1} then reveals how the priori-unknown space affects the learnability of OOD detection.

\begin{restatable}{theorem}{thmCone}
\label{T1}
Given domain spaces $\mathscr{D}_{XY}$ and $\mathscr{D}_{XY}'=\{D_{XY}^{\alpha}:\forall D_{XY}\in \mathscr{D}_{XY}, \forall \alpha\in [0,1)\}$, then
\\
1) $\mathscr{D}_{XY}'$ is a priori-unknown space and $\mathscr{D}_{XY}\subset \mathscr{D}_{XY}'$;\\
2) if $\mathscr{D}_{XY}$ is a priori-unknown space, then Definition \ref{D0} and Definition \ref{D2} are \textbf{equivalent};\\
3) OOD detection is strongly learnable in $\mathscr{D}_{XY}$ \textbf{if and only if} OOD detection is learnable in $\mathscr{D}_{XY}'$.
%\begin{equation*}
%\begin{split}
% \mathbb{E}_{S\sim D^n_{X_{\rm I}Y_{\rm I}}}[ R^{\rm out}_D(\mathbf{A}(S))-& \inf_{h\in \mathcal{H}}R^{\rm out}_D(h)]\leq \epsilon(n),
% \\
 %\mathbb{E}_{S\sim D^n_{X_{\rm I}Y_{\rm I}}}[ R^{\rm in}_D(\mathbf{A}(S))-& \inf_{h\in \mathcal{H}}R^{\rm in}_D(h)] \leq \epsilon(n).
% \end{split}
%\end{equation*}
%The above inequalities imply that for any $\alpha \in [0,1]$,
\end{restatable}
% \end{theorem}
\vspace{-0.1cm}
The second result of Theorem \ref{T1} bridges the learnability and strong learnability, which implies that if an algorithm $\mathbf{A}$ is consistent with respect to a prior-unknown space, then this algorithm $\mathbf{A}$ can address the imbalance issue between ID and OOD distributions, and the priori-unknown issue well. Based on Theorem \ref{T1}, we focus on our theory in the prior-unknown spaces. Furthermore, to demystify the learnability of OOD detection, we introduce five representative priori-unknown spaces:

\noindent$\bullet$ Single-distribution space $\mathscr{D}_{XY}^{D_{XY}}$.  For a domain $D_{XY}$, $\mathscr{D}_{XY}^{D_{XY}}:=\{D_{XY}^{\alpha}: \forall \alpha \in [0,1)\}$.
 \vspace{0.2cm}
% %\vspace{0.5cm}
\\
\noindent$\bullet$ Total space $\mathscr{D}_{XY}^{\rm all}$, which consists of all domains.
\vspace{0.2cm}
% \vspace{-0.5cm}
\\
\noindent$\bullet$ Separate space $\mathscr{D}_{XY}^{s}$, which consists of all domains that satisfy the separate condition, that is for any $D_{XY}\in \mathscr{D}_{XY}^{s}$,
% \begin{equation*}
$
    {\rm supp} D_{X_{\rm O}}\cap   {\rm supp} D_{X_{\rm I}}=\emptyset,
$
% \end{equation*}
where ${\rm supp}$ means the support set.
\vspace{0.2cm}
% \vspace{-0.5cm}
\\
\noindent $\bullet$ Finite-ID-distribution space $\mathscr{D}_{XY}^{F}$, which is a prior-unknown space satisfying that the number of distinct ID joint distributions $D_{X_{\rm I}Y_{\rm I}}$ in $\mathscr{D}_{XY}^{F}$ is finite, \textit{i.e.}, $|\{D_{X_{\rm I}Y_{\rm I}}:\forall D_{XY}\in \mathscr{D}_{XY}^{F}\}|<+\infty$.
\vspace{0.1cm}
%\vspace{-0.2em}
\\
\noindent$\bullet$ Density-based space $\mathscr{D}^{\mu,b}_{XY}$, which is a prior-unknown space consisting of some domains satisfying that: for any $D_{XY}$, there exists a density function $f$ with $1/b\leq f\leq b$ in ${\rm supp}\mu$ and $0.5*D_{X_{\rm I}}+0.5*D_{X_{\rm O}}= \int f {\rm d}\mu$, where $\mu$ is a measure defined over $\mathcal{X}$. Note that if $\mu$ is discrete, then $D_{X}$ is a discrete distribution; and if $\mu$ is the Lebesgue measure, then $D_{X}$ is a continuous distribution.

The above representative spaces widely exist in real applications. For example, 1) if the images from different semantic labels with different styles are clearly different, then those images can form a distribution belonging to a separate space $\mathscr{D}_{XY}^{s}$; and 2) when designing an algorithm, we only have finite ID datasets, \textit{e.g.}, CIFAR-10, MNIST, SVHN, and ImageNet, to build a model. Then, finite-ID-distribution space $\mathscr{D}_{XY}^{F}$ can handle this real scenario. Note that the single-distribution space is a special case of the finite-ID-distribution space. In this paper, we mainly discuss these five spaces. 

\begin{figure}[!t]
    \centering
   \footnotesize
    \subfigure[ID and OOD Distributions]{
    \raisebox{0.22\height}{\includegraphics[width=0.33\linewidth]{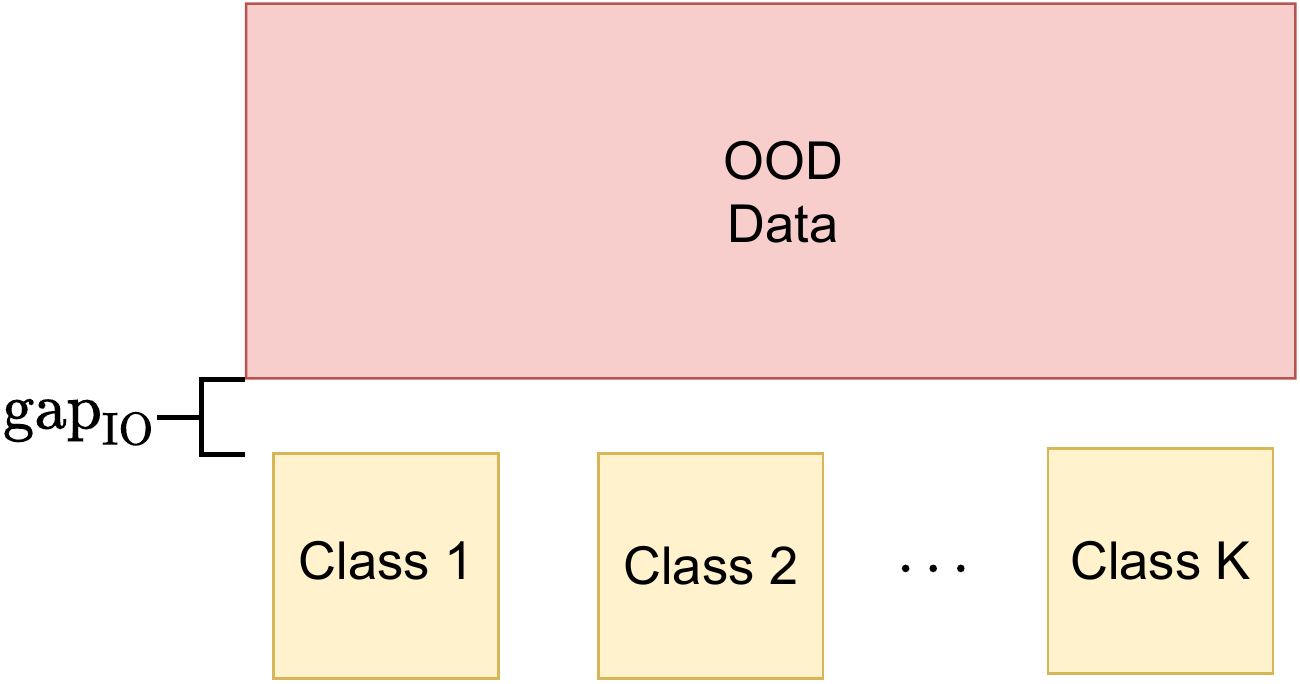}}}~~~~
    \subfigure[Overlap Exists]{
    \includegraphics[width=0.3\linewidth]{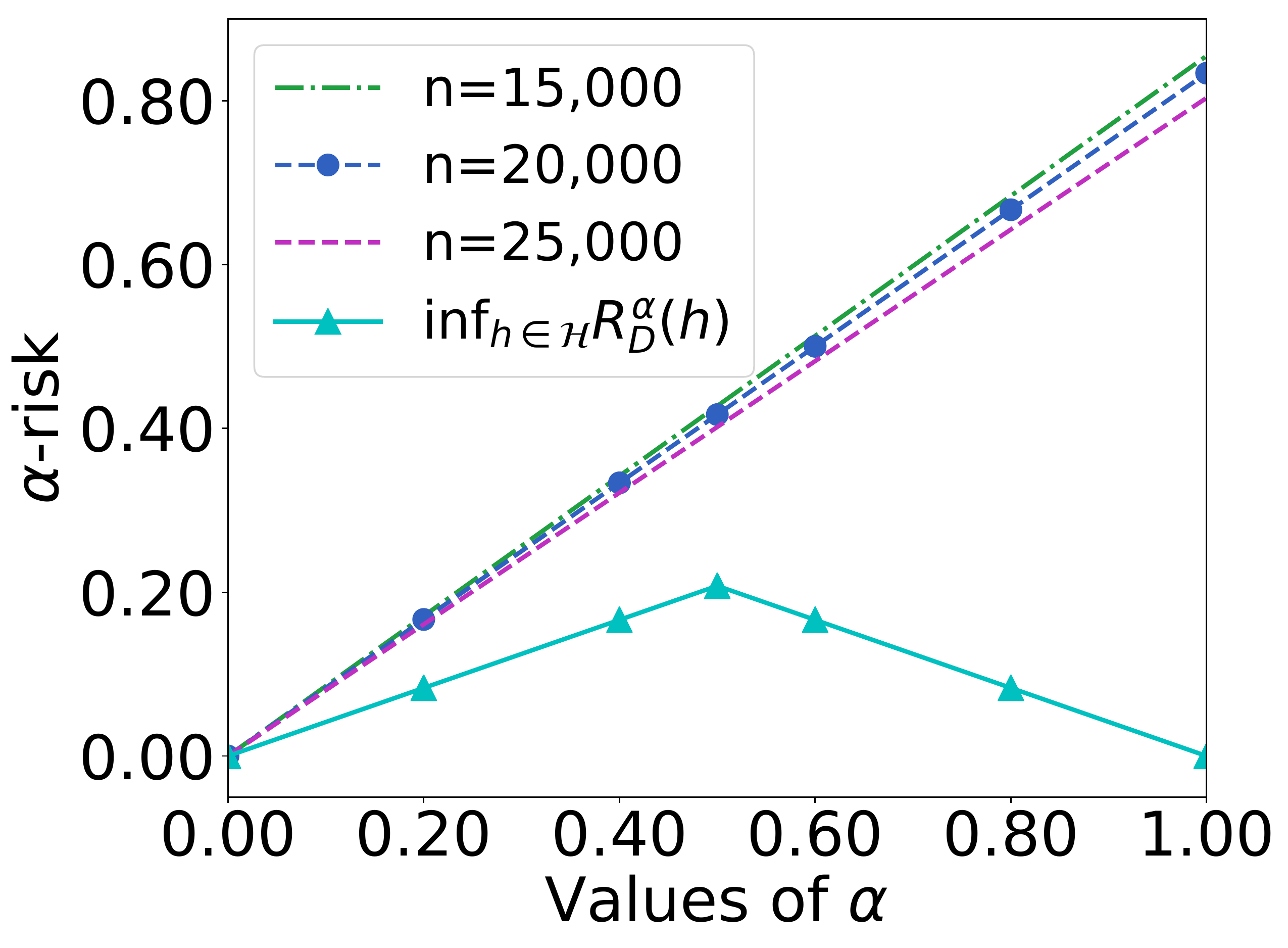}}
    \subfigure[No Overlap]{
    \includegraphics[width=0.3\linewidth]{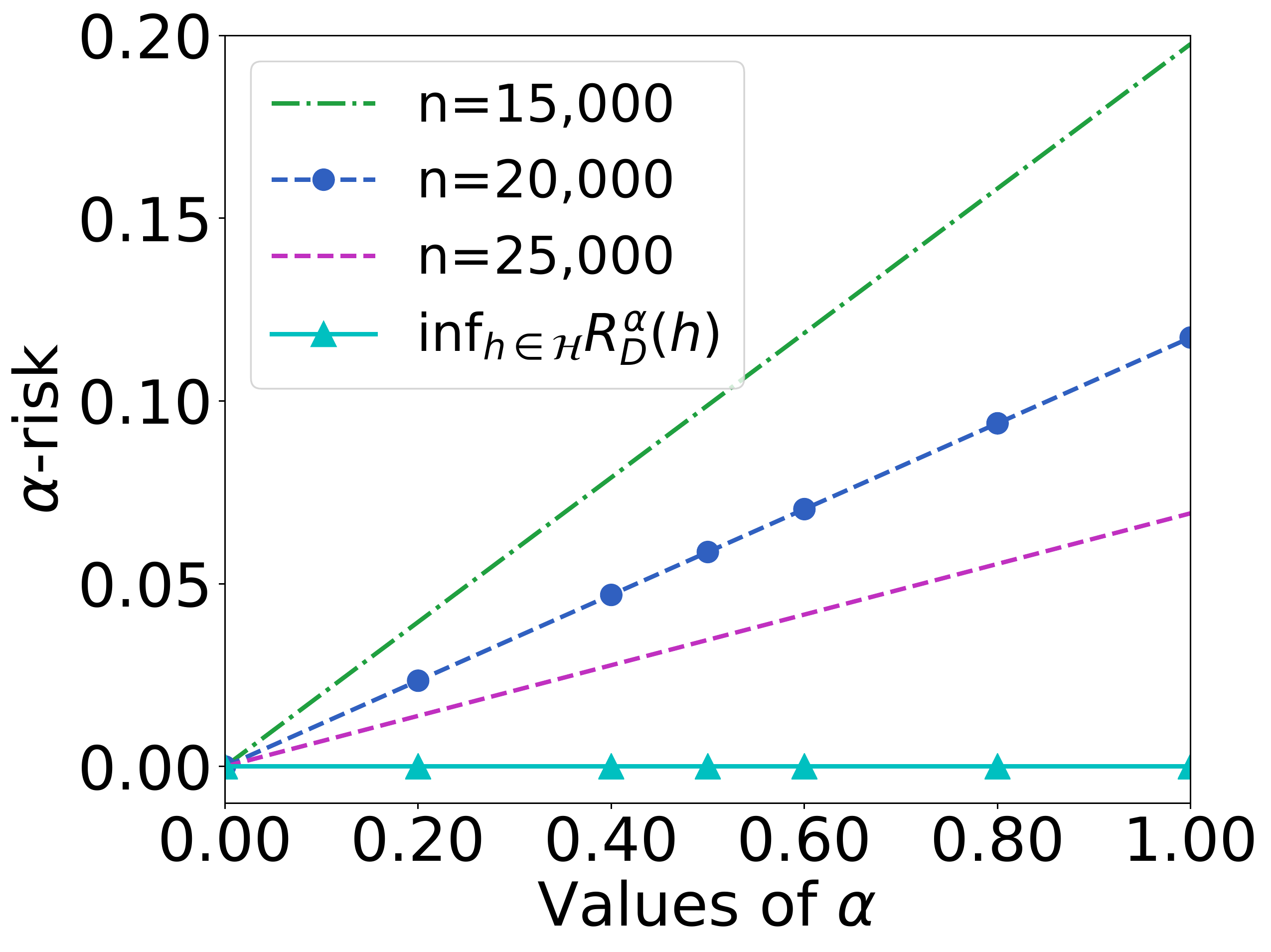}}
    \vspace{-1em}
    \caption{\footnotesize Illustration of $\inf_{h\in \mathcal{H}}R_{D}^{\alpha}({h})$ (solid lines with triangle marks) and the estimated $\mathbb{E}_{S\sim D^n_{\rm in}}R^{\alpha}_D(\mathbf{A}(S))$ (dash lines) with $\alpha\in [0,1)$ in different scenarios, where $D_{\rm in}=D_{X_{\rm I}Y_{\rm I}}$ and the algorithm $\mathbf{A}$ is the free-energy OOD detection method \cite{liu2020energy}. 
    Subfigure (a) shows the ID and OOD distributions. In (a), ${\rm gap}_{\rm IO}$ represents the distance between the support sets of ID and OOD distributions.
    In (b), since there is an overlap between ID and OOD data, the solid line is a ployline. In (c), since there is no overlap between ID and OOD data, we can check that $\inf_{h\in \mathcal{H}}R_{D}^{\alpha}({h})$ forms a straight line (the solid line). 
    However, since dash lines are always straight lines, two observations can be obtained from (b) and (c): 1) dash lines cannot approximate the solid ployline in (b), which implies the unlearnability of OOD detection; and 2) the solid line in (c) is a straight line and may be approximated by the dash lines in (c). The above observations motivate us to propose Condition~\ref{C1}.
    % Details of this experiment can be found in Appendix~\ref{Asec:detail_of_fig}.
    }
    \label{fig:only}
    \vspace{-1.5em}
\end{figure}

%\vspace{-0.8em}
\section{Impossibility Theorems for OOD Detection}

In this section, we first give a necessary condition for the learnability of OOD detection. Then, we show this necessary condition does not hold in the total space $\mathscr{D}_{XY}^{\rm all}$ and the separate space $\mathscr{D}_{XY}^{s}$. 
% Before demonstrating the necessary conditions, OOD convex decomposition and convex domains are introduced below.
% \begin{Definition}[OOD Convex Decomposition and Convex Domains]\label{OODCD}
% Given any domain $D_{XY}\in \mathcal{D}_{XY}$, we say joint distributions $Q_1,...,Q_l$, which are defined over $\mathcal{X}\times \mathcal{Y}^{\rm out}$, are the OOD convex decomposition for $D_{XY}$, if 
% \begin{equation*}
% D_{XY}=(1-\sum_{j=1}^l \lambda_j)D_{X_{\rm I}Y_{\rm I}}+\sum_{j=1}^l \lambda_j Q_j,
% \end{equation*}
% for some  $(\lambda_1,...,\lambda_l)\in \Delta_l^{\rm o}$. We also say domain $D_{XY}\in \mathcal{D}_{XY}$ is an OOD convex domain corresponding to OOD convex decomposition $Q_1,...,Q_l$, if for any $(\alpha_1,...,\alpha_l)\in \Delta_l^{\rm o}$, 
% \begin{equation*}
%   (1-\sum_{j=1}^l \alpha_j)D_{X_{\rm I}Y_{\rm I}}+\sum_{j=1}^l \alpha_j Q_j\in \mathcal{D}_{XY}.
% \end{equation*}
% \end{Definition}

\textbf{Necessary Condition.} We find a necessary condition for the learnability of OOD detection, \textit{i.e.}, Condition \ref{C1}, motivated by the experiments in Figure \ref{fig:only}. Details of Figure \ref{fig:only} can be found in Appendix~\ref{Asec:detail_of_fig}. 
%In {Appendix} \ref{SD}, a general version (Condition \ref{C2}) is offered.
\begin{Condition}[Linear Condition]\label{C1}
For any $D_{XY}\in \mathscr{D}_{XY}$ and any $\alpha \in [0,1)$,
\vspace{-0.1cm}
\begin{equation*}
   \inf_{h\in \mathcal{H}}R_D^{\alpha}(h)= (1-\alpha)\inf_{h\in \mathcal{H}}R_D^{\rm in}(h)+\alpha \inf_{h\in \mathcal{H}}R_D^{\rm out}(h).
\end{equation*}
\end{Condition}
\vspace{-0.2em}
%Then, Theorem \ref{T2} shows Condition \ref{C1} is a necessary condition for the learnability of OOD detection.
% \begin{theorem}
%\begin{restatable}{theorem}{thmNessry}
%\label{T2}
%Given a domain space $\mathscr{D}_{XY}$ and a hypothesis space $\mathcal{H}$, if OOD detection is learnable in the domain space $\mathscr{D}_{XY}$ for the hypothesis space $\mathcal{H}$, then linear condition (i.e., Condition \ref{C1}) holds.
%\end{restatable}
% \end{theorem}
% \vspace{-0.2cm}
To reveal the importance of Condition \ref{C1}, Theorem \ref{T3} shows that Condition \ref{C1} is a \textit{necessary and sufficient} condition for the learnability of OOD detection if the $\mathscr{D}_{XY}$ is the single-distribution space. 

\fbox{
\parbox{0.98\textwidth}{
% \begin{theorem}
\vspace{-0.4em}
\begin{restatable}{theorem}{thmCondoneIff}
\label{T3}
Given a hypothesis space $\mathcal{H}$ and a domain $D_{XY}$, OOD detection is learnable in the single-distribution space $\mathscr{D}_{XY}^{D_{XY}}$ for $\mathcal{H}$ \textbf{if and only if} linear condition (i.e., Condition \ref{C1}) holds.
\end{restatable}
% \end{theorem}
\vspace{-0.3em}
}
}

Theorem \ref{T3} implies that Condition \ref{C1} is important for the learnability of OOD detection. Due to the simplicity of single-distribution space, Theorem \ref{T3} implies that  Condition \ref{C1} is the necessary condition for the learnability of OOD detection in the prior-unknown space, see Lemma \ref{C1andC2} in Appendix \ref{SD}.

%This also implies that it is worthy studying the learnability of OOD detection by using Condition \ref{C1}.

\textbf{Impossibility Theorems.} Here, we first study whether Condition \ref{C1} holds in the total space $\mathscr{D}_{XY}^{\rm all}$. If Condition \ref{C1} does not hold, then OOD detection is not learnable.
% Since Condition \ref{C1} is a necessary condition for the learnability of OOD detection, it is interesting to investigate if this condition can hold in general scenarios, e.g., the total space $\mathscr{D}_{XY}^{\rm all}$. 
Theorem~\ref{T5} shows that Condition \ref{C1} is not always satisfied, 
especially, when there is an overlap between the ID and OOD distributions:
\begin{Definition}[Overlap Between ID and OOD]
\label{def:overlap}
We say a domain $D_{XY}$ has overlap between ID and OOD distributions, if there is a $\sigma$-finite measure $\tilde{\mu}$ such that $D_X$ is absolutely continuous with respect to $\tilde{\mu}$, and
%\vspace{-0.3cm}
% \begin{equation*}
$
 \tilde{\mu}(A_{\rm overlap})>0,~\textit{ where}
$
% \end{equation*}
$A_{\rm overlap}= \{\mathbf{x}\in \mathcal{X}:f_{\rm I}(\mathbf{x})>0~ \textit{and}~ f_{\rm O}(\mathbf{x})>0\}$. Here $f_{\rm I}$ and $f_{\rm O}$ are the representers of $D_{X_{\rm I}}$ and $D_{X_{\rm O}}$ in Radon–Nikodym Theorem \cite{cohn2013measure}, 
\begin{equation*}
  D_{X_{\rm I}} = \int f_{\rm I} {\rm d}\tilde{\mu},~~~ D_{X_{\rm O}} = \int f_{\rm O} {\rm d}\tilde{\mu}.
\end{equation*}
\end{Definition}
% In this section, we briefly discuss how the overlap between OOD and ID effects the learnability of OOD detection.
% \begin{theorem}
\begin{restatable}{theorem}{thmImpOne}
\label{T5}
Given a hypothesis space $\mathcal{H}$ and  a prior-unknown space $\mathscr{D}_{XY}$, if there is $D_{XY}\in \mathscr{D}_{XY}$, which has overlap between ID and OOD, and $\inf_{h\in \mathcal{H}} R_{D}^{\rm in}(h)=0$ and $\inf_{h\in \mathcal{H}} R_{D}^{\rm out}(h)=0$, then Condition \ref{C1} does not hold. Therefore, OOD detection is not learnable in $\mathscr{D}_{XY}$ for $\mathcal{H}$.
\end{restatable}
% \end{theorem}
\vspace{-0.1cm}
Theorem~\ref{T5} clearly shows that under proper conditions, Condition \ref{C1} does not hold, if there exists a domain whose ID and OOD distributions have overlap. By Theorem \ref{T5}, we can obtain that  
the OOD detection is not learnable in the total space $\mathscr{D}^{\rm all}_{XY}$ for any non-trivial hypothesis space $\mathcal{H}$.

\fbox{
\parbox{0.98\textwidth}{
% \begin{theorem}[Impossibility Theorem for Total Space]
\vspace{-0.4em}
\begin{restatable}[Impossibility Theorem for Total Space]{theorem}{thmImpTotal}
\label{T4}
OOD detection is not learnable in the total space $\mathscr{D}^{\rm all}_{XY}$ for $\mathcal{H}$, if $|\phi\circ\mathcal{H}|>1$, where $\phi$ maps ID labels to $1$ and maps OOD labels to $2$.
\end{restatable}
% \end{theorem}
\vspace{-0.4em}
}
}

Since the overlaps between ID and OOD distributions may cause that Condition~\ref{C1} does not hold, we then consider studying
the learnability of OOD detection in the separate space $\mathscr{D}_{XY}^s$, where there are no overlaps between the ID
and OOD distributions. However, Theorem \ref{T12} shows that even if we consider the separate space, the OOD detection is still not learnable in some scenarios. Before introducing the impossibility theorem for separate space, \textit{i.e.}, Theorem \ref{T12}, we need a mild assumption:
\vspace{-0.3cm}
\begin{assumption}[Separate Space for OOD]\label{ass1}
A hypothesis space $\mathcal{H}$ is separate for OOD data, if for each data point $\mathbf{x}\in \mathcal{X}$, there exists at least one hypothesis function $h_{\mathbf{x}}\in \mathcal{H}$ such that $h_{\mathbf{x}}(\mathbf{x})=K+1$.
\end{assumption}
\vspace{-0.2cm}
Assumption \ref{ass1} means that every data point $\mathbf{x}$ has the possibility to be detected as OOD data. Assumption \ref{ass1} is mild and can be satisfied by many hypothesis spaces, {\textit{e.g.}, the FCNN-based hypothesis space} (Proposition \ref{Pr1} in Appendix \ref{SK}), score-based hypothesis space  (Proposition \ref{P2} in Appendix \ref{SK}) and universal kernel space. Next, we use \textit{Vapnik–Chervonenkis} (VC) dimension \cite{mohri2018foundations} to measure the size of hypothesis space, and study the learnability of OOD detection in $\mathscr{D}_{XY}^{s}$ based on the VC dimension. 
\vspace{-0.3cm}
% \begin{theorem}

\fbox{
\parbox{0.98\textwidth}{
\vspace{-0.5em}
\begin{restatable}[Impossibility Theorem for Separate Space]{theorem}{thmImpSeptwo}
\label{T12}
 If Assumption \ref{ass1} holds, 
  ${\rm VCdim}(\phi\circ \mathcal{H})<+\infty$ and $\sup_{{ h}\in \mathcal{H}} |\{\mathbf{x}\in \mathcal{X}: { h}(\mathbf{x})\in \mathcal{Y}\}|=+\infty$, then
  OOD detection is not learnable in separate space $\mathscr{D}_{XY}^{s}$ for $\mathcal{H}$, where $\phi$ maps ID labels to ${1}$ and maps OOD labels to $2$.
\end{restatable}
\vspace{-0.4em}
}}

% Furthermore, Theorem \ref{T12} implies the following corollary.

%\vspace{-0.2cm}
The finite VC dimension normally implies the learnability of supervised learning. However, in our results, the finite VC dimension cannot guarantee the learnability of OOD detection in the separate space, which reveals the difficulty of the OOD detection.
Although the above impossibility theorems are frustrating, there is still room to discuss the conditions in Theorem~\ref{T12}, and to find out the proper conditions for ensuring the learnability of OOD detection in the separate space (see Sections \ref{S6} and \ref{S7}).

\vspace{-0.3em}
\section{When OOD Detection Can Be Successful}\label{S6}
\vspace{-0.3em}
Here, we discuss when the OOD detection can be learnable in the separate space $\mathscr{D}_{XY}^s$, finite-ID-distribution space $\mathscr{D}_{XY}^{F}$ and density-based space $\mathscr{D}_{XY}^{\mu,b}$. We first study the separate space $\mathscr{D}_{XY}^s$.

\textbf{OOD Detection in the Separate Space.} 
Theorem \ref{T12} has indicated that ${\rm VCdim}(\phi\circ \mathcal{H})=+\infty$ or $\sup_{{ h}\in \mathcal{H}} |\{\mathbf{x}\in \mathcal{X}: { h}(\mathbf{x})\in \mathcal{Y}\}|<+\infty$ is necessary to ensure the learnability of OOD detection in $\mathscr{D}_{XY}^{s}$ if Assumption \ref{ass1} holds. However, generally, hypothesis spaces generated by feed-forward neural networks with proper activation functions have finite VC dimension \cite{Peter2019Nearly,Karpinski1997polynomial}. 
% In addition, the PAC learning theory for supervised learning has shown that if a binary classification hypothesis space has infinite VC-dimension, then the hypothesis space is not PAC learnable.
Therefore, we study the learnability of OOD detection in the case that $|\mathcal{X}|<+\infty$, which implies that $\sup_{{ h}\in \mathcal{H}} |\{\mathbf{x}\in \mathcal{X}: { h}(\mathbf{x})\in \mathcal{Y}\}|<+\infty$. Additionally, Theorem~\ref{T24} also implies that $|\mathcal{X}|<+\infty$ is the necessary and sufficient condition for the learnability of OOD detection in separate space, when the hypothesis space is generated by FCNN. Hence, $|\mathcal{X}|<+\infty$ may be necessary in the space $\mathscr{D}_{XY}^s$.

% \textbf{One-class OOD Detection.} 
For simplicity, we first discuss the case that $K = 1$, \textit{i.e.}, the one-class novelty detection.
We show the {necessary and sufficient} condition for the learnability of OOD detection in $\mathscr{D}_{XY}^{s}$, when $|\mathcal{X}|<+\infty$.

\fbox{
\parbox{0.97\textwidth}{
\vspace{-0.4em}
\begin{restatable}{theorem}{thmPosbSep}
\label{T13}
Let $K=1$ and $|\mathcal{X}|<+\infty$. Suppose that Assumption \ref{ass1} holds and the constant function $h^{\rm in}:=1 \in \mathcal{H}$. Then OOD detection is learnable in $\mathscr{D}_{XY}^{s}$ for $\mathcal{H}$ \textbf{if and only if} $\mathcal{H}_{\rm all}-\{h^{\rm out}\} \subset \mathcal{H}$, where $\mathcal{H}_{\rm all}$ is the hypothesis space consisting of all hypothesis functions, and $h^{\rm out}$ is a constant function that $h^{\rm out}:=2$, here $1$ represents ID data and $2$ represents OOD data.
\end{restatable}
\vspace{-0.4em}
}
}
%\vspace{-0.1em}

The condition $h^{\rm in}\in \mathcal{H}$ presented in Theorem \ref{T13} is mild. {Many practical hypothesis spaces satisfy this condition, {\textit{e.g.}, the FCNN-based hypothesis space} (Proposition \ref{Pr1} in Appendix \ref{SK}), score-based hypothesis space  (Proposition \ref{P2} in Appendix \ref{SK}) and universal kernel-based hypothesis space.} Theorem \ref{T13} implies that if $K=1$ and OOD detection is learnable in $\mathscr{D}^{s}_{XY}$ for $\mathcal{H}$, then the hypothesis space $\mathcal{H}$ should contain almost all hypothesis functions, implying that if the OOD detection can be learnable in the distribution-agnostic case, then a large-capacity model is necessary.
 
%The condition $|\mathcal{X}|<+\infty$ is important for the separate space $\mathscr{D}_{XY}^{s}$. However, Theorem \ref{T16} shows that in $\tau$-margin space $\mathscr{D}_{XY}^{\tau}$, the condition $|\mathcal{X}|<+\infty$ can be eased. Specifically, in $\tau$-margin space, we only need $\mathcal{X}$ to be a bounded set.

% \begin{theorem}

%\vspace{-0.2cm}
%Note that when the feature space $\mathcal{X}$ is a bounded and discrete set,  the universal kernel-based hypothesis space satisfies the condition that $\mathcal{H}_{\rm all}-\{h^{\rm out}\} \subset \mathcal{H}$, even if $|\mathcal{X}|=+\infty$.  
%Theorem~\ref{T16} shows that the OOD detection is promising to be addressed in theory, when the domains belong to $\mathscr{D}_{XY}^{\tau}$. 

% \textbf{Multi-class OOD Detection.} 
Next, we extend Theorem \ref{T13} to a general case, \textit{i.e.}, $K>1$. 
When $K>1$, we will first use a binary classifier $h^b$ to classify the ID and OOD data. Then, for the ID data identified by $h^b$, an ID hypothesis function $h^{\rm in}$ will be used to classify them into corresponding ID classes. 
We state this strategy as follows: given a hypothesis space $\mathcal{H}^{\rm in}$ for ID distribution and a binary classification hypothesis space $\mathcal{H}^{\rm b}$ introduced in Section \ref{S3}, we use $\mathcal{H}^{\rm in}$ and $\mathcal{H}^{\rm b}$ to construct an OOD detection's hypothesis space $\mathcal{H}$, which consists of all hypothesis functions $h$ satisfying the following condition: there exist $h^{\rm in}\in \mathcal{H}^{\rm in}$ and $h^{\rm b}\in \mathcal{H}^b$ such that for any $\mathbf{x}\in \mathcal{X}$,
\begin{equation}\label{Eq.dot}
    h(\mathbf{x}) = i,~~\textnormal{ if}~h^{\rm in}(\mathbf{x})=i~ \textnormal{ and }~h^{\rm b}(\mathbf{x})=1; \textnormal{ otherwise},  ~h(\mathbf{x}) =  K+1.
\end{equation}
%\begin{equation}\label{Eq.dot}
 %   h(\mathbf{x}) = \left \{
  %  \begin{aligned}
  % &~~~~~~~~~i,~~\textnormal{ if}~h^{\rm in}(\mathbf{x})=i~ \textnormal{ and }~h^{\rm b}(\mathbf{x})=1;\\ 
%   &  K+1,~~\textnormal{ if}~h^{\rm b}(\mathbf{x})=2.
%    \end{aligned}
 %   \right.
%\end{equation}
We use $\mathcal{H}^{\rm in} \bullet \mathcal{H}^{\rm b} $ to represent a hypothesis space consisting of all $h$ defined in Eq. (\ref{Eq.dot}). In addition, we also need an additional condition for the loss function $\ell$. This condition is shown as follows:
%\begin{assumption}[Agnostic PAC Learnability in Supervised Learning for Classification \cite{shalev2010learnability}]\label{ass2}
%There exist an algorithm $\mathbf{A}^{\rm in}: \cup_{n=1}^{+\infty}(\mathcal{X}\times)^n\rightarrow \mathcal{H}^{\rm in}$ and a monotonically decreasing
%sequence $\epsilon_{\rm cons}^{\rm in}(n)$, such that $\epsilon_{\rm cons}^{\rm in}(n)\rightarrow 0$, as $n\rightarrow +\infty$, and for any joint distribution $D_{XY}$ defined over $\mathcal{X}\times$, 
%\begin{align*}
%    \mathbb{E}_{S\sim D^n_{XY}}\big[ R_D(\mathbf{A}^{\rm in}(S))- \inf_{h^{\rm in}\in \mathcal{H}^{\rm in}}R_D(h^{\rm in})\big] \leq \epsilon_{\rm cons}^{\rm in}(n).
%\end{align*}
%\end{assumption}
%\vspace{-1em}
%If $\mathcal{H}^{\rm in}$ has finite Natarajan dimension \cite{shalev2014understanding} or finite Graph dimension \cite{DBLP:journals/jmlr/DanielySBS11}, then Assumption \ref{ass2} holds.

\begin{Condition}\label{C3}
$
\ell(y_2,y_1)\leq \ell(K+1,y_1)$, for any in-distribution labels $y_1$ and $y_2\in \mathcal{Y}.
$
\end{Condition}
% \begin{theorem}
\begin{restatable}{theorem}{thmPosbMultione}
\label{T17}
Let $|\mathcal{X}|<+\infty$ and $\mathcal{H}=\mathcal{H}^{\rm in} \bullet \mathcal{H}^{\rm b}$. If $\mathcal{H}_{\rm all}-\{h^{\rm out}\} \subset \mathcal{H}^{\rm b}$ and Condition \ref{C3} holds, then OOD detection is learnable in $\mathscr{D}_{XY}^{s}$ for $\mathcal{H}$, where $\mathcal{H}_{\rm all}$ and $h^{\rm out}$ are defined in Theorem \ref{T13}.
\end{restatable}
% \end{theorem}

% \end{theorem}

% Note that, Theorem \ref{T17} is the corollaries of Theorems \ref{T13}. 
% Theorem \ref{T17.1} indicates that it is possible to realize the learnability of OOD detection in the $\tau$-margin space $\mathscr{D}_{XY}^{\tau}$.

\textbf{OOD Detection in the Finite-ID-Distribution Space.} Since researchers can only collect finite ID datasets as the training data in the process of algorithm design, it is worthy to study the learnability of OOD detection in the finite-ID-distribution space $\mathscr{D}_{XY}^F$. We first show two necessary concepts below. 
\begin{Definition}[ID Consistency] Given a domain space $\mathscr{D}_{XY}$, we say any two domains $D_{XY}\in \mathscr{D}_{XY}$ and $D_{XY}'\in \mathscr{D}_{XY}$ are ID consistency, if $D_{X_{\rm I}Y_{\rm I}}=D_{X_{\rm I}Y_{\rm I}}'$. We use the notation $\sim$ to represent the ID consistency, i.e., $D_{XY}\sim D_{XY}'$ if and only if $D_{XY}$ and $D_{XY}'$ are ID consistency.
\end{Definition}
It is easy to check that the ID consistency $\sim$ is an equivalence relation. Therefore, we define the set $[D_{XY}]:=\{D_{XY}'\in \mathscr{D}_{XY}:D_{XY}\sim D_{XY}'\}$ as the equivalence class with respect to space $\mathscr{D}_{XY}$.% Then, we introduce the following condition. %The set of all equivalence classes with respect to $\mathscr{D}_{XY}$ is called the quotient set, \textit{i.e.}, ${\mathscr{D}_{XY}}{/\sim}:=\{[D_{XY}]:D_{XY}\in \mathscr{D}_{XY}\}$.

%Given an equivalence class $[D_{XY}]$ with respect to $\mathscr{D}_{XY}^{\mu,M}$, we  set $D_{XY}^{\rm max}$ as the maximum domain with respect to $[D_{XY}]$, \textit{i.e.}, $D_{XY}^{\rm max}:=0.5*D_{XY|Y\in \mathcal{Y}}+0.5*D_{\mu}$, where $D_{\mu}$ is an OOD distribution, whose marginal distribution is $\mu|_A$, here $A=\cup_{D_{XY}'\in[D_{XY}]} {\rm supp} D_X'$ and $\mu|_A$ is a measure restricted to $A$, \textit{i.e.}, $\mu|_A(B)=\mu(A\cap B)/\mu(A)$. Next, we introduce the compatibility condition as follows. 

\begin{Condition}[Compatibility]\label{Con2}
    For any equivalence class $[D_{XY}']$ with respect to $\mathscr{D}_{XY}$ and any $\epsilon>0$, there exists a hypothesis function $h_{\epsilon}\in \mathcal{H}$ such that for any domain $D_{XY}\in [D_{XY}']$,
 \begin{equation*}
 h_{\epsilon}\in \{ h' \in \mathcal{H}: R_D^{\rm out}(h') \leq \inf_{h\in \mathcal{H}} R_D^{\rm out}(h)+\epsilon\} \cap   \{ h' \in \mathcal{H}: R_D^{\rm in}(h') \leq \inf_{h\in \mathcal{H}} R_D^{\rm in}(h)+\epsilon\}.
 \end{equation*}
\end{Condition}
%\vspace{-1em}

In Appendix \ref{SD}, Lemma \ref{Lemma1} has implied that Condition \ref{Con2} is a general version of Condition \ref{C1}. Next, Theorem \ref{T-SET} indicates that Condition \ref{Con2} is the \textit{necessary and sufficient condition} in the space $\mathscr{D}_{XY}^F$.

\fbox{
\parbox{0.98\textwidth}{
\vspace{-0.4em}
\begin{restatable}{theorem}{thmPosbtennew}\label{T-SET}
Suppose that $\mathcal{X}$ is a bounded set. OOD detection is learnable in the finite-ID-distribution space $\mathscr{D}_{XY}^{F}$ for $\mathcal{H}$ \textbf{if and only if} the compatibility condition (i.e., Condition \ref{Con2}) holds. Furthermore, the learning rate $\epsilon_{\rm cons}(n)$ can attain ${O}(1/\sqrt{n^{1-\theta}})$, for any $\theta\in (0,1)$.
\end{restatable}
\vspace{-0.4em}
}}
%\vspace{-0.3em}

Theorem~\ref{T-SET} shows that, in the process of algorithm design, OOD detection cannot be successful without the compatibility condition. Theorem \ref{T-SET} also implies that Condition \ref{Con2} is essential for the learnability of OOD detection. This motivates us to study whether OOD detection can be successful in more general spaces (\textit{e.g.}, the density-based space), when the compatibility condition holds.

\textbf{OOD Detection in the Density-based Space.} To ensure that Condition \ref{Con2} holds, we consider a basic assumption in learning theory---\textit{Realizability Assumption} (see Appendix \ref{SB.2}), \textit{i.e.}, for any $D_{XY}\in \mathscr{D}_{XY}$, there exists $h^*\in \mathcal{H}$ such that $R_D(h^*)=0$. We discover that in the density-based space $\mathscr{D}_{XY}^{\mu,b}$, Realizability Assumption can conclude the compatibility condition (\textit{i.e.}, Condition \ref{Con2}). Based on this observation, we can prove the following theorem:% \textit{i.e.}, \textit{there exists a hypothesis function $h^*\in \mathcal{H}$ such that $R_{D}(h^*)=0$}. 

\fbox{
\parbox{0.98\textwidth}{
\vspace{-0.6em}
\begin{restatable}{theorem}{thmPosbtennewtwo}\label{T-SET2}
Given a density-based space $\mathscr{D}_{XY}^{\mu,b}$, if $\mu(\mathcal{X})<+\infty$, the Realizability Assumption holds, then when $\mathcal{H}$ has finite Natarajan dimension \cite{shalev2014understanding}, OOD detection is learnable in $\mathscr{D}_{XY}^{\mu,b}$ for $\mathcal{H}$. Furthermore, the learning rate $\epsilon_{\rm cons}(n)$ can attain ${O}(1/\sqrt{n^{1-\theta}})$, for any $\theta\in (0,1)$.
\end{restatable}
\vspace{-0.2em}
}}
%\vspace{-0.1em}
%Resum

To further investigate the importance and necessary of Realizability Assumption, Theorem \ref{T24.3} has indicated that in some practical scenarios, Realizability Assumption is the necessary and sufficient condition for the learnability of OOD detection in the density-based space. Therefore, Realizability Assumption may be indispensable for the learnability of OOD detection in some practical scenarios.

% \begin{remark}

% \end{remark}

\section{Connecting Theory to Practice}\label{S7}
In Section \ref{S6}, we have shown the successful scenarios where OOD detection problem can be addressed in theory. In this section, we will discuss how the proposed theory is applied to two representative hypothesis spaces---neural-network-based hypothesis spaces and score-based hypothesis spaces. 
% Lastly, we extend our analysis to modern \emph{convolutional neural networks} (CNNs) {in Appendix} {\ref{SO}}.
% \subsection{Fully-connected Neural Networks}

\textbf{Fully-connected Neural Networks.}
 Given a sequence $\mathbf{q}=(l_1,l_2,...,l_g)$, where $l_i$ and $g$ are positive integers and $g>2$, we use $g$ to represent the \textbf{\textit{depth}} of neural network and use $l_i$ to represent the \textbf{\textit{width}} of the $i$-th layer. After the activation function $\sigma$ is selected\footnote{We consider the \emph{rectified linear unit} (ReLU) function as the default activation function $\sigma$, which is defined by $\sigma(x)=\max \{x,0 \}$,~$\forall$ $ x\in \mathbb{R}$. 
\textit{{We will not repeatedly mention the definition of $\sigma$ in the rest of our paper}}.
}, we can obtain the architecture of FCNN according to the sequence $\mathbf{q}$. Let $\mathbf{f}_{\mathbf{w},\mathbf{b}}$ be the function generated by FCNN with weights $\mathbf{w}$ and bias $\mathbf{b}$. An FCNN-based scoring function space is defined as:
$
    \mathcal{F}_{\mathbf{q}}^{\sigma}:=\{\mathbf{f}_{\mathbf{w},\mathbf{b}}:\forall~ \textnormal{weights}~\mathbf{w},~\forall~  \textnormal{bias}~ \mathbf{b}\}.
$ In addition, for simplicity, given any two sequences $\mathbf{q}=(l_1,...,l_g)$ and $\mathbf{q}'=(l_1',...,l_{g'}')$, we use the notation
$
    \mathbf{q}\lesssim\mathbf{q}'
$
to represent the  following equations and inequalities:
\begin{equation*}
1)~ g \leq g', l_1=l_1', l_g=l_{g'}'; ~~~~~2)~ l_i\leq l'_i,~\forall i=1,...,g-1;~~\textnormal{and}~~~3)~
l_{g-1}\leq l'_i, ~\forall i=g,...,g'-1.
\end{equation*}
In Appendix \ref{SL}, Lemma \ref{L7-contain} shows $\mathbf{q}\lesssim\mathbf{q}'\Rightarrow \mathcal{F}_{\mathbf{q}}^{\sigma}\subset \mathcal{F}_{\mathbf{q}'}^{\sigma}$. We use $\lesssim$ to compare the sizes of FCNNs.

\textbf{FCNN-based Hypothesis Space.}
Let $l_g=K+1$. The FCNN-based scoring function space $\mathcal{F}_{\mathbf{q}}^{\sigma}$ can induce an FCNN-based hypothesis space. For any $\mathbf{f}_{\mathbf{w},\mathbf{b}}\in \mathcal{F}_{\mathbf{q}}^{\sigma}$, the induced hypothesis function  is:
\begin{equation*}
    h_{\mathbf{w},\mathbf{b}}:=\argmax_{k\in\{1,...,K+1\}} {f}^k_{\mathbf{w},\mathbf{b}}, ~\textnormal{where}~{f}^k_{\mathbf{w},\mathbf{b}}~\textnormal{is ~the~}k\textnormal{-th}~\textnormal{coordinate~of}~\mathbf{f}_{\mathbf{w},\mathbf{b}}.
\end{equation*}
 Then, the FCNN-based hypothesis space is defined as
$
  \mathcal{H}_{\mathbf{q}}^{\sigma}:= \{h_{\mathbf{w},\mathbf{b}}:\forall~ \textnormal{weights}~\mathbf{w},~\forall~\textnormal{bias}~ \mathbf{b}\}.
$
%There are several works \cite{Fang2021learning, Zhong2021Bridging}, which use the FCNN-based hypothesis spaces $\mathcal{H}_{\mathbf{q}}^{\sigma}$ in their papers. %Next, we introduce an important proposition for the FCNN-based hypothesis space $\mathcal{H}_{\mathbf{q}}^{\sigma}$.

% \begin{Proposition}
%\begin{restatable}{Proposition}{propOne}
%\label{P1}
%For any sequence $\mathbf{q}=(l_1,...l_g)$  with $l_1=d$ and $l_g=K+1$, the constant functions $h_{1}$,...,$h_{K+1}\in \mathcal{H}_{\mathbf{q}}^{\sigma}$, where $h_{i}:=i$. Therefore, Assumption \ref{ass1} holds for the hypothesis space $\mathcal{H}_{\mathbf{q}}^{\sigma}$.
%\end{restatable}
% \end{Proposition}

%We note that, in many works \cite{}, the bias $\mathbf{b}_g=\mathbf{0}$. It is easy to check that when $\mathbf{b}_g=\mathbf{0}$ and $g>2$, Proposition \ref{P1} holds.
% Then, we introduce the fully-convolutional network (FCN)

% \subsection{Score-based Hypothesis Space}
\textbf{Score-based Hypothesis Space.}
Many OOD detection algorithms detect OOD data by using a score-based strategy. That is, given a threshold $\lambda$, a scoring function space $\mathcal{F}_{l}\subset \{\mathbf{f}:\mathcal{X}\rightarrow \mathbb{R}^l\}$ and a scoring function $E: \mathcal{F}_l\rightarrow \mathbb{R}$, then $\mathbf{x}$ is regarded as ID data if and only if $E(\mathbf{f}(\mathbf{x}))\geq \lambda$. 
We introduce several representative scoring functions $E$ as follows: for any $\mathbf{f}=[f^1,...,f^l]^{\top}\in \mathcal{F}_{l}$,\\
$\bullet$ softmax-based function \cite{Hendrycks2017abaseline} and temperature-scaled function \cite{liang2018enhancing}: $\lambda\in (\frac{1}{l},1)$ and $T>0$,
\begin{equation}\label{score1}
    E(\mathbf{f}) =\max_{k\in\{1,...,l\}}  \frac{\exp{(f^k)}}{\sum_{c=1}^l \exp{(f^c)}},~~~~~~~E(\mathbf{f}) =\max_{k\in\{1,...,l\}}  \frac{\exp{(f^k/T)}}{\sum_{c=1}^l \exp{(f^c/T)}};
\end{equation}
% $\bullet$ temperature-scaled function \cite{liang2018enhancing}
% \begin{equation}\label{score2}
%     E(\mathbf{f}) =\max_{k\in\{1,...,K\}}  \frac{\exp{(f^k/T)}}{\sum_{c=1}^K \exp{(f^c/T)}},~~~ \lambda\in (\frac{1}{K},1);
% \end{equation}
$\bullet$ energy-based function \cite{liu2020energy}: $\lambda\in (0,+\infty) $ and $T>0$,
\begin{equation}
    E(\mathbf{f}) =T\log \sum_{c=1}^l  \exp{(f^c/T)}.~~~ \label{score3}
    % \\
    % E(\mathbf{f}) =\sum_{c=1}^K \log   (1+\exp{(f^c)}),~~~  \lambda\in (0,+\infty). \label{score4}
\end{equation}
% \begin{equation}\label{score3}
% \begin{split}
%     E(\mathbf{f}) =T\log \sum_{c=1}^K  \exp{(f^c/T)},~~~ \lambda\in (0,+\infty);
%     \end{split}
% \end{equation}
% \begin{equation}\label{score4}
% \begin{split}
%     E(\mathbf{f}) =\sum_{c=1}^K \log   (1+\exp{(f^c)}),~~~  \lambda\in (0,+\infty).
%     \end{split}
% \end{equation}

Using $E$, $\lambda$ and $\mathbf{f}\in \mathcal{F}_{\mathbf{q}}^{\sigma}$, we have a classifier: $h^{\lambda}_{\mathbf{f},E}(\mathbf{x})=1$, if $E(\mathbf{f}(\mathbf{x}))\geq \lambda$; otherwise, $h^{\lambda}_{\mathbf{f},E}(\mathbf{x})=2$, where $1$ represents the ID data and $2$ represents the OOD data. Hence, a binary classification hypothesis space $\mathcal{H}^b$, which consists of all $h^{\lambda}_{\mathbf{f},E}$, is generated. We define $
    \mathcal{H}^{{\sigma},\lambda}_{{\mathbf{q}},E} := \{h^{\lambda}_{\mathbf{f},E}:\forall \mathbf{f}\in \mathcal{F}_{\mathbf{q}}^{\sigma}\}
$.
% $h^{\lambda}_{\mathbf{f},E}$:
%\begin{equation*}
 %   h^{\lambda}_{\mathbf{f},E}(\mathbf{x}) := \left \{
  %  \begin{aligned}
 %  &~~~~1,~~\textnormal{if}~E(\mathbf{f}(\mathbf{x}))\geq \lambda;\\ 
 %  &~~~~  2,~~\textnormal{if}~E(\mathbf{f}(\mathbf{x}))< \lambda,
  %  \end{aligned}
 %   \right.
%\end{equation*}
%Next, we introduce an important proposition for the space $\mathcal{H}^{{\sigma},\lambda}_{{\mathbf{q}},E}$.

% \begin{Proposition}
%\begin{restatable}{Proposition}{propTwo}
%\label{P2}
%For any sequence $\mathbf{q}=(l_1,...,l_g)$ satisfying that $l_1=d$ and $l_g=K$, if $\{\mathbf{v}\in \mathbb{R}^K:E(\mathbf{v}) \geq \lambda\}\neq \emptyset$ and  $\{\mathbf{v}\in \mathbb{R}^K:E(\mathbf{v}) < \lambda\}\neq \emptyset$, then the constant functions $h_{1}$ and $h_{2}$ belong to $\mathcal{H}^{{\sigma},\lambda}_{{\mathbf{q}},E}$, where $h_{1}:=1$ and $h_{2}:=2$. Therefore, Assumption \ref{ass1} holds for the hypothesis space $\mathcal{H}^{{\sigma},\lambda}_{{\mathbf{q}},E}$.
%\end{restatable}
% \end{Proposition}

% When 
% \subsection{Learnability of OOD Detection in Different Hypothesis Spaces}
\textbf{Learnability of OOD Detection in Different Hypothesis Spaces.}
Next, we present applications of our theory regarding the above two practical and important hypothesis spaces $\mathcal{H}_{\mathbf{q}}^{\sigma}$ and $\mathcal{H}_{\mathbf{q},E}^{\sigma,\lambda}$.

% \begin{theorem}

\begin{restatable}{theorem}{thmAppFCNN}
\label{T24}
 Suppose that Condition \ref{C3} holds and the hypothesis space $\mathcal{H}$ is FCNN-based or score-based, i.e., $\mathcal{H}=\mathcal{H}_{\mathbf{q}}^{\sigma}$ or  $\mathcal{H}=\mathcal{H}^{\rm in}\bullet \mathcal{H}^{\rm b}$, where  $\mathcal{H}^{\rm in}$ is an ID hypothesis space, $\mathcal{H}^{\rm b}=\mathcal{H}^{{\sigma},\lambda}_{{\mathbf{q}},E}$ and $\mathcal{H}=\mathcal{H}^{\rm in}\bullet \mathcal{H}^{\rm b}$ is introduced below Eq. \eqref{Eq.dot}, here
 $E$ is introduced in Eqs. \eqref{score1} or \eqref{score3}. Then
% \\
\\
$~~~~~~~~~~~~~~~~~~~~~~~~$ \fbox{
 \parbox{0.7\textwidth}{
There is a sequence $\mathbf{q}=(l_1,...,l_g)$  such that OOD detection is learnable in the separate space $\mathscr{D}^s_{XY}$ for $\mathcal{H}$ \textbf{if and only if} $|\mathcal{X}|<+\infty$.
 }}
 \\
Furthermore, if $|\mathcal{X}|<+\infty$, then there exists a sequence $\mathbf{q}=(l_1,...,l_g)$ such that for any sequence $\mathbf{q}'$ satisfying that $\mathbf{q}\lesssim\mathbf{q}'$, OOD detection is learnable in  $\mathscr{D}^s_{XY}$ for $\mathcal{H}$.
\end{restatable}

Theorem~\ref{T24} states that 1) when the hypothesis space is FCNN-based or score-based, the finite feature space is the necessary and sufficient condition for the learnability of OOD detection in the separate space; and 2) a larger architecture of FCNN has a greater probability to achieve the learnability of OOD detection in the separate space.
Note that when we select  Eqs. \eqref{score1} or \eqref{score3} as the scoring function $E$, Theorem~\ref{T24} also shows that the selected scoring functions $E$ can guarantee the learnability of OOD detection, which is a theoretical support for the representative works \cite{liang2018enhancing,liu2020energy,Hendrycks2017abaseline}. Furthermore, Theorem \ref{T24.3} also offers theoretical supports for these works in the density-based space, when $K=1$. %Next, we use Theorem~\ref{Cor2} to study the learnability in the separate space for representative works \cite{liang2018enhancing,liu2020energy,Hendrycks2017abaseline} whose scoring functions $E$ are in Eqs.~\eqref{score1} and \eqref{score3}. 

% \fbox{
% \parbox{0.98\textwidth}{
% % \begin{corollary}
% 
% \vspace{-0.6em}
% % \end{corollary}
% }
% }
% \begin{theorem}
% \end{theorem}
%\vspace{-0.15cm}

%Note that, in Theorem~\ref{T27}, we have not provided a  sufficient and necessary condition similar to Theorem \ref{T24}. The reason is that when $|\mathcal{X}|=+\infty$, we cannot ensure that $\mathcal{H}^{{\sigma},\lambda}_{{\mathbf{q}},E}$ has finite VC dimension. If $E$ is constructed by a neural network or a polynomial \cite{Peter2019Nearly}, then it is promising to get the sufficient and necessary condition. However, in representative OOD detection algorithms, the selected scoring functions $E$ are too complicated (\textit{e.g.}, Eqs.~\eqref{score1} and \eqref{score3}. Therefore, we leave studying such sufficient and necessary conditions to future work.  Lastly, we use Theorem~\ref{T27} to study several representative works whose scoring functions $E$ are shown in Eqs.~\eqref{score1} and \eqref{score3}.

\begin{restatable}{theorem}{thmAppFCNNtwo}
\label{T24.3}
 Suppose that each domain $D_{XY}$ in $\mathscr{D}_{XY}^{\mu,b}$  is attainable, i.e., $\argmin_{h\in \mathcal{H}}R_{D}(h)\neq \emptyset$ (the finite discrete domains satisfy this). Let $K=1$ and the hypothesis space $\mathcal{H}$ be score-based $($$\mathcal{H}=\mathcal{H}_{\mathbf{q},E}^{\sigma,\lambda}$, where $E$ is in Eqs. \eqref{score1} or \eqref{score3}$)$ or  FCNN-based $($$\mathcal{H}=\mathcal{H}_{\mathbf{q}}^{\sigma}$$)$.
% \\
% \fbox{
% \parbox{0.98\textwidth}{
If $\mu(\mathcal{X})<+\infty$, then the following four conditions are \textbf{equivalent}:
\\
\fbox{
 \parbox{0.98\textwidth}{
 Learnability in $\mathscr{D}_{XY}^{\mu,b}$ for  $\mathcal{H}$ $\iff$ Condition \ref{C1} 
 $\iff$
 Realizability Assumption $\iff$ Condition \ref{Con2}
}}
\end{restatable}
Theorem \ref{T24.3} still holds if the function space $\mathcal{F}_{\mathbf{q}}^{\sigma}$ is generated by Convolutional Neural Network.
\textbf{Overlap and Benefits of Multi-class Case.}
We investigate when the hypothesis space is FCNN-based or score-based, what will happen if there exists an overlap between the ID and OOD distributions?

% \begin{theorem}
\begin{restatable}{theorem}{thmImpOverlapFCNN}
\label{overlapcase}
Let $K=1$ and the hypothesis space $\mathcal{H}$ be score-based $($$\mathcal{H}=\mathcal{H}_{\mathbf{q},E}^{\sigma,\lambda}$, where $E$ is in Eqs. \eqref{score1} or \eqref{score3}$)$ or FCNN-based $($$\mathcal{H}=\mathcal{H}_{\mathbf{q}}^{\sigma}$$)$.  Given a prior-unknown space $\mathscr{D}_{XY}$, if there exists a domain $D_{XY}\in \mathscr{D}_{XY}$, which has an overlap between ID and OOD distributions (see Definition \ref{def:overlap}), then OOD detection is not learnable in the domain space $\mathscr{D}_{XY}$ for  $\mathcal{H}$.
\end{restatable}
% \end{theorem}
%\vspace{-0.1cm}

When $K=1$ and the hypothesis space is FCNN-based or score-based, Theorem \ref{overlapcase} shows that overlap between ID and OOD distributions is the sufficient condition for the unlearnability of OOD detection.
% This result in Theorem \ref{overlapcase} also holds in the score-based hypothesis space ({see Appendix {\ref{SN.2}}}).
Theorem \ref{overlapcase} takes roots in the conditions $\inf_{h\in \mathcal{H}} R_{D}^{\rm in}(h)=0$ and $\inf_{h\in \mathcal{H}} R_{D}^{\rm out}(h)=0$. However, when $K>1$, we can ensure $\inf_{h\in \mathcal{H}}R_{D}^{\rm in}(h)>0$ if ID distribution $D_{X_{\rm I}Y_{\rm I}}$ has overlap between ID classes. By this observation, we conjecture that when $K>1$, OOD detection is learnable in some special cases where overlap exists, even if the hypothesis space is FCNN-based or score-based.
% \vspace{-0.5cm}
\section{Discussion}

\textbf{Understanding Far-OOD Detection.} Many existing works \cite{Hendrycks2017abaseline,DBLP:journals/corr/abs-2204-05306} study the far-OOD detection issue. Existing benchmarks include 1) MNIST \cite{DBLP:journals/spm/Deng12} as ID dataset, and Texture \cite{kylberg2011kylberg}, CIFAR-$10$ \cite{krizhevsky2010convolutional} or Place$365$ \cite{DBLP:journals/pami/ZhouLKO018} as OOD datasets; and  2) CIFAR-$10$ \cite{krizhevsky2010convolutional} as ID dataset, and MNIST \cite{DBLP:journals/spm/Deng12}, or Fashion-MNIST \cite{DBLP:journals/pami/ZhouLKO018} as OOD datasets. 
In far-OOD case, we find that the ID and OOD datasets have different semantic labels and different styles. From the theoretical view, we can define far-OOD detection tasks as follows: for $\tau>0$, a domain space $\mathscr{D}_{XY}$ is $\tau$-far-OOD, if for any domain $D_{XY}\in \mathscr{D}_{XY}$, 
\begin{equation*}
    {\rm dist}({{\rm supp} D_{X_{\rm O}}},{{\rm supp} D_{X_{\rm I}}})>\tau.
\end{equation*}
Theorems \ref{T17}, \ref{T-SET} and \ref{T24} imply that under appropriate hypothesis space, $\tau$-far-OOD detection is learnable. In Theorem \ref{T17}, the condition $|\mathcal{X}|<+\infty$ is necessary for the separate space. However, one can prove that in the  far-OOD case, when $\mathcal{H}^{\rm in}$ is agnostic PAC learnable for ID distribution, the results in  Theorem \ref{T17} still holds, if the condition $|\mathcal{X}|<+\infty$ is replaced by a weaker condition that $\mathcal{X}$ is compact. In addition, it is notable that when $\mathcal{H}^{\rm in}$ is agnostic PAC learnable for ID distribution and $\mathcal{X}$ is compact, the KNN-based OOD
detection algorithm \cite{sun2022knn} is consistent in the  $\tau$-far-OOD case.

\textbf{Understanding Near-OOD Detection.} When the ID and OOD datasets have similar semantics or styles,  OOD detection tasks become more challenging. \cite{ren2021asimplefix,DBLP:conf/nips/FortRL21} consider this issue and name it near-OOD detection. Existing benchmarks include  1) MNIST \cite{DBLP:journals/spm/Deng12} as ID dataset, and Fashion-MNIST \cite{DBLP:journals/pami/ZhouLKO018} or Not-MNIST \cite{notmnist} as OOD datasets; and 2) CIFAR-$10$ \cite{krizhevsky2010convolutional} as ID dataset, and CIFAR-$100$ \cite{cifar} as OOD dataset. From the theoretical view, some near-OOD tasks may imply the overlap condition, \textit{i.e.} Definition \ref{def:overlap}. Therefore, Theorems \ref{T5} and \ref{overlapcase} imply that near-OOD detection may be not learnable. 
Developing a theory to understand the feasibility of near-OOD detection is still an \textit{open question}.

\textbf{Understanding One-class Novelty Detection.}  In one-class novelty detection and semantic anomaly detection (\textit{i.e.} $K=1$), Theorem \ref{T13} has revealed that it is necessary to use a large-capacity model to ensure the good generalization in the separate space. Theorem \ref{T5} and Theorem \ref{overlapcase} suggest that we should try to avoid the overlap between ID and OOD distributions in the one-class case. If the overlap cannot be avoided, we suggest considering the multi-class OOD detection instead of the one-class case.  Additionally, in the density-based space, Theorem \ref{T24.3} has shown that it is necessary to select a suitable hypothesis space satisfying the Realizability Assumption to ensure the learnability of OOD detection in the density-based space. Generally, a large-capacity model can be helpful to guarantee that the Realizability Assumption holds.

% \textbf{FPR and AUROC Metrics.} FPR (\textit{i.e.}, False Positive Rate) and AUROC (\textit{i.e.}, Area under the Receiver
% Operating Characteristic Curve) metrics are commonly used metrics in OOD detection \citep{Yang2021generalized}. TPR (\textit{i.e.}, True Positive Rate) is a corresponding metric with respect to FPR. Generally, we hope that an OOD detection algorithm can approximate the optimal TPR and FPR performances. With minor adjustment, our theory can also be used to understand OOD detection under the TPR and FPR metrics. However, due to the complexity of AUROC, we still cannot give an OOD detection theory under AUROC metric. Understanding OOD detection under AUROC metric is still an \textit{open question}.

\section{Related Work}

We briefly review the related theoretical works below. See Appendix~\ref{Asec:relatedWork} for detailed related works.

\textbf{OOD Detection Theory.}
\citep{zhang2021understanding} understands the OOD detection via goodness-of-fit tests and typical set hypothesis, and
% discover that relevant OOD distributions can lie in the high likelihood regions of a data distribution. \citep{zhang2021understanding} 
argues that minimal density estimation errors can lead to OOD detection failures without assuming an overlap between ID and OOD distributions. 
Beyond \citep{zhang2021understanding}, \citep{morteza2022provable} paves a new avenue to designing provable OOD detection algorithms.
Compared to \citep{morteza2022provable,zhang2021understanding}, our theory focuses on the PAC learnable theory of OOD detection 
% and aims to characterize the learnability of OOD detection to answer the question: Is OOD detection agnostic PAC learnable?
 and identifies several necessary and sufficient conditions for the learnability of OOD detection, opening a door to study OOD detection in theory.
% . If detectors are generated by FCNN, our theory (Theorem~\ref{overlapcase}) shows that overlap is the sufficient condition to the failure of learnability of OOD detection, which is a complementary to \citep{zhang2021understanding}. In addition,
% we identify several necessary and sufficient conditions for the learnability of OOD detection, which opens a door to studying OOD detection in theory.
% Compared to \citep{morteza2022provable}, our paper aims to characterize the learnability of OOD detection to answer the question: Is OOD detection agnostic PAC learnable?

% \textbf{Supervised Learning Theory.}

\textbf{Open-set Learning Theory.} \citep{liu2018open} and \cite{Fang2020Open,Luo2020progressive} propose the agnostic PAC learning bounds for open-set detection and open-set domain adaptation, respectively. Unfortunately, \citep{Fang2020Open,liu2018open,Luo2020progressive} all require that the test data are indispensable during the training process.  To investigate open-set learning (OSL) \textit{without accessing the test data} during training, \citep{Fang2021learning} proposes and investigates the \textit{almost} {agnostic} PAC learnability for OSL. However, the assumptions used in \citep{Fang2021learning}  are very strong and unpractical.

\textbf{Learning Theory for Classification with Reject Option.} Many works \cite{DBLP:journals/tit/Chow70,DBLP:journals/corr/abs-2101-12523} also investigate the \emph{classification with reject option} (CwRO) problem, which is similar to OOD detection in some cases.  \cite{cortes2016learning,DBLP:conf/nips/CortesDM16,DBLP:conf/nips/NiCHS19,DBLP:conf/icml/CharoenphakdeeC21,DBLP:journals/jmlr/BartlettW08} study the learning theory and propose the PAC learning bounds for CwRO. However, compared to our work regarding OOD detection, existing CwRO theories mainly focus on how the ID risk $R^{\rm in}_D$ (\textit{i.e.}, the risk that ID data is wrongly classified) is influenced by special rejection rules. Our theory not only focuses on the ID risk, but also pays attention to the OOD risk.

\textbf{Robust Statistics.} In the field of robust statistics \cite{rousseeuw2011robust}, researchers aim to propose estimators and testers that can mitigate the negative effects of outliers (similar to OOD data). The proposed estimators are supposed to be independent of the potentially high dimensionality of the data \cite{ronchetti2009robust,diakonikolas2020outlier,diakonikolas2021outlier}. Existing works \cite{diakonikolas2021outlier_robust,cheng2021outlier,diakonikolas2022outlier} in the field have identified and resolved the statistical limits of outlier robust statistics by constructing estimators and proving impossibility results. In the future, it is a promising and interesting research direction to study the robustness of OOD detection based on robust statistics.

\textbf{PQ Learning Theory.} Under some conditions, PQ learning theory \cite{DBLP:conf/nips/GoldwasserKKM20,DBLP:conf/alt/KalaiK21}  can be regarded
as the PAC theory for OOD detection in the semi-supervised
or transductive learning cases, \textit{i.e.}, test data are required during training. Besides, \cite{DBLP:conf/nips/GoldwasserKKM20,DBLP:conf/alt/KalaiK21} aim to give the PAC estimation under Realizability Assumption \cite{shalev2014understanding}. Our theory does not only study the PAC estimation in the realization cases, but also studies the other
cases, which are more difficult than PAC theory under Realizability Assumption.

\section{Conclusions and Future Works}

Detecting OOD data has shown its significance in improving the reliability of machine learning. 
However, very few works discuss OOD detection in theory, which hinders real-world applications of OOD detection algorithms. 
In this paper, we are the \emph{first} to provide the PAC theory for OOD detection. Our results imply that we cannot expect a universally consistent algorithm to handle all scenarios in OOD detection. 
Yet, it is still possible to make OOD detection learnable in certain scenarios.
For example, when we design OOD detection algorithms, we normally only have finite ID datasets. In this real scenario, Theorem \ref{T-SET} provides a necessary and sufficient condition for the success of OOD detection.
Our theory reveals many necessary and sufficient conditions for the learnability of OOD detection, hence \emph{opening a door} to studying the learnability of OOD detection. In the future,  we will focus on studying the robustness of OOD detection based on robust statistics  \cite{diakonikolas2021outlier_robust,DBLP:journals/corr/abs-1911-05911}.
% In addition, our theory also applies to several representative OOD detection works, providing strong theoretical supports for their practical successes.
\section*{Acknowledgment}
JL and ZF were supported by the Australian Research Council (ARC) under
FL190100149. YL is supported by the AFOSR Young Investigator Program Award. BH
was supported by the RGC Early Career Scheme No. 22200720 and NSFC Young Scientists Fund
No. 62006202. ZF would also like to thank Prof. Peter Bartlett and Dr. Tongliang Liu for productive discussions.

\bibliography{ood_theory}
\bibliographystyle{unsrtnat}

%\section*{Checklist}

% %%% BEGIN INSTRUCTIONS %%%
% The checklist follows the references.  Please
% read the checklist guidelines carefully for information on how to answer these
% questions.  For each question, change the default \answerTODO{} to \answerYes{},
% \answerNo{}, or \answerNA{}.  You are strongly encouraged to include a {\bf
% justification to your answer}, either by referencing the appropriate section of
% your paper or providing a brief inline description.  For example:
% \begin{itemize}
%   \item Did you include the license to the code and datasets? \answerNA{}
%   \item Did you include the license to the code and datasets? \answerNA{}
%   \item Did you include the license to the code and datasets? \answerNA{}
% \end{itemize}
% Please do not modify the questions and only use the provided macros for your
% answers.  Note that the Checklist section does not count towards the page
% limit.  In your paper, please delete this instructions block and only keep the
% Checklist section heading above along with the questions/answers below.
% %%% END INSTRUCTIONS %%%

\section*{Checklist}

\begin{enumerate}

\item For all authors...
\begin{enumerate}
  \item Do the main claims made in the abstract and introduction accurately reflect the paper's contributions and scope?
    \answerYes{}
  \item Did you describe the limitations of your work?
    \answerYes{See Appendix~\ref{Scheck}}
  \item Did you discuss any potential negative societal impacts of your work?
    \answerYes{See Appendix~\ref{Scheck}}
  \item Have you read the ethics review guidelines and ensured that your paper conforms to them?
    \answerYes{}
\end{enumerate}

\item If you are including theoretical results...
\begin{enumerate}
  \item Did you state the full set of assumptions of all theoretical results?
    \answerYes{}
        \item Did you include complete proofs of all theoretical results?
    \answerYes{}
\end{enumerate}

\item If you ran experiments...
\begin{enumerate}
  \item Did you include the code, data, and instructions needed to reproduce the main experimental results (either in the supplemental material or as a URL)?
    \answerNA{}
  \item Did you specify all the training details (e.g., data splits, hyperparameters, how they were chosen)?
    \answerNA{}
        \item Did you report error bars (e.g., with respect to the random seed after running experiments multiple times)?
    \answerNA{}
        \item Did you include the total amount of compute and the type of resources used (e.g., type of GPUs, internal cluster, or cloud provider)?
    \answerNA{}
\end{enumerate}

\item If you are using existing assets (e.g., code, data, models) or curating/releasing new assets...
\begin{enumerate}
  \item If your work uses existing assets, did you cite the creators?
    \answerNA{}
  \item Did you mention the license of the assets?
    \answerNA{}
  \item Did you include any new assets either in the supplemental material or as a URL?
    \answerNA{}
  \item Did you discuss whether and how consent was obtained from people whose data you're using/curating?
    \answerNA{}
  \item Did you discuss whether the data you are using/curating contains personally identifiable information or offensive content?
    \answerNA{}
\end{enumerate}

\item If you used crowdsourcing or conducted research with human subjects...
\begin{enumerate}
  \item Did you include the full text of instructions given to participants and screenshots, if applicable?
    \answerNA{}
  \item Did you describe any potential participant risks, with links to Institutional Review Board (IRB) approvals, if applicable?
    \answerNA{}
  \item Did you include the estimated hourly wage paid to participants and the total amount spent on participant compensation?
    \answerNA{}
\end{enumerate}

\end{enumerate}

%%%%%%%%%%%%%%%%%%%%%%%%%%%%%%%%%%%%%%%%%%%%%%%%%%%%%%%%%%%%

%%%%%%%%%%%%%%%%%%%%%%%%%%%%%%%%%%%%%%%%%%%%%%%%%%%%%%%%%%%%

\clearpage

\appendix

%\tableofcontents
\DoToC
\newpage
\section{Detailed Related Work}\label{Asec:relatedWork}

\textbf{OOD Detection Algorithms.} 
We will briefly review many representative OOD detection algorithms in three categories.
% OOD detection algorithms mainly include three categories: density-based methods \cite{morningstar2021density}, distance-based methods \cite{lee2018asimple} and classification-based methods \cite{Hendrycks2017abaseline,liang2018enhancing,liu2020energy}. 
1) Classification-based methods use an ID classifier to detect OOD data \cite{Hendrycks2017abaseline}\footnote{Note that, some methods assume that OOD data are available in advance \cite{Hendrycks2019deep,Dhamija2018reducing}. 
% These methods have much better performance than other methods due to the appearance of OOD data. 
However, the exposure of OOD data is a strong assumption \cite{Yang2021generalized}. We do not consider this situation in our paper.}. Representative works consider using the maximum softmax score \cite{Hendrycks2017abaseline}, temperature-scaled score \cite{ren2019likelihood} and energy-based score \cite{liu2020energy,Wang2021can} to identify OOD data. 
2) Density-based methods aim to estimate an ID distribution and identify the low-density area as OOD data \cite{zong2018deep}. 3) The recent development of generative models provides promising ways to make them successful in OOD detection \cite{Pidhorskyi2018generative,Nalisnick2019do,ren2019likelihood,Kingma2018glow,Xiao2020likelihood}.
Distance-based methods are based on the assumption that OOD data should be relatively far away from the centroids of ID classes \cite{lee2018asimple}, including Mahalanobis distance \cite{lee2018asimple,ren2021asimplefix}, cosine similarity \cite{zaeemzadeh2021out}, and kernel similarity \cite{Amersfoort2020uncertainty}.

Early works consider using the maximum softmax score to express the ID-ness \cite{Hendrycks2017abaseline}. Then, temperature scaling functions are used to amplify the separation between the ID and OOD data \cite{ren2019likelihood}. Recently, researchers propose hyperparameter-free energy scores to improve the OOD uncertainty estimation \cite{liu2020energy,Wang2021can}. 
Additionally, researchers also consider using the information contained in gradients to help improve the performance of OOD detection \cite{Huang2021on}.

Except for the above algorithms, researchers also study the situation, where auxiliary OOD data can be obtained during the training process \cite{Hendrycks2019deep,Dhamija2018reducing}. These methods are called outlier exposure, and have much better performance than the above methods due to the appearance of OOD data. However, the exposure of OOD data is a strong assumption \cite{Yang2021generalized}. Thus, researchers also consider generating OOD data to help the separation of OOD and ID data \cite{Vernekar2019out}. In this paper, we do not make an assumption that OOD data are available during training, since this assumption may not hold in real world.

\textbf{OOD Detection Theory.}
\citep{zhang2021understanding} rejects the typical set hypothesis, the claim that relevant OOD distributions can lie in high likelihood regions of data distribution, as implausible.
% understands the OOD detection via goodness-of-fit tests and typical set hypothesis, and discovers that relevant OOD distributions can lie in the high likelihood regions of a data distribution.
\citep{zhang2021understanding} argues that minimal density estimation errors can lead to OOD detection failures {without assuming} an overlap between ID and OOD distributions. Compared to \citep{zhang2021understanding}, our theory focuses on the PAC learnable theory of OOD detection. If detectors are generated by FCNN, our theory (Theorem~\ref{overlapcase}) shows that the overlap is the sufficient condition to the failure of learnability of OOD detection, which is complementary to \citep{zhang2021understanding}. In addition, we identify several necessary and sufficient conditions for the learnability of OOD detection, which opens a door to studying OOD detection in theory.
Beyond \citep{zhang2021understanding}, \citep{morteza2022provable} paves a new avenue to designing provable OOD detection algorithms. Compared to \citep{morteza2022provable}, our paper aims to characterize the learnability of OOD detection to answer the question: is OOD detection PAC learnable?

% \textbf{Supervised Learning Theory.}

\textbf{Open-set Learning Theory.} \citep{liu2018open} is the first to propose the agnostic PAC guarantees for open-set detection. Unfortunately, the test data must be used during the training process. \citep{Fang2020Open} considers the open-set domain adaptation (OSDA)  \cite{Luo2020progressive} and proposes the first learning bound for OSDA. \citep{Fang2020Open} mainly depends on the positive-unlabeled learning techniques \cite{kiryo2017positive,Ishida2018binary,chen2021large}. However, similar to \citep{liu2018open}, the test data must be available during training. To study open-set learning (OSL) \textit{without accessing the test data} during training, \citep{Fang2021learning} proposes and studies the {almost} PAC learnability for OSL, which is motivated by transfer learning \cite{What_Transferred_Dong_CVPR2020,9695325}. In our paper, we study the PAC learnability for OOD detection, which is an {open problem} proposed by \citep{Fang2021learning}.

\textbf{Learning Theory for Classification with Reject Option.} Many works \cite{DBLP:journals/tit/Chow70,DBLP:journals/corr/abs-2101-12523} also investigate the \emph{classification with reject option} (CwRO) problem, which is similar to OOD detection in some cases.  \cite{cortes2016learning,DBLP:conf/nips/CortesDM16,DBLP:conf/nips/NiCHS19,DBLP:conf/icml/CharoenphakdeeC21,DBLP:journals/jmlr/BartlettW08} study the learning theory and propose the agnostic PAC learning bounds for CwRO. However, compared to our work regarding OOD detection, existing CwRO theories mainly focus on how the ID risk (\textit{i.e.}, the risk that ID data is wrongly classified) is influenced by special rejection rules. Our theory not only focuses on the ID risk, but also pays attention to the OOD risk.

\textbf{Robust Statistics.} In the field of robust statistics \cite{rousseeuw2011robust}, researchers aim to propose estimators and testers that can mitigate the negative effects of outliers (similar to OOD data). The proposed estimators are supposed to be independent of the potentially high dimensionality of the data \cite{ronchetti2009robust,diakonikolas2020outlier,diakonikolas2021outlier}. Existing works \cite{diakonikolas2021outlier_robust,cheng2021outlier,diakonikolas2022outlier} in the field have identified and resolved the statistical limits of outlier robust statistics by constructing estimators and proving impossibility results. In the future, it is a promising and interesting research direction to study the robustness of OOD detection based on robust statistics.

\textbf{PQ Learning Theory.} Under some conditions, PQ learning theory \cite{DBLP:conf/nips/GoldwasserKKM20,DBLP:conf/alt/KalaiK21}  can be regarded
as the PAC theory for OOD detection in the semi-supervised
or transductive learning cases, \textit{i.e.}, test data are required during the training process. Additionally,  PQ learning theory in \cite{DBLP:conf/nips/GoldwasserKKM20,DBLP:conf/alt/KalaiK21} aims to give the PAC estimation under Realizability Assumption \cite{shalev2014understanding}. Our theory focuses on the
PAC theory in different cases, which is more difficult and more practical than PAC theory under Realizability Assumption.  

% \vspace{-0.4em}

\section{Limitations and Potential Negative Societal Impacts}\label{Scheck}

\textbf{Limitations.} The main limitation of our work lies in that we do not answer the most general question:

\textit{Given any hypothesis space $\mathcal{H}$ and  space $\mathscr{D}_{XY}$, what is the necessary and sufficient condition to ensure the PAC learnability of OOD detection?}

However, this question is still difficult to be addressed, due to 
limited mathematical skills. Yet, based on our observations and the main results in our paper, we believe the following result may hold:

\textbf{Conjecture}: \textit{If $\mathcal{H}$ is agnostic learnable for supervised learning, then OOD detection is learnable in $\mathscr{D}_{XY}$ \textbf{if and only if} compatibility condition (i.e., Condition \ref{Con2}) holds.}

We leave this question as a future work.

\textbf{Potential Negative Societal Impacts.} Since our paper is a theoretical paper and the OOD detection problem is significant to ensure the safety of deploying existing machine learning algorithms, there are no potential negative societal impacts in our paper.

\section{Discussions and Details about Experiments in Figure \ref{fig:only}}\label{SA}
In this section, we summarize our main results, then give the details of the experiments in Figure \ref{fig:only}.
\subsection{Summary}
We summarize our main results as follows:

$\bullet$ A necessary condition (\textit{i.e.}, Condition \ref{C1}) for the learnability of OOD detection is proposed. Theorem \ref{T3} shows that Condition \ref{C1} is the \emph{necessary and sufficient condition} for the learnability of OOD detection, when the domain space is the single-distribution space $\mathscr{D}_{XY}^{D_{XY}}$. This implies the Condition \ref{C1} is the necessary condition for the learnability of OOD detection.
\vspace{0.1cm}

$\bullet$ Theorem \ref{T5} has shown that the overlap between ID and OOD data can lead the failures of OOD detection under some mild assumptions. Furthermore, Theorem \ref{overlapcase} shows that when $K=1$, the overlap is the sufficient condition for the failures of OOD detection, when the hypothesis space is FCNN-based or score-based.
\vspace{0.1cm}

$\bullet$ Theorem \ref{T4} provides an impossibility theorem for the total space $\mathscr{D}_{XY}^{\rm all}$. OOD detection  is not learnable in $\mathscr{D}_{XY}^{\rm all}$ for any non-trivial hypothesis space.
\vspace{0.1cm}

$\bullet$ Theorem \ref{T12} gives impossibility theorems for the separate space $\mathscr{D}_{XY}^{s}$. To ensure the impossibility theorems hold, mild assumptions are required. Theorem \ref{T12} also implies that OOD detection may be learnable in the separate space $\mathscr{D}_{XY}^{s}$, if the feature space is finite, \emph{i.e.}, $|\mathcal{X}|<+\infty$. Additionally, Theorem \ref{T24} implies that the finite feature space may be the necessary condition to ensure the learnability of OOD detection in the separate space.
\vspace{0.1cm}

$\bullet$ When $|\mathcal{X}|<+\infty$ and $K=1$, Theorem \ref{T13} provides the  \emph{necessary and sufficient condition} for the learnability of OOD detection in the separate space $\mathscr{D}_{XY}^{s}$. Theorem \ref{T13} implies that if the OOD detection can be learnable in the distribution-agnostic case, then a large-capacity model is necessary.
 Based on Theorem \ref{T13}, Theorem \ref{T17} studies the learnability in the $K>1$ case. 
\vspace{0.1cm}

% Then, Theorems \ref{T16} and \ref{T17.1} provide the positive answers for the learnability of OOD detection in $\tau$-margin space $\mathscr{D}_{XY}^{\tau}$.
% \\
$\bullet$ The compatibility condition (\textit{i.e.}, Condition \ref{Con2}) for the learnability of OOD detection is proposed. Theorem \ref{T-SET} shows that Condition \ref{Con2} is the \emph{necessary and sufficient condition} for the learnability of OOD detection in the finite-ID-distribution space $\mathscr{D}_{XY}^{F}$. This also implies Condition \ref{Con2} is the necessary condition for any prior-unknown space. Note that we can only collect finite ID datasets to build models. Hence, Theorem \ref{T-SET} can handle the most practical scenarios.
\vspace{0.1cm}

$\bullet$ To further understand the importance of the compatibility condition (Condition \ref{Con2}). Theorem \ref{T-SET2} considers the density-based space $\mathscr{D}_{XY}^{\mu,b}$. We discover that Realizability Assumption implies the  compatibility condition in the density-based space. Based on this observation, we prove that OOD detection is learnable in $\mathscr{D}_{XY}^{\mu,b}$ under Realizability Assumption.
\vspace{0.1cm}

$\bullet$ Theorem \ref{T24} gives practical applications of our theory. In this theorem, we discover that the finite feature space is a \emph{necessary and sufficient condition} for the learnability of OOD detection in  the separate space $\mathscr{D}_{XY}^{s}$, when the hypothesis space is FCNN-based or score-based. 
\vspace{0.1cm}

 $\bullet$ Theorem \ref{T24.3} has shown that when $K=1$ and the hypothesis space is FCNN-based or score-based, Realizability Assumption, Condition \ref{Con2}, Condition \ref{C1} and the learnability of OOD detection in the density-based space $\mathcal{D}_{XY}^{\mu,b}$ are all  \textit{equivalent}.
\vspace{0.1cm}

$\bullet$ \textbf{Meaning of Our Theory.}
In classical statistical learning theory, the generalization theory guarantees that a well-trained classifier can be generalized well on the test set as long as the training and test sets are from the same distribution \cite{shalev2014understanding,mohri2018foundations}. 
However, since the OOD data are unseen during the training process, it is very difficult to determine whether the generalization theory holds for OOD detection.

Normally, OOD data are unseen and can be various. We hope that there exists an algorithm that can be used for the various OOD data instead of some certain OOD data, which is the reason why the generalization theory for OOD detection needs to be developed.
In this paper, we investigate the generalization theory regarding OOD detection and point out when the OOD detection can be successful. 
Our theory is based on the PAC learning theory. The impossibility theorems and the given necessary and sufficient conditions outlined provide important perspectives from which to think about OOD detection. 

\subsection{Details of Experiments in Figure~\ref{fig:only}}
\label{Asec:detail_of_fig}

In this subsection, we present details of the experiments in Figure~\ref{fig:only}, including data generation, configuration and OOD detection procedure.

\begin{figure*}[!t]
    \centering
    \scriptsize
    \subfigure[ID and OOD Distributions]{
    \raisebox{0.1\height}{\includegraphics[width=0.5\linewidth]{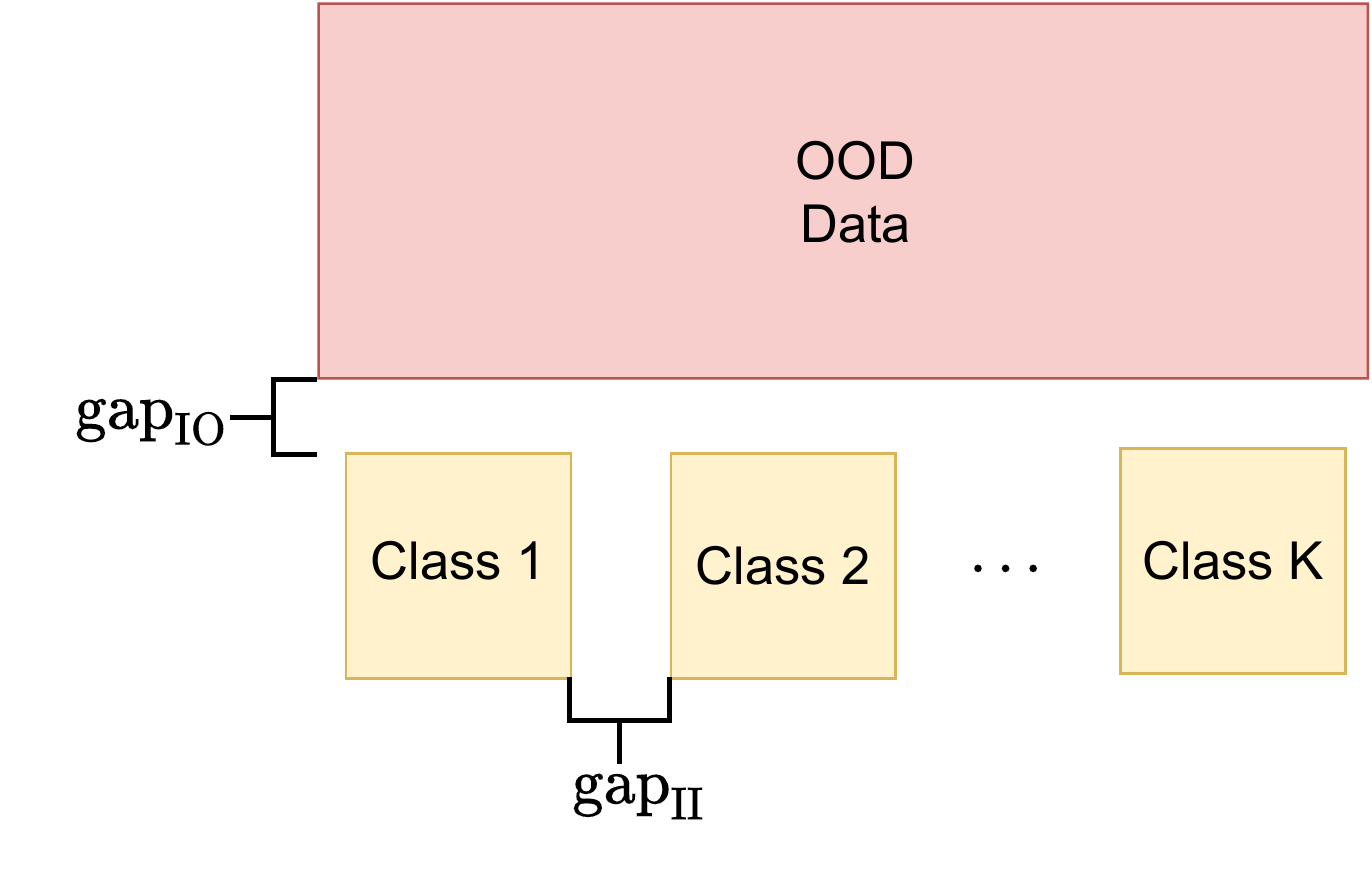}}~~~~~~~~~~~~~~~~~~~~~~~~~~~}\\
    \subfigure[Illustration of ID and OOD Data]{
    \raisebox{0.1\height}{\includegraphics[width=1\linewidth]{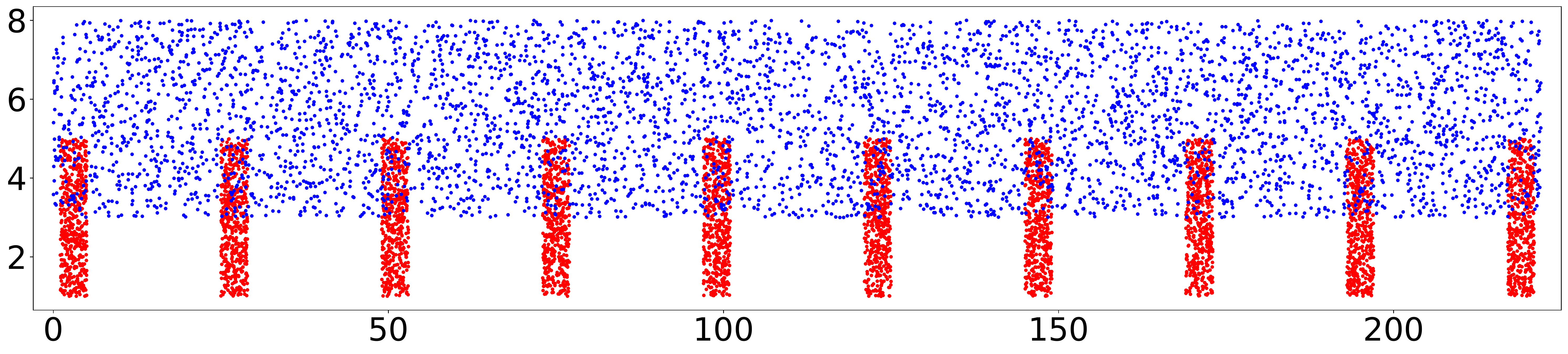}} }
    \caption{ID and OOD distributions in Figure~\ref{fig:only}.}\label{fig2}
\end{figure*}

\textbf{Data Generation.} ID and OOD data are drawn from the following \emph{uniform} (U) distributions (note that we use ${\rm U}(\mathbf{I})$ to present the uniform distribution in region $\mathbf{I}$).
\\
$\bullet$ The marginal distribution of ID distribution for class $c$: for any $c\in \{1,...,10\}$,
\begin{align}
D_{X_{\rm I}|Y_{\rm I}=c} = {\rm U}(\mathbf{I}_c),~{\rm where}~\mathbf{I}_c=[d_c,d_c+4]\times [1,5],% ,
\end{align}
here $d_i=5+{\rm gap}_{\rm II}*(i-1) + 4(i-2)$ and ${\rm gap}_{\rm II}$ is a positive constant.
\\
$\bullet$ The class-prior probability for class $c$: for any $c\in \{1,...,10\}$,
\begin{equation*}
    D_{Y_{\rm I}}(y=c)= \frac{1-\alpha}{10}.
\end{equation*}
\\
$\bullet$ The marginal distribution of OOD distribution:
\begin{align}
D_{X_{\rm O}} = {\rm U}(\mathbf{I}_{\rm out}),~{\rm where}~ \mathbf{I}_{\rm out}=[d_1-1,d_{10}+5]\times [5+{\rm gap}_{\rm IO},10+{\rm gap}_{\rm IO}].
\end{align}
Figure \ref{fig2} shows the OOD and ID distributions, when ${\rm gap}_{\rm II}=20$ and ${\rm gap}_{\rm IO}=-2$. In Figure~\ref{fig:only}, we draw $n$ data from ID distribution ($n=15,000, 20,000, 25,000$) and $25,000$ data from the OOD distribution.

\textbf{Configuration.} The architecture of ID classifier is a four-layer FCNN. The number of neurons in hidden layers is set to $100$, and the number of neurons of output layer is set to $10$. These neurons use sigmoid activations. We use the Adam optimizer \citep{Kingma2015adam} to optimize the network's parameters (with the $\ell_2$ loss). The learning rate is set to $0.001$, and the max number of training iterations is set to $10,000$. Within each iteration, we use full batch to update the network's parameters. ${\rm gap}_{\rm II}$ is set to $20$ in our experiments. In Figure~\ref{fig:only}b, ${\rm gap}_{\rm IO}=-2$ (the overlap exists, see Figure~\ref{fig2}), and in Figure~\ref{fig:only}c, ${\rm gap}_{\rm IO}=100$ (no overlap).

\textbf{OOD Detection Procedure.} We first train an ID classifier with $n$ data drawn from the ID distribution. Then, according to \citep{liu2020energy}, we apply the free-energy score to identify the OOD data and calculate the $\alpha$-risk (with the $0$-$1$ loss). We repeat the above detection procedure $20$ times and report the average $\alpha$-risk in Figure~\ref{fig:only}. Note that, following \citep{liu2020energy}, we choose the threshold used by the free-energy method so that $95\%$ of ID data are correctly identified as the ID classes by the OOD detector.

\newpage
\section{Notations}\label{SB}
\subsection{Main Notations and Their Descriptions}\label{SB.1}
In this section, we summarize important notations in Table \ref{Table_conncept}.
\begin{table}[h]
\caption{{Main notations and their descriptions.}}
\vspace{-1em}
\begin{center}\label{Table_conncept}
\footnotesize
\begin{tabular}{p{5cm}p{8.2cm}}
\hline
Notation & ~~~~~~~~~~Description  \\ \hline
$\bullet$ \textbf{Spaces and Labels} & \\
$d$ and $\mathcal{X}\subset \mathbb{R}^d$&  the feature dimension of data point and feature space\\
$\mathcal{Y}$&  ID label space $\{1,...,K\}$\\
$K+1$&  $K+1$ represents the OOD labels\\
$\mathcal{Y}_{\rm all}$&  $\mathcal{Y}\cup \{K+1\}$\\
$\bullet$ \textbf{Distributions} & \\
$X_{\rm I}$, $X_{\rm O}$, $Y_{\rm I}$, $Y_{\rm O}$ & ID feature, OOD feature, ID label, OOD label random variables\\
$D_{X_{\rm I}Y_{\rm I}}$, $D_{X_{\rm O}Y_{\rm O}}$&ID joint distribution and OOD joint distribution\\
$D_{XY}^{\alpha}$& $D_{XY}^{\alpha}=(1-\alpha)D_{X_{\rm I}Y_{\rm I}}+\alpha D_{X_{\rm O}Y_{\rm O}},~~\forall \alpha\in [0,1]$\\
$\pi^{\rm out}$& class-prior probability for OOD distribution\\
$D_{XY}$& $D_{XY}=(1-\pi^{\rm out})D_{X_{\rm I}Y_{\rm I}}+\pi^{\rm out} D_{X_{\rm O}Y_{\rm O}}$, called domain
\\
$D_{X_{\rm I}}, D_{X_{\rm O}},D_X$& marginal distributions for $D_{X_{\rm I}Y_{\rm I}}$, $D_{X_{\rm O}Y_{\rm O}}$ and $D_{XY}$, respectively
\\
$\bullet$ \textbf{Domain Spaces} & \\
$\mathscr{D}_{XY}$& domain space consisting of some domains\\
$\mathscr{D}_{XY}^{\rm all}$& total space\\
$\mathscr{D}_{XY}^{s}$& seperate space\\
$\mathscr{D}_{XY}^{D_{XY}}$& single-distribution space\\ 
$\mathscr{D}_{XY}^{F}$& finite-ID-distribution space\\
$\mathscr{D}_{XY}^{\mu,b}$& density-based space\\
$\bullet$ \textbf{Loss Function, Function Spaces} & \\
$\ell(\cdot,\cdot)$& loss: $\mathcal{Y}_{\rm all}\times \mathcal{Y}_{\rm all}\rightarrow \mathbb{R}_{\geq 0}$:  $\ell(y_1,y_2)=0$ if and only if $y_1=y_2$\\
$\mathcal{H}$& hypothesis space 
\\ 
$\mathcal{H}^{\rm in}$& ID hypothesis space 
\\ 
$\mathcal{H}^{\rm b}$& hypothesis space in binary
classification 
\\ 
$\mathcal{F}_l$& scoring function space consisting some $l$ dimensional vector-valued functions
\\ 
$\bullet$ \textbf{Risks and Partial Risks} & \\
$R_D(h)$& risk corresponding to $D_{XY}$\\
$R_D^{\rm in}(h)$& partial risk corresponding to $D_{X_{\rm I}Y_{\rm I}}$\\
$R_D^{\rm out}(h)$& partial risk corresponding to $D_{X_{\rm O}Y_{\rm O}}$\\
$R_D^{\alpha}(h)$& $\alpha$-risk corresponding to $D_{XY}^{\alpha}$
\\
$\bullet$ \textbf{Fully-Connected Neural Networks} & \\
$\mathbf{q}$&  a sequence $(l_1,...,l_g)$ to represent the architecture of FCNN\\
$\sigma$&  activation function. In this paper, we use ReLU function\\
$\mathcal{F}_{\mathbf{q}}^{\sigma}$&  FCNN-based scoring function space\\
$\mathcal{H}_{\mathbf{q}}^{\sigma}$&  FCNN-based hypothesis space\\
$\mathbf{f}_{\mathbf{w},\mathbf{b}}$& FCNN-based scoring function, which is from $\mathcal{F}_{\mathbf{q}}^{\sigma}$\\
${h}_{\mathbf{w},\mathbf{b}}$& FCNN-based hypothesis function, which is from $\mathcal{H}_{\mathbf{q}}^{\sigma}$\\
$\bullet$ \textbf{Score-based Hypothesis Space} & \\
$E$ & scoring function\\
$\lambda$& threshold\\
$\mathcal{H}_{\mathbf{q},E}^{\sigma, \lambda}$&  score-based hypothesis
space---a binary classification space\\
${h}_{\mathbf{f},E}^{\lambda}$&  score-based hypothesis
function---a binary classifier\\
\hline
\end{tabular}
\end{center}
\end{table}
\\
Given $\mathbf{f}=[f^1,...,f^l]^{\top}$, for any $\mathbf{x}\in \mathcal{X}$,
\begin{equation*}
    \argmax_{k
    \in \{1,...,l\}} f^k(\mathbf{x}):=\max \{k\in \{1,...,l\}: f^k(\mathbf{x}) \geq f^i(\mathbf{x}), \forall i=1,...,l\},
\end{equation*}
where $f^k$ is the $k$-th coordinate of $\mathbf{f}$ and $f^i$ is the $i$-th coordinate of $\mathbf{f}$. 
The above definition about $\argmax$ aims to overcome some special cases. For example, there exist ${k}_1$, ${k}_2$ ($k_1< k_2$) such that $f^{k_1}(\mathbf{x})=f^{k_2}(\mathbf{x})$ and $f^{k_1}(\mathbf{x})> f^{i}(\mathbf{x})$, $f^{k_2}(\mathbf{x})> f^{i}(\mathbf{x})$, $\forall i\in \{1,...,l\}{- }\{k_1,k_2\}$. 
Then, according to the above definition, $k_2=\argmax_{k\in\{1,...,l\}} f^k(\mathbf{x})$.
\subsection{Realizability Assumption}\label{SB.2}
\begin{assumption}[Realizability Assumption]\label{Reaass}
A domain space $\mathscr{D}_{XY}$ and hypothesis space $\mathcal{H}$ satisfy the Realizability Assumption, if for each domain $D_{XY}\in \mathscr{D}_{XY}$, there exists at least one hypothesis function $h^{*}\in \mathcal{H}$ such that $R_D(h^*)=0$.
\end{assumption}
\subsection{Learnability and PAC learnability}\label{SB.3}
Here we give a proof to show that Learnability given in Definition \ref{D0} and PAC learnability are equivalent.
\\

\textbf{First}, we prove that Learnability concludes the PAC learnability.
\\

According to Definition \ref{D0},
\begin{equation*}
    \mathbb{E}_{S\sim D^n_{X_{\rm I}Y_{\rm I}}} R_{D}(\mathbf{A}(S))\leq \inf_{h\in \mathcal{H}}R_{D}(h) + \epsilon_{\rm cons}(n),
\end{equation*}
which implies that
\begin{equation*}
    \mathbb{E}_{S\sim D^n_{X_{\rm I}Y_{\rm I}}} [R_{D}(\mathbf{A}(S))-\inf_{h\in \mathcal{H}}R_{D}(h)]\leq   \epsilon_{\rm cons}(n).
\end{equation*}
Note that $R_{D}(\mathbf{A}(S))-\inf_{h\in \mathcal{H}}R_{D}(h) \geq 0$. Therefore, by Markov's inequality, we have \begin{equation*}
    \mathbb{P}(R_{D}(\mathbf{A}(S))-\inf_{h\in \mathcal{H}}R_{D}(h)<\epsilon)>1-  \mathbb{E}_{S\sim D^n_{X_{\rm I}Y_{\rm I}}} [R_{D}(\mathbf{A}(S))-\inf_{h\in \mathcal{H}}R_{D}(h)]/\epsilon \geq 1- \epsilon_{\rm cons}(n)/\epsilon.
\end{equation*}
 Because $\epsilon_{\rm cons}(n)$ is monotonically decreasing, we can find a smallest $m$ such that $\epsilon_{\rm cons}(m) \geq \epsilon\delta$ and $\epsilon_{\rm cons}(m-1) < \epsilon\delta$, for $\delta\in (0,1)$. We define that $m(\epsilon,\delta)=m$. Therefore, for any $\epsilon>0$ and $\delta\in (0,1)$, there exists a function $m(\epsilon,\delta)$ such that when $n>m(\epsilon,\delta)$, with the probability at least $1-\delta$, we have
 \begin{equation*}
     R_{D}(\mathbf{A}(S))-\inf_{h\in \mathcal{H}}R_{D}(h)<\epsilon,
 \end{equation*}
 which is the definition of  PAC learnability.
 \\

\textbf{Second}, we prove that the PAC learnability concludes Learnability.
\\

 PAC-learnability: for any $\epsilon>0$ and $0<\delta<1$, there exists a function $m(\epsilon,\delta)>0$ such that when the sample size $n>m(\epsilon,\delta)$, we have that with the probability at least $1-\delta>0$,
\begin{equation*}
    R_D(\mathbf{A}(S))-\inf_{h\in \mathcal{H}}  R_D(h) \leq \epsilon.
\end{equation*}

Note that the loss $\ell$ defined in Section \ref{S3} has upper bound (because $\mathcal{Y}\cup \{K+1\}$ is a finite set). We assume the upper bound of $\ell$ is $M$. Hence, according to the definition of PAC-learnability, when the sample size $n>m(\epsilon,\delta)$, we have that 
\begin{equation*}
 \mathbb{E}_S [R_D(\mathbf{A}(S))-\inf_{h\in \mathcal{H}}  R_D(h)] \leq \epsilon(1-\delta)+2M\delta < \epsilon+2M\delta.
\end{equation*}
If we set $\delta = \epsilon$, then when the sample size $n>m(\epsilon,\epsilon)$, we have that 
\begin{equation*}
 \mathbb{E}_S [R_D(\mathbf{A}(S))-\inf_{h\in \mathcal{H}}  R_D(h)] < (2M+1)\epsilon,
\end{equation*}
this implies that
\begin{equation*}
 \lim_{n\rightarrow +\infty}\mathbb{E}_S [R_D(\mathbf{A}(S))-\inf_{h\in \mathcal{H}}  R_D(h)]=0,
\end{equation*}
which implies the Learnability in Definition \ref{D0}. We have completed this proof.

\subsection{Explanations for Some Notations in Section \ref{S3}}

First, we explain the concept that $S\sim {D}_{X_{I}Y_{I}}^n$ in Eq. \eqref{issue-definition1}.
\\

$S=\{(\mathbf{x}^1,{y}^1),...,(\mathbf{x}^n,{y}^n)\}$ is training data drawn {independent and identically distributed}  from $D_{X_{\rm I}Y_{\rm I}}$.
\\

$D_{X_{\rm I}Y_{\rm I}}^n$ denotes the probability over $n$-tuples induced
by applying $D_{X_{\rm I}Y_{\rm I}}$ to pick each element of the tuple independently of the other
members of the tuple.
\\

Because these samples are i.i.d. drawn $n$ times, researchers often use ''$S\sim D_{X_{\rm I}Y_{\rm I}}^n$" to represent a sample set $S$ (of size $n$) whose each element is drawn i.i.d. from $D_{X_{\rm I}Y_{\rm I}}$.
\\

Second, we explain the concept ''$+$" in $(1-\pi^{\rm out}) D_{X_{\rm I}}+\pi^{\rm out} D_{X_{\rm O}}$.
\\

For convenience, let $P=(1-\pi^{\rm out}) D_{X_{\rm I}}$ and $Q=\pi^{\rm out} D_{X_{\rm O}} $. It is clear that $P$ and $Q$ are measures. Then $P+Q$ is also a measure, which is defined as follows: for any measurable set $A\subset \mathcal{X}$, we have
\begin{equation*}
    (P+Q)(A)=P(A)+Q(A).
\end{equation*}
For example, when $P$ and $Q$ are discrete measures, then $P+Q$ is also discrete measure: for any $\mathbf{x}\in \mathcal{X}$,
\begin{equation*}
    (P+Q)(\mathbf{x})=P(\mathbf{x})+Q(\mathbf{x}).
\end{equation*}
When $P$ and $Q$ are continuous measures with density functions $f$ and $g$, then $P+Q$ is also continuous measure with density function $f+g$: for any measurable $A\subset \mathcal{X}$,
\begin{equation*}
    P(A) = \int_A f(\mathbf{x}) {\rm d} \mathbf{x},~~~Q(A) = \int_A g(\mathbf{x}) {\rm d} \mathbf{x},
\end{equation*}
then
\begin{equation*}
    (P+Q)(A) = \int_A f(\mathbf{x})+ g(\mathbf{x}) {\rm d} \mathbf{x}.
\end{equation*}
\\

Third, we explain the concept $
   \mathbb{E}_{(\mathbf{x},y)\sim D_{XY}} \ell({ h}(\mathbf{x}),{y}).
$

The concept $\mathbb{E}_{(\mathbf{x},y)\sim D_{XY}} \ell({ h}(\mathbf{x}),{y})$ can be computed as follows:
\begin{equation*}
    \mathbb{E}_{(\mathbf{x},y)\sim D_{XY}} \ell({ h}(\mathbf{x}),{y}) = \int_{\mathcal{X}\times \mathcal{Y}_{\rm all}} \ell({ h}(\mathbf{x}),{y}) {\rm d}D_{XY}(\mathbf{x},y).
\end{equation*}

For example, when $D_{XY}$ is a finite discrete distribution: let $\mathcal{Z}=\{(\mathbf{x}^1,y^1),...,(\mathbf{x}^m,y^m)\}$ be the support set of $D_{XY}$, and assume that $a^i$ is the probability for $(\mathbf{x}^i,y^i)$, \textit{i.e.}, $a^i = D_{XY}(\mathbf{x}^i,y^i)$. Then
\begin{equation*}
\begin{split}
    \mathbb{E}_{(\mathbf{x},y)\sim D_{XY}} \ell({ h}(\mathbf{x}),{y}) &= \int_{\mathcal{X}\times \mathcal{Y}_{\rm all}} \ell({ h}(\mathbf{x}),{y}) {\rm d}D_{XY}(\mathbf{x},y)\\ & = \frac{1}{m} \sum_{i=1}^m a^i\ell({ h}(\mathbf{x}^i),{y}^i).
    \end{split}
\end{equation*}
When $D_X$ is a continuous distribution with density $f$, and $D_{Y|X}(Y=k|X=\mathbf{x})$ ($k$-th class-conditional distribution for $\mathbf{x}$) is $a^k(\mathbf{x})$, then 
\begin{equation*}
\begin{split}
    \mathbb{E}_{(\mathbf{x},y)\sim D_{XY}} \ell({ h}(\mathbf{x}),{y}) &= \int_{\mathcal{X}\times \mathcal{Y}_{\rm all}} \ell({ h}(\mathbf{x}),{y}) {\rm d}D_{XY}(\mathbf{x},y)\\ & = \int_{\mathcal{X}} \sum_{k=1}^{K+1} \ell({ h}(\mathbf{x}),k) f(\mathbf{x})a^k(\mathbf{x})  {\rm d} \mathbf{x},
    \end{split}
\end{equation*}
where $D_{Y|X}(Y=k|X=\mathbf{x})$ is the $k$-th class-conditional distribution.
%\\
%Given $\mathbf{f}=[f^1,...,f^l]^{\top}$, for any $\mathbf{x}\in \mathcal{X}$,
%\begin{equation*}
%    \argmax_{k
%    \in \{1,...,l\}} f^k(\mathbf{x}):=\max \{k\in \{1,...,l\}: f^k(\mathbf{x}) \geq f^i(\mathbf{x}), \forall i=1,...,l\},
%\end{equation*}
%where $f^k$ is the $k$-th coordinate of $\mathbf{f}$ and $f^i$ is the $i$-th coordinate of $\mathbf{f}$. 
%The above definition about $\argmax$ aims to overcome some special cases. For example, there exist ${k}_1$, ${k}_2$ ($k_1< k_2$) such that $f^{k_1}(\mathbf{x})=f^{k_2}(\mathbf{x})$ and $f^{k_1}(\mathbf{x})> f^{i}(\mathbf{x})$, $f^{k_2}(\mathbf{x})> f^{i}(\mathbf{x})$, $\forall i\in \{1,...,l\}{- }\{k_1,k_2\}$. 
%Then, according to the above definition, $k_2=\argmax_{k\in\{1,...,l\}} f^k(\mathbf{x})$.
\newpage
\section{Proof of Theorem \ref{T1}}\label{SC}

\thmCone*
\begin{proof}[Proof of Theorem \ref{T1}.] 
$~$

\textbf{Proof of the First Result.}

To prove that $\mathscr{D}_{XY}'$ is a priori-unknown space, we need to show that for any $D_{XY}^{\alpha'} \in \mathscr{D}_{XY}'$, then $D_{XY}^{\alpha} \in \mathscr{D}_{XY}'$ for any $\alpha\in [0,1)$.

According to the definition of $\mathscr{D}_{XY}'$, for any $D_{XY}^{\alpha'} \in \mathscr{D}_{XY}'$, we can find a domain $D_{XY}\in \mathscr{D}_{XY}$, which can be written as $D_{XY}=(1-\pi^{\rm out})D_{X_{\rm I}Y_{\rm I}}+ \pi^{\rm out}D_{X_{\rm O}Y_{\rm O}}$ (here $\pi^{\rm out}\in [0,1)$) such that
\begin{equation*}
    D_{XY}^{\alpha'} = (1-\alpha')D_{X_{\rm I}Y_{\rm I}}+\alpha' D_{X_{\rm O}Y_{\rm O}}.
\end{equation*}

Note that $D_{XY}^{\alpha} = (1-\alpha)D_{X_{\rm I}Y_{\rm I}}+\alpha D_{X_{\rm O}Y_{\rm O}}$.

Therefore, based on the definition of $\mathscr{D}_{XY}'$, for any $\alpha\in [0,1)$, $ D_{XY}^{\alpha}\in \mathscr{D}_{XY}'$, which implies that $\mathscr{D}_{XY}'$ is a prior-known space. Additionally, for any $D_{XY}\in \mathscr{D}_{XY}$, we can rewrite $D_{XY}$ as $D_{XY}^{\pi_{\rm out}}$, thus $D_{XY}=D_{XY}^{\pi_{\rm out}}\in \mathscr{D}_{XY}'$, which implies that $\mathscr{D}_{XY}\subset \mathscr{D}_{XY}'$.

%For any $\alpha\in [0,1)$, we can find

%To prove that $\mathscr{D}_{XY}'$ is a priori-unknown space, we need to show that for any $D_{XY}^{\alpha'} \in \mathscr{D}_{XY}'$,
%\begin{equation*}
  %  D_{XY}^{\alpha}\in \mathscr{D}_{XY}',~\forall~\alpha\in [0,1).
%\end{equation*}

%It is clear that $D_{XY}^{\alpha'}$ has ID joint distribution $D_{X_{\rm I}Y_{\rm I}}$ and OOD joint distribution $D_{X_{\rm O}Y_{\rm O}}$

%According to Definition \ref{D3}, we can prove the first result easily. We omit it.  The third result is a simple conclusion of the second result. Hence, we focus on proving the second result.

\textbf{Proof of the Second Result.}

\textbf{First,} we prove that Definition \ref{D0} concludes Definition \ref{D2}, if $\mathscr{D}_{XY}$ is a prior-unknown space:

\fbox{
\parbox{0.98\textwidth}{
The domain space $\mathscr{D}_{XY}$ is a priori-unknown space, and OOD detection  is learnable in $\mathscr{D}_{XY}$ for $\mathcal{H}$. \\
$~~~~~~~~~~~~~~~~~~~~~~~~~~~~~~~~~~~~~~~~~~~~~~~~~~~~~~~~~~~~~~~~~~~~~~~~~~~~~~{\Huge \Downarrow}$
\\
OOD detection is strongly learnable in $\mathscr{D}_{XY}$ for $\mathcal{H}$: there exist an algorithm $\mathbf{A}: \cup_{n=1}^{+\infty}(\mathcal{X}\times\mathcal{Y})^n\rightarrow \mathcal{H}$, and a monotonically decreasing sequence $\epsilon(n)$, such that
$\epsilon(n)\rightarrow 0$, as $n\rightarrow +\infty$
\begin{equation*}
\begin{split}
 \mathbb{E}_{S\sim D^n_{X_{\rm I}Y_{\rm I}}} \big[R^{\alpha}_D(\mathbf{A}(S))& -\inf_{h\in \mathcal{H}}R^{\alpha}_D(h)\big] \leq \epsilon(n), ~~~\forall \alpha\in [0,1],~\forall D_{XY}\in \mathscr{D}_{XY}.
 \end{split}
\end{equation*}
}
}
\\

\noindent In the priori-unknown space, for any $D_{XY}\in \mathscr{D}_{XY}$, we have that for any $\alpha\in [0,1)$,
\begin{equation*}
    D_{XY}^{\alpha}=(1-\alpha)D_{X_{\rm I}Y_{\rm I}}+\alpha D_{X_{\rm O}Y_{\rm O}}\in \mathscr{D}_{XY}.
\end{equation*}
Then, according to the definition of learnability of OOD detection, we have an algorithm $\mathbf{A}$ and a monotonically decreasing sequence
$\epsilon_{\rm cons}(n)\rightarrow 0$, as $n\rightarrow +\infty$, such that for any $\alpha\in [0,1)$,
\begin{equation*}
    \mathbb{E}_{S\sim D^n_{X_{\rm I}Y_{\rm I}}} R_{D^{\alpha}}(\mathbf{A}(S))\leq \inf_{h\in \mathcal{H}}R_{D^{\alpha}}(h) + \epsilon_{\rm cons}(n),~~(\textnormal{{by~the~property~of~priori}-{\rm unknown~space}})
\end{equation*}
where 
\begin{equation*}
     R_{D^{\alpha}}(\mathbf{A}(S))=\int_{\mathcal{X}\times \mathcal{Y}_{\rm all}} \ell(\mathbf{A}(S)(\mathbf{x}),y){\rm d}D^{\alpha}_{XY}(\mathbf{x},y),~~~R_{D^{\alpha}}(h)=\int_{\mathcal{X}\times \mathcal{Y}_{\rm all}} \ell(h(\mathbf{x}),y){\rm d}D^{\alpha}_{XY}(\mathbf{x},y).
\end{equation*}
Since $R_{D^{\alpha}}(\mathbf{A}(S))=R_{D}^{\alpha}(\mathbf{A}(S))$ and $R_{D^{\alpha}}(h)=R_{D}^{\alpha}(h)$, we have that
\begin{equation}\label{Eq1}
    \mathbb{E}_{S\sim D^n_{X_{\rm I}Y_{\rm I}}} R_{D}^{\alpha}(\mathbf{A}(S))\leq \inf_{h\in \mathcal{H}}R_{D}^{\alpha}(h) + \epsilon_{\rm cons}(n),~~~ \forall \alpha\in[0,1).
\end{equation}
{Next},
we consider the case that $\alpha=1$. Note that
\begin{equation}\label{Riskinequality1}
   \liminf_{\alpha\rightarrow 1} \inf_{h\in\mathcal{H}} R_{D}^{\alpha}(h) \geq \liminf_{\alpha \rightarrow 1} \alpha \inf_{h\in\mathcal{H}} R_{D}^{\rm out}(h) =  \inf_{h\in\mathcal{H}} R_{D}^{\rm out}(h).
\end{equation}
Then, we assume that $h_{\epsilon} \in \mathcal{H}$ satisfies that
\begin{equation*}
    R_{D}^{\rm out}(h_{\epsilon})-\inf_{h\in\mathcal{H}} R_{D}^{\rm out}(h) \leq \epsilon.
\end{equation*}
It is obvious that
\begin{equation*}
    R_{D}^{\alpha}(h_{\epsilon}) \geq  \inf_{h\in\mathcal{H}} R_{D}^{\alpha}(h).
\end{equation*}
Let $\alpha\rightarrow 1$. Then, for any $\epsilon>0$,
\begin{equation*}
    R_{D}^{\rm out}(h_{\epsilon})=\lim_{\alpha\rightarrow 1} R_{D}^{\alpha}(h_{\epsilon})=\limsup_{\alpha\rightarrow 1}R_{D}^{\alpha}(h_{\epsilon}) \geq \limsup_{\alpha\rightarrow 1} \inf_{h\in\mathcal{H}} R_{D}^{\alpha}(h),
\end{equation*}
which implies that
\begin{equation}\label{Riskinequality2}
    \inf_{h\in\mathcal{H}} R_{D}^{\rm out}(h) = \lim_{\epsilon\rightarrow 0}  R_{D}^{\rm out}(h_{\epsilon})\geq \lim_{\epsilon\rightarrow 0}\limsup_{\alpha\rightarrow 1} \inf_{h\in\mathcal{H}} R_{D}^{\alpha}(h)= \limsup_{\alpha\rightarrow 1} \inf_{h\in\mathcal{H}} R_{D}^{\alpha}(h).
\end{equation}
Combining Eq. (\ref{Riskinequality1}) with Eq. (\ref{Riskinequality2}), we have
\begin{equation}\label{Riskinequality3}
    \inf_{h\in\mathcal{H}} R_{D}^{\rm out}(h) = \limsup_{\alpha\rightarrow 1} \inf_{h\in\mathcal{H}} R_{D}^{\alpha}(h) = \liminf_{\alpha\rightarrow 1} \inf_{h\in\mathcal{H}} R_{D}^{\alpha}(h),
\end{equation}
which implies that
\begin{equation}\label{Riskinequality4}
    \inf_{h\in\mathcal{H}} R_{D}^{\rm out}(h) = \lim_{\alpha\rightarrow 1} \inf_{h\in\mathcal{H}} R_{D}^{\alpha}(h).
\end{equation}
Note that
\begin{equation*}
     \mathbb{E}_{S\sim D^n_{X_{\rm I}Y_{\rm I}}} R_{D}^{\alpha}(\mathbf{A}(S)) = (1-\alpha)\mathbb{E}_{S\sim D^n_{X_{\rm I}Y_{\rm I}}} R_{D}^{\rm in}(\mathbf{A}(S))+\alpha \mathbb{E}_{S\sim D^n_{X_{\rm I}Y_{\rm I}}} R_{D}^{\rm out}(\mathbf{A}(S)).
\end{equation*}
Hence, Lebesgue's Dominated Convergence Theorem \cite{cohn2013measure} implies that
\begin{equation}\label{Riskinequality5}
    \lim_{\alpha\rightarrow 1}\mathbb{E}_{S\sim D^n_{X_{\rm I}Y_{\rm I}}} R_{D}^{\alpha}(\mathbf{A}(S)) = \mathbb{E}_{S\sim D^n_{X_{\rm I}Y_{\rm I}}} R_{D}^{\rm out}(\mathbf{A}(S)).
\end{equation}

Using Eq. (\ref{Eq1}), we have that
\begin{equation}\label{Eq18}
   \lim_{\alpha \rightarrow 1} \mathbb{E}_{S\sim D^n_{X_{\rm I}Y_{\rm I}}} R_{D}^{\alpha}(\mathbf{A}(S))\leq  \lim_{\alpha \rightarrow 1} \inf_{h\in \mathcal{H}}R_{D}^{\alpha}(h) + \epsilon_{\rm cons}(n).
\end{equation}
Combining Eq. (\ref{Riskinequality4}), Eq. (\ref{Riskinequality5}) with Eq. (\ref{Eq18}), we obtain that
\begin{equation*}
  \mathbb{E}_{S\sim D^n_{X_{\rm I}Y_{\rm I}}} R_{D}^{\rm out}(\mathbf{A}(S))\leq  \inf_{h\in \mathcal{H}}R_{D}^{\rm out}(h) + \epsilon_{\rm cons}(n).
\end{equation*}
Since $R_D^{\rm out}(\mathbf{A}(S))=R_{D}^{1}(\mathbf{A}(S))$ and $R_D^{\rm out}(h)=R_{D}^{1}(h)$, we obtain that \begin{equation}\label{Eq2}
  \mathbb{E}_{S\sim D^n_{X_{\rm I}Y_{\rm I}}} R_{D}^{1}(\mathbf{A}(S))\leq  \inf_{h\in \mathcal{H}}R_{D}^{1}(h) + \epsilon_{\rm cons}(n).
\end{equation}
\\
Combining Eq. (\ref{Eq1}) and Eq. (\ref{Eq2}), we have proven that: if the  domain space $\mathscr{D}_{XY}$ is a priori-unknown space, then OOD detection  is learnable in $\mathscr{D}_{XY}$ for $\mathcal{H}$.\\
$~~~~~~~~~~~~~~~~~~~~~~~~~~~~~~~~~~~~~~~~~~~~~~~~~~~~~~~~~~~~~~~~~~~~~~~~~~~~~~~{\Huge \Downarrow}$
\\
OOD detection is strongly learnable in $\mathscr{D}_{XY}$ for $\mathcal{H}$: there exist an algorithm $\mathbf{A}: \cup_{n=1}^{+\infty}(\mathcal{X}\times\mathcal{Y})^n\rightarrow \mathcal{H}$, and a monotonically decreasing sequence $\epsilon(n)$, such that
$\epsilon(n)\rightarrow 0$, as $n\rightarrow +\infty$,
\begin{equation*}
\begin{split}
 \mathbb{E}_{S\sim D^n_{X_{\rm I}Y_{\rm I}}} R^{\alpha}_D(\mathbf{A}(S))&\leq \inf_{h\in \mathcal{H}}R^{\alpha}_D(h) + \epsilon(n),~~~\forall \alpha\in [0,1],~\forall D_{XY}\in \mathscr{D}_{XY}.
 \end{split}
\end{equation*}

\noindent \textbf{Second}, we prove that Definition \ref{D2} concludes Definition \ref{D0}:
\\
\fbox{
\parbox{1\textwidth}{
OOD detection is strongly learnable in $\mathscr{D}_{XY}$ for $\mathcal{H}$: there exist an algorithm $\mathbf{A}: \cup_{n=1}^{+\infty}(\mathcal{X}\times\mathcal{Y})^n\rightarrow \mathcal{H}$, and a monotonically decreasing sequence $\epsilon(n)$, such that
$\epsilon(n)\rightarrow 0$, as $n\rightarrow +\infty$,
\begin{equation*}
\begin{split}
 \mathbb{E}_{S\sim D^n_{X_{\rm I}Y_{\rm I}}} \big[R^{\alpha}_D(\mathbf{A}(S))&- \inf_{h\in \mathcal{H}}R^{\alpha}_D(h)\big] \leq \epsilon(n),~~~\forall \alpha\in [0,1],~\forall D_{XY}\in \mathscr{D}_{XY}.
 \end{split}
\end{equation*}
$~~~~~~~~~~~~~~~~~~~~~~~~~~~~~~~~~~~~~~~~~~~~~~~~~~~~~~~~~~~~~~~~~~~~~~~~~~~~~~~~{\Huge \Downarrow}$\\
${~~~~~~~~~~~~~~~~~~~~~~~~~~~~~~~~~~~~~~~~~~~~~~~}$OOD detection is learnable in $\mathscr{D}_{XY}$ for $\mathcal{H}$.
}
}

If we set $\alpha=\pi^{\rm out}$, then $
 \mathbb{E}_{S\sim D^n_{X_{\rm I}Y_{\rm I}}} R^{\alpha}_D(\mathbf{A}(S))\leq \inf_{h\in \mathcal{H}}R^{\alpha}_D(h) + \epsilon(n)
$ implies that
\begin{equation*}
\begin{split}
    \mathbb{E}_{S\sim D^n_{X_{\rm I}Y_{\rm I}}} R_D(\mathbf{A}(S)) \leq \inf_{h\in \mathcal{H}} R_D(h) + \epsilon(n),
    \end{split}
\end{equation*}
which means that OOD detection  is learnable in $\mathscr{D}_{XY}$ for $\mathcal{H}$. We have completed this proof.

\textbf{Proof of the Third Result.} 

The third result is a simple conclusion of the second result. Hence, we omit it.
\end{proof}
\section{Proof of Theorem \ref{T3}}\label{SD}

Before introducing the proof of Theorem \ref{T3}, we extend Condition \ref{C1} to a general version (Condition \ref{C2}). Then, Lemma  \ref{C1andC2} proves that Conditions \ref{C1} and \ref{C2} are the necessary conditions for the learnability of OOD detection. First, we provide the details of Condition \ref{C2}.

 Let  $\Delta_l^{\rm o} =\{(\lambda_1,...,\lambda_l): \sum_{j=1}^l \lambda_j<1~ {\rm and}~ \lambda_j \geq 0, \forall j=1,...,l\}$, where $l$ is a positive integer. Next, we introduce an important definition as follows:
\begin{Definition}[OOD Convex Decomposition and Convex Domain]\label{OODCD}
Given any domain $D_{XY}\in \mathscr{D}_{XY}$, we say joint distributions $Q_1,...,Q_l$, which are defined over $\mathcal{X}\times \{K+1\}$, are the OOD convex decomposition for $D_{XY}$, if 
\begin{equation*}
D_{XY}=(1-\sum_{j=1}^l \lambda_j)D_{X_{\rm I}Y_{\rm I}}+\sum_{j=1}^l \lambda_j Q_j,
\end{equation*}
for some  $(\lambda_1,...,\lambda_l)\in \Delta_l^{\rm o}$. We also say domain $D_{XY}\in \mathscr{D}_{XY}$ is an OOD convex domain corresponding to OOD convex decomposition $Q_1,...,Q_l$, if for any $(\alpha_1,...,\alpha_l)\in \Delta_l^{\rm o}$, 
\begin{equation*}
   (1-\sum_{j=1}^l \alpha_j)D_{X_{\rm I}Y_{\rm I}}+\sum_{j=1}^l \alpha_j Q_j\in \mathscr{D}_{XY}.
\end{equation*}
\end{Definition}
We extend the  linear condition (Condition \ref{C1}) to a multi-linear scenario.
\begin{Condition}[Multi-linear Condition]\label{C2}
    For each OOD convex domain $D_{XY}\in \mathscr{D}_{XY}$ corresponding to OOD convex decomposition $Q_1,...,Q_l$, the following function 
    \begin{equation*}
    f_{D,Q}(\alpha_1,...,\alpha_{l}):= \inf_{h\in \mathcal{H}}  \Big ((1-\sum_{j=1}^{l}\alpha_j) R_D^{\rm in}(h) + \sum_{j=1}^l \alpha_j R_{Q_j}(h) \Big ),~~~\forall (\alpha_1,...,\alpha_{l})\in\Delta_l^{\rm o}
    \end{equation*}
  satisfies that
\begin{equation*}
    f_{D,Q}(\alpha_1,...,\alpha_{l})= (1-\sum_{j=1}^{l}\alpha_j) f_{D,Q}(\mathbf{0})+\sum_{j=1}^{l}\alpha_j f_{D,Q}({\bm \alpha}_j),
    \end{equation*}
    where $\mathbf{0}$ is the $1\times l$ vector, whose elements are $0$, and ${\bm \alpha}_j$ is the $1\times l$ vector, whose $j$-th element is $1$ and other elements are $0$.
\end{Condition}
When $l=1$ and the domain space $\mathscr{D}_{XY}$ is a priori-unknown space, Condition \ref{C2} degenerates into  Condition \ref{C1}.
%Similarly, given a different joint distribution $Q_{XY}$, we can define $R_{Q,k}(h),R_{Q,u}(h)$ and $R_{Q}^{\alpha}(h)$.
Lemma \ref{C1andC2} shows that Condition \ref{C2} is necessary for the learnability of OOD detection.
\vspace{0.5cm}
\begin{lemma}\label{C1andC2}
Given a priori-unknown space $\mathscr{D}_{XY}$ and a hypothesis space $\mathcal{H}$, if OOD detection  is learnable in $\mathscr{D}_{XY}$ for $\mathcal{H}$, then Conditions \ref{C1} and \ref{C2} hold.
\end{lemma}
\begin{proof}[Proof of Lemma \ref{C1andC2}]
$~$

Since Condition \ref{C1} is a special case of Condition \ref{C2}, we only need to prove that Condition \ref{C2} holds. 

 For any OOD convex domain $D_{XY}\in \mathscr{D}_{XY}$ corresponding to OOD convex decomposition $Q_1,...,Q_l$,  and any $(\alpha_1,...,\alpha_l) \in \Delta_l^{\rm o}$, we set
 \begin{equation*}
     Q^{\bm \alpha}= \frac{1}{\sum_{i=1}^{l}\alpha_i} \sum_{j=1}^l \alpha_j Q_j.
 \end{equation*}
Then, we define
\begin{equation*}
    D_{XY}^{\bm \alpha}=(1-\sum_{i=1}^{l}\alpha_i)D_{X_{\rm I}Y_{\rm I}}+(\sum_{i=1}^{l}\alpha_i)  Q^{\bm \alpha},~\textnormal{which~belongs~to~}\mathscr{D}_{XY}.
\end{equation*}
Let
\begin{equation*}
     R^{\bm \alpha}_D(h)=\int_{\mathcal{X}\times \mathcal{Y}_{\rm all}} \ell(h(\mathbf{x}),y){\rm d}D_{XY}^{\bm \alpha}(\mathbf{x},y).
\end{equation*}

Since OOD detection is learnable in $\mathscr{D}_{XY}$ for $\mathcal{H}$,  there exist an algorithm $\mathbf{A}: \cup_{n=1}^{+\infty}(\mathcal{X}\times\mathcal{Y})^n\rightarrow \mathcal{H}$, and a monotonically decreasing sequence $\epsilon(n)$, such that
$\epsilon(n)\rightarrow 0$, as $n\rightarrow +\infty$, and
\begin{equation*}
 0\leq  \mathbb{E}_{S\sim D^n_{X_{\rm I}Y_{\rm I}}} R^{\bm \alpha}_D(\mathbf{A}(S)) - \inf_{h\in \mathcal{H}}R^{\bm \alpha}_D(h)\leq \epsilon(n).
\end{equation*}

Note that 
\begin{equation*}
\begin{split}
    &\mathbb{E}_{S\sim D^n_{X_{\rm I}Y_{\rm I}}} R^{\bm \alpha}_D(\mathbf{A}(S))=(1-\sum_{j=1}^{l}\alpha_j) \mathbb{E}_{S\sim D^n_{X_{\rm I}Y_{\rm I}}} R^{\rm in}_D(\mathbf{A}(S)) +\sum_{j=1}^{l}\alpha_j \mathbb{E}_{S\sim D^n_{X_{\rm I}Y_{\rm I}}}R_{Q_j}(\mathbf{A}(S)),
    \end{split}
\end{equation*}
and
\begin{equation*}
    \inf_{h\in \mathcal{H}}R^{\bm \alpha}_D(h) = f_{D,Q}(\alpha_1,...,\alpha_{l}),
\end{equation*}
where
\begin{equation*}
    R_{Q_j}(\mathbf{A}(S)) = \int_{\mathcal{X}\times \{K+1\}} \ell(\mathbf{A}(S)(\mathbf{x}),y){\rm d}Q_j(\mathbf{x},y).
\end{equation*}
Therefore, we have that for any $ (\alpha_1,...,\alpha_l)\in \Delta_l^{\rm o}$,
\begin{equation}\label{Eq::converge}
\begin{split}
  &\big |(1-\sum_{j=1}^{l}\alpha_j) \mathbb{E}_{S\sim D^n_{X_{\rm I}Y_{\rm I}}} R^{\rm in}_D(\mathbf{A}(S)) +\sum_{j=1}^{l}\alpha_j \mathbb{E}_{S\sim D^n_{X_{\rm I}Y_{\rm I}}}R_{Q_j}(\mathbf{A}(S)) - f_{D,Q}(\alpha_1,...,\alpha_{l})\big |\leq \epsilon(n).
  \end{split}
\end{equation}
Let 
\begin{equation*}
g_n(\alpha_1,...,\alpha_l)=(1-\sum_{j=1}^{l}\alpha_j) \mathbb{E}_{S\sim D^n_{X_{\rm I}Y_{\rm I}}} R^{\rm in}_D(\mathbf{A}(S)) +\sum_{j=1}^{l}\alpha_j \mathbb{E}_{S\sim D^n_{X_{\rm I}Y_{\rm I}}}R_{Q_j}(\mathbf{A}(S)).
\end{equation*}
\noindent Note that Eq. \eqref{Eq::converge} implies that
\begin{equation}\label{aim5}
\begin{split}
    \lim_{n\rightarrow +\infty } g_n(\alpha_1,...,\alpha_l) &= f_{D,Q}(\alpha_1,...,\alpha_{l}),~~\forall (\alpha_1,...,\alpha_l)\in \Delta_l^{\rm o},
   \\ 
   \lim_{n\rightarrow +\infty } g_n(\mathbf{0}) &= f_{D,Q}(\mathbf{0}).
    \end{split}
\end{equation}

\textbf{Step 1.} Since ${\bm \alpha}_j\notin \Delta_l^{\rm o}$, we need to prove that
\begin{equation}
    \lim_{n\rightarrow +\infty} \mathbb{E}_{S\sim D^n_{X_{\rm I}Y_{\rm I}}} R_{Q_j}(\mathbf{A}(S))  = f({\bm \alpha}_j), \textit{i.e.},  \lim_{n\rightarrow +\infty} g_n({\bm \alpha}_j)  = f({\bm \alpha}_j),
\end{equation}
where ${\bm \alpha}_j$ is the $1\times l$ vector, whose $j$-th element is $1$ and other elements are $0$.

Let $\tilde{D}_{XY}=0.5*D_{X_{\rm I}Y_{\rm I}}+0.5* Q_j$. The second result of Theorem \ref{T1} implies that
\begin{equation*}
\begin{split}
 \mathbb{E}_{S\sim D^n_{X_{\rm I}Y_{\rm I}}} R^{\rm out}_{\tilde{D}}(\mathbf{A}(S))\leq \inf_{h\in \mathcal{H}}R^{\rm out}_{\tilde{D}}(h) + \epsilon(n).
 \end{split}
\end{equation*}
Since $R^{\rm out}_{\tilde{D}}(\mathbf{A}(S))=R_{Q_j}(\mathbf{A}(S))$ and $R^{\rm out}_{\tilde{D}}(h)=R_{Q_j}(h)$,
\begin{equation*}
    \mathbb{E}_{S\sim D^n_{X_{\rm I}Y_{\rm I}}} R_{Q_j}(\mathbf{A}(S))\leq \inf_{h\in \mathcal{H}}R_{Q_j}(h) + \epsilon(n).
\end{equation*}
Note that $ \inf_{h\in \mathcal{H}}R_{Q_j}(h)  \leq \mathbb{E}_{S\sim D^n_{X_{\rm I}Y_{\rm I}}} R_{Q_j}(\mathbf{A}(S))$. We have
\begin{equation}\label{aim1}
   0\leq  \mathbb{E}_{S\sim D^n_{X_{\rm I}Y_{\rm I}}} R_{Q_j}(\mathbf{A}(S))- \inf_{h\in \mathcal{H}}R_{Q_j}(h)\leq \epsilon(n).
\end{equation}
Eq. \eqref{aim1} implies that
\begin{equation}\label{aim2}
   \lim_{n\rightarrow +\infty} \mathbb{E}_{S\sim D^n_{X_{\rm I}Y_{\rm I}}} R_{Q_j}(\mathbf{A}(S))= \inf_{h\in \mathcal{H}}R_{Q_j}(h).
\end{equation}

We note that $\inf_{h\in \mathcal{H}}R_{Q_j}(h)=f_{D,Q}({\bm \alpha}_j)$. Therefore,
\begin{equation}\label{aim3}
   \lim_{n\rightarrow +\infty} \mathbb{E}_{S\sim D^n_{X_{\rm I}Y_{\rm I}}} R_{Q_j}(\mathbf{A}(S))= f_{D,Q}({\bm \alpha}_j),~ \textit{i.e.}, \lim_{n\rightarrow +\infty} g_n({\bm \alpha}_j)  = f({\bm \alpha}_j).
\end{equation}

\textbf{Step 2.} It is easy to check that for any $(\alpha_1,...,\alpha_l)\in \Delta_l^{\rm o}$,
\begin{equation}\label{finalequ}
\begin{split}
    \lim_{n\rightarrow +\infty } g_n(\alpha_1,...,\alpha_l) &=   \lim_{n\rightarrow +\infty } \big ((1-\sum_{j=1}^{l}\alpha_j)g_n(\mathbf{0})+ \sum_{j=1}^{l}\alpha_j g_n({\bm \alpha}_j) \big)\\&=(1-\sum_{j=1}^{l}\alpha_j) \lim_{n\rightarrow +\infty } g_n(\mathbf{0})+ \sum_{j=1}^{l}\alpha_j \lim_{n\rightarrow +\infty } g_n({\bm \alpha}_j).
    \end{split}
\end{equation}
According to Eq. \eqref{aim5} and Eq. \eqref{aim3}, we have
\begin{equation}\label{finalequ1}
\begin{split}
    & \lim_{n\rightarrow +\infty } g_n(\alpha_1,...,\alpha_l)=f_{D,Q}(\alpha_1,...,\alpha_{l}),~~\forall (\alpha_1,...,\alpha_l)\in \Delta_l^{\rm o},
    \\ &
    \lim_{n\rightarrow +\infty } g_n(\mathbf{0})=f_{D,Q}(\mathbf{0}),
       \\ &  \lim_{n\rightarrow +\infty} g_n({\bm \alpha}_j)  = f({\bm \alpha}_j),
     \end{split}
\end{equation}
 Combining Eq. (\ref{finalequ1}) with Eq. (\ref{finalequ}), we complete the proof.
\end{proof}
\vspace{0.5cm}

\begin{lemma}\label{Lemma1}
\begin{equation*}
 \inf_{h\in \mathcal{H}}R_D^{\alpha}(h)=(1-\alpha)\inf_{h\in \mathcal{H}}R_D^{\rm in}(h)+\alpha\inf_{h\in \mathcal{H}}R_D^{\rm out}(h),~\forall \alpha\in [0,1),
 \end{equation*}
% {~}
 %\\
 %{~}
 \textbf{if and only if} for any $\epsilon>0$,
 \begin{equation*}
     \{ h' \in \mathcal{H}: R_D^{\rm in}(h') \leq \inf_{h\in \mathcal{H}} R_D^{\rm in}(h)+2\epsilon\} \cap  \{ h' \in \mathcal{H}: R_D^{\rm out}(h') \leq \inf_{h\in \mathcal{H}} R_D^{\rm out}(h)+2\epsilon\}\neq \emptyset.
 \end{equation*}
 \end{lemma}
 \begin{proof}[Proof of Lemma \ref{Lemma1}]
For the sake of convenience, we set $f_D(\alpha)=\inf_{h\in \mathcal{H}}R_D^{\alpha}(h)$, for any $\alpha \in [0,1]$.

\noindent  \textbf{First}, we prove that $f_D(\alpha)= (1-\alpha)f_D(0)+\alpha f_D(1)$, ~$\forall \alpha\in [0,1)$ implies \begin{equation*}
     \{ h' \in \mathcal{H}: R_D^{\rm in}(h') \leq \inf_{h\in \mathcal{H}} R_D^{\rm in}(h)+2\epsilon\} \cap  \{ h' \in \mathcal{H}: R_D^{\rm out}(h') \leq \inf_{h\in \mathcal{H}} R_D^{\rm out}(h)+2\epsilon \}\neq \emptyset.
 \end{equation*}
 For any $\epsilon>0$ and $0\leq \alpha<1$, we can find $h_{\epsilon}^{\alpha}\in \mathcal{H}$ satisfying that
 \begin{equation*}
     R_D^{\alpha}(h_{\epsilon}^{\alpha}) \leq \inf_{h\in \mathcal{H}} R_D^{\alpha}(h)+\epsilon.
 \end{equation*}
 Note that
 \begin{equation*}
     \inf_{h\in \mathcal{H}} R_D^{\alpha}(h)= \inf_{h\in \mathcal{H}} \Big ((1-\alpha)R_D^{\rm in}(h)+\alpha R_D^{\rm out}(h) \Big )\geq (1-\alpha)\inf_{h\in \mathcal{H}} R_D^{\rm in}(h)+\alpha \inf_{h\in \mathcal{H}} R_D^{\rm out}(h).
 \end{equation*}
 Therefore,
 \begin{equation}\label{L1.eq1}
(1-\alpha)\inf_{h\in \mathcal{H}} R_D^{\rm in}(h)+\alpha \inf_{h\in \mathcal{H}} R_D^{\rm out}(h)   \leq  \inf_{h\in \mathcal{H}} R_D^{\alpha}(h) \leq R_D^{\alpha}(h_{\epsilon}^{\alpha}) \leq \inf_{h\in \mathcal{H}} R_D^{\alpha}(h)+\epsilon.
 \end{equation}
 Note that $f_D(\alpha)= (1-\alpha)f_D(0)+\alpha f_D(1), \forall \alpha\in [0,1)$, \textit{i.e.},
 \begin{equation}\label{L1.eq2}
 \inf_{h\in \mathcal{H}} R_D^{\alpha}(h)= (1-\alpha)\inf_{h\in \mathcal{H}} R_D^{\rm in}(h)+\alpha \inf_{h\in \mathcal{H}} R_D^{\rm out}(h), \forall \alpha\in [0,1).
 \end{equation}
 
\noindent  Using Eqs. (\ref{L1.eq1}) and (\ref{L1.eq2}), we have that for any $0\leq \alpha<1$,
\begin{equation}\label{AboveEq}
\epsilon \geq \big |R^{\alpha}_D(h_{\epsilon}^{\alpha})-\inf_{h\in \mathcal{H}} R_D^{\alpha}(h)\big |=  \big |(1-\alpha)\big (R_D^{\rm in}(h_{\epsilon}^{\alpha})-\inf_{h\in \mathcal{H}} R_D^{\rm in}(h)\big)+\alpha\big (R_D^{\rm out}(h_{\epsilon}^{\alpha})-\inf_{h\in \mathcal{H}} R_D^{\rm out}(h)\big) \big |.
\end{equation}
Since $R_D^{\rm out}(h_{\epsilon}^{\alpha})-\inf_{h\in \mathcal{H}} R_D^{\rm out}(h)\geq 0$ and $R_D^{\rm in}(h_{\epsilon}^{\alpha})-\inf_{h\in \mathcal{H}} R_D^{\rm in}(h)\geq 0$, Eq. (\ref{AboveEq}) implies that: for any $0< \alpha<1$,
\begin{equation*}
\begin{split}
  &  R_D^{\rm in}(h_{\epsilon}^{\alpha}) \leq \inf_{h\in \mathcal{H}} R_D^{\rm in}(h)+\epsilon /(1-\alpha),
  \\ &
  R_D^{\rm out}(h_{\epsilon}^{\alpha}) \leq \inf_{h\in \mathcal{H}} R_D^{\rm out}(h)+\epsilon /\alpha.
\end{split}
\end{equation*}
Therefore,
\begin{equation*}
  h_{\epsilon}^\alpha \in \{ h' \in \mathcal{H}: R_D^{\rm in}(h') \leq \inf_{h\in \mathcal{H}} R_D^{\rm in}(h)+\epsilon/(1-\alpha)\} \cap  \{ h' \in \mathcal{H}: R_D^{\rm out}(h') \leq \inf_{h\in \mathcal{H}} R_D^{\rm out}(h)+\epsilon/\alpha \}.
 \end{equation*}
 If we set $\alpha=0.5$, we obtain that for any $\epsilon>0$,
\begin{equation*}
 \{ h' \in \mathcal{H}: R_D^{\rm in}(h') \leq \inf_{h\in \mathcal{H}} R_D^{\rm in}(h)+2\epsilon \}\cap  \{ h' \in \mathcal{H}: R_D^{\rm out}(h') \leq \inf_{h\in \mathcal{H}} R_D^{\rm out}(h)+2\epsilon \}\neq \emptyset.
 \end{equation*}
 \\
\textbf{Second}, we prove that for any $\epsilon>0$, if 
\begin{equation*}
     \{ h' \in \mathcal{H}: R_D^{\rm in}(h') \leq \inf_{h\in \mathcal{H}} R_D^{\rm in}(h)+2\epsilon\} \cap  \{ h' \in \mathcal{H}: R_D^{\rm out}(h') \leq \inf_{h\in \mathcal{H}} R_D^{\rm out}(h)+2\epsilon\}\neq \emptyset,
 \end{equation*}
 then $f_D(\alpha)= (1-\alpha)f_D(0)+\alpha f_D(1)$, for any $\alpha\in [0,1)$.
 \\
 \\
Let $h_{\epsilon}\in  \{ h' \in \mathcal{H}: R_D^{\rm in}(h') \leq \inf_{h\in \mathcal{H}} R_D^{\rm in}(h)+2\epsilon\} \cap  \{ h' \in \mathcal{H}: R_D^{\rm out}(h') \leq \inf_{h\in \mathcal{H}} R_D^{\rm out}(h)+2\epsilon\}$.

Then,
\begin{equation*}
 \inf_{h\in \mathcal{H}} R_D^{\alpha}(h) \leq   R_D^{\alpha}(h_{\epsilon})\leq (1-\alpha) \inf_{h\in \mathcal{H}} R_D^{\rm in}(h)+\alpha \inf_{h\in \mathcal{H}} R_D^{\rm out}(h)+2\epsilon \leq \inf_{h\in \mathcal{H}} R_D^{\alpha}(h)+2\epsilon,
\end{equation*}
which implies that $|f_D(\alpha)-(1-\alpha)f_D(0)-\alpha f_D(1)|\leq 2\epsilon$.

As $\epsilon \rightarrow 0$, $|f_D(\alpha)-(1-\alpha)f_D(0)-\alpha f_D(1)|\leq 0$. We have completed the proof.
 \end{proof}
 \vspace{0.5cm}
\thmCondoneIff*
\begin{proof}[Proof of Theorem \ref{T3}]
Based on Lemma \ref{C1andC2}, we obtain that
 Condition \ref{C1} is the necessary condition for the learnability of OOD detection in the single-distribution space $\mathscr{D}_{XY}^{D_{XY}}$. Next, it suffices to prove that Condition \ref{C1} is the sufficient  condition for the learnability of OOD detection in the single-distribution space $\mathscr{D}_{XY}^{D_{XY}}$.
 We use Lemma \ref{Lemma1} to prove the sufficient condition. 
 
 Let $\mathscr{F}$ be the infinite sequence set that consists of all infinite sequences, whose coordinates are hypothesis functions, \textit{i.e.},
 \begin{equation*}
 \mathscr{F}=\{{\bm h}=(h_1,...,h_n,...): \forall h_n\in \mathcal{H}, n=1,....,+\infty\}.
 \end{equation*}
For each ${\bm h}\in \mathscr{F}$, there is a corresponding algorithm $\mathbf{A}_{\bm h}$\footnote{In this paper, we regard an algorithm as a mapping from $\cup_{n=1}^{+\infty}(\mathcal{X}\times\mathcal{Y})^n$ to $\mathcal{H}$. So we can design an algorithm like this.}: $\mathbf{A}_{\bm h}(S)=h_n,~{\rm if}~|S|=n$. $\mathscr{F}$ generates an algorithm class $\mathscr{A}=\{\mathbf{A}_{\bm h}: \forall {\bm h}\in \mathscr{F}\}$. We select a consistent algorithm from the algorithm class $\mathscr{A}$. 

We construct a special infinite sequence $\tilde{{\bm h}}=(\tilde{h}_1,...,\tilde{h}_n,...)\in \mathscr{F}$. For each positive integer $n$, we select $\tilde{h}_n$ from $  \{ h' \in \mathcal{H}: R_D^{\rm in}(h') \leq \inf_{h\in \mathcal{H}} R_D^{\rm in}(h)+2/n\} \cap  \{ h' \in \mathcal{H}: R_D^{\rm out}(h') \leq \inf_{h\in \mathcal{H}} R_D^{\rm out}(h)+2/n\}$ (the existence of $\tilde{h}_n$ is based on Lemma \ref{Lemma1}). It is easy to check that
 \begin{equation*}
 \begin{split}
  & \mathbb{E}_{S\sim D^n_{X_{\rm I}Y_{\rm I}}}  R_D^{\rm in}(\mathbf{A}_{\tilde{{\bm h}}}(S))\leq  \inf_{h\in \mathcal{H}} R_D^{\rm in}(h)+2/n.
   \\
   & \mathbb{E}_{S\sim D^n_{X_{\rm I}Y_{\rm I}}}  R_D^{\rm out}(\mathbf{A}_{\tilde{{\bm h}}}(S))\leq  \inf_{h\in \mathcal{H}} R_D^{\rm out}(h)+2/n.
    \end{split}
 \end{equation*}
 Since $(1-\alpha)\inf_{h\in \mathcal{H}} R_D^{\rm in}(h)+\alpha \inf_{h\in \mathcal{H}} R_D^{\rm out}(h)\leq \inf_{h\in \mathcal{H}} R_D^{\rm \alpha}(h)$, we obtain that for any $\alpha\in [0,1]$, 
 \begin{equation*}
 \begin{split}
  & \mathbb{E}_{S\sim D^n_{X_{\rm I}Y_{\rm I}}}  R_D^{\alpha}(\mathbf{A}_{\tilde{{\bm h}}}(S))\leq  \inf_{h\in \mathcal{H}} R_D^{\alpha}(h)+2/n.
    \end{split}
 \end{equation*}
We have completed this proof.
\end{proof}

\section{Proofs of Theorem \ref{T5} and Theorem \ref{T4} }\label{SF}
\subsection{Proof of Theorem \ref{T5}}
\thmImpOne*

\begin{proof}[Proof of Theorem \ref{T5}]
We \textbf{first} explain how we get $f_{\rm I}$ and $f_{\rm O}$ in Definition {\ref{def:overlap}}. Since $D_X$ is absolutely continuous respect to $\mu$ ($D_X\ll \mu$), then $D_{X_{\rm I}}\ll \mu$ and $D_{X_{\rm O}}\ll \mu$. By Radon-Nikodym Theorem \cite{cohn2013measure}, we know there exist two non-negative functions defined over $\mathcal{X}$: $f_{\rm I}$
and $f_{\rm O}$ such that for any $\mu$-measurable set $A\subset \mathcal{X}$,
\begin{equation*}
    D_{X_{\rm I}}(A)=\int_{A} f_{\rm I}(\mathbf{x}){\rm d} \mu(\mathbf{x}),~~D_{X_{\rm O}}(A)=\int_{A} f_{\rm O}(\mathbf{x}){\rm d} \mu(\mathbf{x}).
\end{equation*}

\textbf{Second}, we prove that for any $\alpha\in (0,1)$, $\inf_{h\in \mathcal{H}} R_{D}^{\alpha}(h)>0$.

We define $A_{m}=\{\mathbf{x}\in \mathcal{X}: f_{\rm I}(\mathbf{x})\geq \frac{1}{m}~ {\rm and} ~f_{\rm O}(\mathbf{x})\geq \frac{1}{m}\}$. It is clear that 
\begin{equation*}
    \cup_{m=1}^{+\infty} A_m =\{\mathbf{x}\in \mathcal{X}: f_{\rm I}(\mathbf{x})>0~ {\rm and} ~ f_{\rm O}(\mathbf{x})>0\}=A_{\rm overlap},
\end{equation*}
and 
\begin{equation*}
    A_m \subset A_{m+1}.
\end{equation*}

Therefore, 
\begin{equation*}
\lim_{m\rightarrow +\infty}\mu(A_m)=\mu(A_{\rm overlap})>0,
\end{equation*}
which implies that there exists $m_0$ such that
\begin{equation*}
    \mu(A_{m_0})>0.
\end{equation*}

For any $\alpha\in(0,1)$, we define 
$
    c_{\alpha}= \min_{y_1\in \mathcal{Y}_{\rm all}} \big((1-\alpha)\min_{y_2\in \mathcal{Y}} \ell(y_1,y_2)+\alpha \ell(y_1,K+1)\big).
$
It is clear that $c_{\alpha}>0$ for $\alpha\in(0,1)$. Then, for any $h\in \mathcal{H}$,
\begin{equation*}
\begin{split}
    &~~~~~~R_D^{\alpha}(h)\\&=\int_{\mathcal{X}\times \mathcal{Y}_{\rm all}} \ell(h(\mathbf{x}),y){\rm d} D^{\alpha}_{XY}(\mathbf{x},y)\\&=\int_{\mathcal{X}\times \mathcal{Y}} (1-\alpha) \ell(h(\mathbf{x}),y){\rm d} D_{X_{\rm I}Y_{\rm I}}(\mathbf{x},y)+\int_{\mathcal{X}\times \{K+1\}} \alpha \ell(h(\mathbf{x}),y){\rm d} D_{X_{\rm O}Y_{\rm O}}(\mathbf{x},y)\\&\geq \int_{A_{m_0}\times \mathcal{Y}} (1-\alpha) \ell(h(\mathbf{x}),y){\rm d} D_{X_{\rm I}Y_{\rm I}}(\mathbf{x},y)+\int_{A_{m_0}\times \{K+1\}} \alpha \ell(h(\mathbf{x}),y){\rm d} D_{X_{\rm O}Y_{\rm O}}(\mathbf{x},y)\\ & = \int_{A_{m_0}} \big((1-\alpha)\int_{ \mathcal{Y}}  \ell(h(\mathbf{x}),y){\rm d} D_{Y_{\rm I}|X_{\rm I}}(y|\mathbf{x})\big){\rm d}D_{X_{\rm I}}(\mathbf{x})\\&+ \int_{A_{m_0}} \alpha \ell(h(\mathbf{x}),K+1){\rm d} D_{X_{\rm O}}(\mathbf{x})\\ & \geq \int_{A_{m_0}} (1-\alpha)\min_{y_2\in \mathcal{Y}}\ell(h(\mathbf{x}),y_2){\rm d}D_{X_{\rm I}}(\mathbf{x})+ \int_{A_{m_0}} \alpha \ell(h(\mathbf{x}),K+1){\rm d} D_{X_{\rm O}}(\mathbf{x})
    \\ & \geq \int_{A_{m_0}} (1-\alpha)\min_{y_2\in \mathcal{Y}}\ell(h(\mathbf{x}),y_2)f_{\rm I}(\mathbf{x}){\rm d}\mu(\mathbf{x})+ \int_{A_{m_0}} \alpha \ell(h(\mathbf{x}),K+1)f_{\rm O}(\mathbf{x}){\rm d} \mu(\mathbf{x})\\ & \geq \frac{1}{m_0} \int_{A_{m_0}} (1-\alpha)\min_{y_2\in \mathcal{Y}}\ell(h(\mathbf{x}),y_2){\rm d}\mu(\mathbf{x})+ \frac{1}{m_0}\int_{A_{m_0}} \alpha \ell(h(\mathbf{x}),K+1){\rm d} \mu (\mathbf{x})\\ & = \frac{1}{m_0} \int_{A_{m_0}} \big( (1-\alpha)\min_{y_2\in \mathcal{Y}}\ell(h(\mathbf{x}),y_2)+ \alpha \ell(h(\mathbf{x}),K+1) \big) {\rm d} \mu (\mathbf{x}) \geq \frac{c_{\alpha}}{m_0}\mu(A_{m_0})>0.
    \end{split}
\end{equation*}
Therefore,
\begin{equation*}
    \inf_{h\in \mathcal{H}} R_D^{\alpha}(h)\geq \frac{c_{\alpha}}{m_0}\mu(A_{m_0})>0.
\end{equation*}

\textbf{Third}, Condition \ref{C1} indicates that  $\inf_{h\in \mathcal{H}} R_{D}^{\alpha}(h)=(1-\alpha)\inf_{h\in \mathcal{H}}R_{D}^{\rm in}(h)+\alpha \inf_{h\in \mathcal{H}}R_{D}^{\rm in}(h) = 0 $ (here we have used conditions $\inf_{h\in \mathcal{H}}R_D^{\rm in}(h)=0$ and $\inf_{h\in \mathcal{H}}R_D^{\rm out}(h)=0$), which contradicts with $\inf_{h\in \mathcal{H}} R_{D}^{\alpha}(h)>0$ ($\alpha\in (0,1)$). Therefore, Condition \ref{C1} does not hold. Using Lemma \ref{C1andC2}, we obtain that OOD detection in $\mathscr{D}_{XY}$ is not learnable  for $\mathcal{H}$.
\end{proof}

\subsection{Proof of Theorem \ref{T4}}

\thmImpTotal*
\begin{proof}[Proof of Theorem \ref{T4}]
We need to prove that OOD detection is not learnable in the total space $\mathscr{D}_{XY}^{\rm all}$ for $\mathcal{H}$, if $\mathcal{H}$ is non-trivial, \textit{i.e.},
$
\{\mathbf{x}\in \mathcal{X}:\exists h_1,h_2\in \mathcal{H}, \textnormal{s.t.}~ h_1(\mathbf{x})\in \mathcal{Y}, h_2(\mathbf{x})=K+1\}\neq \emptyset.
$

The main idea is to construct a domain $D_{XY}$ satisfying that:
\\
1) the ID and OOD distributions have overlap (Definition {\ref{def:overlap}}); and
2) $R_{D}^{\rm in}(h_1)=0$, $ R_{D}^{\rm out}(h_2)=0$.

According to the condition that $\mathcal{H}$ is non-trivial, we know that there exist $h_1, h_2\in \mathcal{H}$ such that $h_1(\mathbf{x}_1)\in \mathcal{Y}, h_2(\mathbf{x}_1)=K+1$, for some $\mathbf{x}_1\in \mathcal{X}$. We set $D_{XY}= 0.5*\delta_{(\mathbf{x}_1,h_1(\mathbf{x}_1))}+0.5*\delta_{(\mathbf{x}_1,h_2(\mathbf{x}_1))}$, where $\delta$ is the Dirac measure. It is easy to check that $R_{D}^{\rm in}(h_1)=0$, $R_{D}^{\rm out}(h_2)=0$, which implies that $\inf_{h\in \mathcal{H}}R_{D}^{\rm in}(h)=0$ and $\inf_{h\in \mathcal{H}}R_{D}^{\rm out}(h)=0$. In addition, the ID distribution $\delta_{(\mathbf{x}_1,h_1(\mathbf{x}_1))}$ and OOD distribution $\delta_{(\mathbf{x}_1,h_2(\mathbf{x}_1))}$ have overlap $\mathbf{x}_1$.
By using Theorem \ref{T5}, we have completed this proof.
\end{proof}

\section{Proof of Theorem \ref{T12}}\label{SH}
Before proving Theorem \ref{T12}, we need three important lemmas.
\begin{lemma}\label{lemma10} Suppose that $D_{XY}$ is a domain with OOD convex decomposition $Q_1,...,Q_l$ (convex decomposition is given by Definition \ref{OODCD} in Appendix \ref{SD}), and $D_{XY}$ is a finite discrete distribution, then (the definition of $f_{D,Q}$ is given in Condition \ref{C2})
 \begin{equation*}
    f_{D,Q}(\alpha_1,...,\alpha_{l})= (1-\sum_{j=1}^{l}\alpha_j) f_{D,Q}(\mathbf{0})+\sum_{j=1}^{l}\alpha_j f_{D,Q}({\bm \alpha}_j),~~~\forall (\alpha_1,...,\alpha_{l})\in \Delta_l^{\rm o}, 
    \end{equation*}
    \textbf{if and only if} 
    \begin{equation*}
     \argmin_{h\in \mathcal{H}} R_D(h) = \bigcap_{j=1}^l \argmin_{h\in \mathcal{H}} R_{Q_j}(h) \bigcap \argmin_{h\in \mathcal{H}} R_D^{\rm in}(h),
    \end{equation*}
    where $\mathbf{0}$ is the $1\times l$ vector, whose elements are $0$, and ${\bm \alpha}_j$ is the $1\times l$ vector, whose $j$-th element is $1$ and other elements are $0$, and 
 \begin{equation*}   
 R_{Q_j}(h) = \int_{\mathcal{X}\times \{K+1\}} \ell(h(\mathbf{x}),y){\rm d}Q_j(\mathbf{x},y).
\end{equation*}
\end{lemma}
\begin{proof}[Proof of Lemma \ref{lemma10}] To better understand this proof, we recall the definition of  $f_{D,Q}(\alpha_1,...,\alpha_{l})$:
   \begin{equation*}
    f_{D,Q}(\alpha_1,...,\alpha_{l})= \inf_{h\in \mathcal{H}}  \Big ((1-\sum_{j=1}^{l}\alpha_j) R_D^{\rm in}(h) + \sum_{j=1}^l \alpha_j R_{Q_j}(h) \Big ),~~~\forall (\alpha_1,...,\alpha_{l})\in\Delta_l^{\rm o}
    \end{equation*}
%\end{equation*}
\textbf{First}, we prove that if \begin{equation*}
    f_{D,Q}(\alpha_1,...,\alpha_{l})= (1-\sum_{j=1}^{l}\alpha_j) f_{D,Q}(\mathbf{0})+\sum_{j=1}^{l}\alpha_j f_{D,Q}({\bm \alpha}_j),~~~\forall (\alpha_1,...,\alpha_{l})\in \Delta_l^{\rm o},
    \end{equation*}
   then,
   \begin{equation*}
     \argmin_{h\in \mathcal{H}} R_D(h) = \bigcap_{j=1}^l \argmin_{h\in \mathcal{H}}R_{Q_j}(h) \bigcap \argmin_{h\in \mathcal{H}} R_D^{\rm in}(h).
    \end{equation*}
    
\noindent    
Let $D_{XY}=(1-\sum_{j=1}^l \lambda_j)D_{X_{\rm I}Y_{\rm I}}+\sum_{j=1}^l \lambda_j Q_j$, for some $(\lambda_1,...,\lambda_l)\in \Delta_l^{\rm o}$. Since $D_{XY}$ has finite support set, we have
\begin{equation*}
     \argmin_{h\in \mathcal{H}} R_D(h) = \argmin_{h\in \mathcal{H}}\Big ((1-\sum_{j=1}^l \lambda_j) R_D^{\rm in}(h) + \sum_{j=1}^l \lambda_j R_{Q_j}(h)\Big) \neq \emptyset.
\end{equation*}

We can find that $h_{0}\in   \argmin_{h\in \mathcal{H}}\Big ((1-\sum_{j=1}^l \lambda_j) R_D^{\rm in}(h) + \sum_{j=1}^l \lambda_j R_{Q_j}(h)\Big)$. Hence,
    \begin{equation}\label{threeinequality-1}
     (1-\sum_{j=1}^l \lambda_j) R_D^{\rm in}(h_{0}) + \sum_{j=1}^l \lambda_j R_{Q_j}(h_{0})  = \inf_{h\in \mathcal{H}}  \Big ((1-\sum_{j=1}^l \lambda_j) R_D^{\rm in}(h) + \sum_{j=1}^l \lambda_j R_{Q_j}(h) \Big ) .
    \end{equation}
    Note that the condition $f_{D,Q}(\alpha_1,...,\alpha_{l})= (1-\sum_{j=1}^{l}\alpha_j) f_{D,Q}(\mathbf{0})+\sum_{j=1}^{l}\alpha_j f_{D,Q}({\bm \alpha}_j)$ implies 
    \begin{equation}\label{threeinequality-2}
    (1-\sum_{j=1}^l \lambda_j)\inf_{h\in \mathcal{H}}  R_D^{\rm in}(h) + \sum_{j=1}^{l} \lambda_j \inf_{h\in \mathcal{H}}  R_{Q_j}(h)   = \inf_{h\in \mathcal{H}}  \Big ( (1-\sum_{j=1}^l \lambda_j) R_D^{\rm in}(h) + \sum_{j=1}^{l} \lambda_j R_{Q_j}(h) \Big ).
    \end{equation}
    Therefore, Eq. \eqref{threeinequality-1} and Eq. \eqref{threeinequality-2} imply that
    \begin{equation}\label{threeinequality}
    \begin{split}
   (1-\sum_{j=1}^l \lambda_j)\inf_{h\in \mathcal{H}}  R_D^{\rm in}(h) + \sum_{j=1}^{l} \lambda_j \inf_{h\in \mathcal{H}}  R_{Q_j}(h) = (1-\sum_{j=1}^l \lambda_j) R_D^{\rm in}(h_{0}) + \sum_{j=1}^{l} \lambda_j R_{Q_j}(h_{0}).
   \end{split}
    \end{equation}

\noindent   
Since $ R_D^{\rm in}(h_{0})\geq \inf_{h\in \mathcal{H}}  R_D^{\rm in}(h)$ and $R_{Q_j}(h_{0})\geq \inf_{h\in \mathcal{H}}  R_{Q_j}^{\rm in}(h)$, for $j=1,...,l$, then using Eq. \eqref{threeinequality}, we have that
\begin{equation*}
\begin{split}
&R_D^{\rm in}(h_{0})= \inf_{h\in \mathcal{H}}  R_D^{\rm in}(h),
\\
&
R_{Q_j}(h_{0})=  \inf_{h\in \mathcal{H}}  R_{Q_j}(h),~~~\forall j=1,...,l,
\end{split}
\end{equation*}
which implies that
\begin{equation*}
h_{0} \in     \bigcap_{j=1}^{l} \argmin_{h\in \mathcal{H}} R_{Q_j}(h) \bigcap   \argmin_{h\in \mathcal{H}} R_D^{\rm in}(h).
    \end{equation*}
    Therefore, 
    \begin{equation}\label{subsetOOD1}
       \argmin_{h\in \mathcal{H}} R_D(h)\subset  \bigcap_{j=1}^{l} \argmin_{h\in \mathcal{H}} R_{Q_j}(h) \bigcap   \argmin_{h\in \mathcal{H}} R_D^{\rm in}(h).
    \end{equation}
    
 \noindent Additionally, using 
 \begin{equation*}
 f_{D,Q}(\alpha_1,...,\alpha_{l})= (1-\sum_{j=1}^{l}\alpha_j) f_{D,Q}(\mathbf{0})+\sum_{j=1}^{l}\alpha_j f_{D,Q}({\bm \alpha}_j),~\forall (\alpha_1,...,\alpha_{l})\in \Delta_l^{\rm o},
 \end{equation*}
 we obtain that for any $h'\in    \bigcap_{j=1}^{l} \argmin_{h\in \mathcal{H}} R_{Q_j}(h) \bigcap   \argmin_{h\in \mathcal{H}} R_D^{\rm in}(h)$, 
 \begin{equation*}
    \begin{split}
 \inf_{h\in \mathcal{H}}   R_D(h) =& \inf_{h\in \mathcal{H}}  \Big ((1-\sum_{j=1}^{l}\lambda_j) R_D^{\rm in}(h) + \sum_{j=1}^l \lambda_j R_{Q_j}(h) \Big )\\=&(1-\sum_{j=1}^{l}\lambda_j)  \inf_{h\in \mathcal{H}} R_D^{\rm in}(h) + \sum_{j=1}^l \lambda_j  \inf_{h\in \mathcal{H}}  R_{Q_j}(h)\\ =  & (1-\sum_{j=1}^{l}\lambda_j) R_D^{\rm in}(h') + \sum_{j=1}^l \lambda_j R_{Q_j}(h')=R_D(h'),
    \end{split}
    \end{equation*}
    which implies that 
    \begin{equation*}
        h'\in \argmin_{h\in \mathcal{H}} R_D(h).
    \end{equation*}
    Therefore,  
    \begin{equation}\label{subsetOOD2}
    \bigcap_{j=1}^{l} \argmin_{h\in \mathcal{H}} R_{Q_j}(h) \bigcap   \argmin_{h\in \mathcal{H}} R_D^{\rm in}(h) \subset \argmin_{h\in \mathcal{H}} R_D(h).
    \end{equation}
    \\
    \noindent Combining Eq. \eqref{subsetOOD1} with Eq. \eqref{subsetOOD2},  we obtain that 
    \begin{equation*}
    \bigcap_{j=1}^{l} \argmin_{h\in \mathcal{H}} R_{Q_j}(h) \bigcap   \argmin_{h\in \mathcal{H}} R_D^{\rm in}(h)= \argmin_{h\in \mathcal{H}} R_D(h).
    \end{equation*}

\noindent \textbf{Second}, we prove that if
    \begin{equation*}
  \argmin_{h\in \mathcal{H}} R_{D}(h) =    \bigcap_{j=1}^{l}  \argmin_{h\in \mathcal{H}} R_{Q_j}(h) \bigcap  \argmin_{h\in \mathcal{H}} R_D^{\rm in}(h),
    \end{equation*}
   then,
   \begin{equation*}
    f_{D,Q}(\alpha_1,...,\alpha_{l})= (1-\sum_{j=1}^{l}\alpha_j) f_{D,Q}(\mathbf{0})+\sum_{j=1}^{l}\alpha_j f_{D,Q}({\bm \alpha}_j),~~~\forall (\alpha_1,...,\alpha_{l})\in \Delta_l^{\rm o}.
    \end{equation*}

\noindent    We set
    \begin{equation*}
        h_{0} \in \bigcap_{j=1}^{l}  \argmin_{h\in \mathcal{H}} R_{Q_j}(h) \bigcap  \argmin_{h\in \mathcal{H}} R_D^{\rm in}(h),
    \end{equation*}
    then, for any $(\alpha_1,...,\alpha_{l})\in \Delta_l^{\rm o}$,
    \begin{equation*}
    \begin{split}
(1-\sum_{j=1}^{l}\alpha_j)  \inf_{h\in \mathcal{H}} R_D^{\rm in}(h) + \sum_{j=1}^l \alpha_j  \inf_{h\in \mathcal{H}}  R_{Q_j}(h)  &\leq  \inf_{h\in \mathcal{H}}  \Big ((1-\sum_{j=1}^{l}\alpha_j) R_D^{\rm in}(h) + \sum_{j=1}^l \alpha_j R_{Q_j}(h) \Big )\\  & \leq    (1-\sum_{j=1}^{l}\alpha_j) R_D^{\rm in}(h_{0}) + \sum_{j=1}^l \alpha_j R_{Q_j}(h_{0})\\  &=(1-\sum_{j=1}^{l}\alpha_j)  \inf_{h\in \mathcal{H}} R_D^{\rm in}(h) + \sum_{j=1}^l \alpha_j  \inf_{h\in \mathcal{H}}  R_{Q_j}(h).
    \end{split}
    \end{equation*}
    Therefore, for any $(\alpha_1,...,\alpha_{l})\in \Delta_l^{\rm o}$,
      \begin{equation*}
    \begin{split}
& (1-\sum_{j=1}^{l}\alpha_j)  \inf_{h\in \mathcal{H}} R_D^{\rm in}(h) + \sum_{j=1}^l \alpha_j  \inf_{h\in \mathcal{H}}  R_{Q_j}(h) = \inf_{h\in \mathcal{H}}  \Big ((1-\sum_{j=1}^{l}\alpha_j) R_D^{\rm in}(h) + \sum_{j=1}^l \alpha_j R_{Q_j}(h) \Big ),
    \end{split}
    \end{equation*}
    which implies that: for any $(\alpha_1,...,\alpha_{l})\in \Delta_l^{\rm o}$, 
    \begin{equation*}
        f_{D,Q}(\alpha_1,...,\alpha_{l})= (1-\sum_{j=1}^{l}\alpha_j) f_{D,Q}(\mathbf{0})+\sum_{j=1}^{l}\alpha_j f_{D,Q}({\bm \alpha}_j).
    \end{equation*}
    \noindent We have completed this proof.
\end{proof}
\newpage

\begin{lemma}
\label{T9}
Suppose that Assumption \ref{ass1} holds.
% $\mathcal{H}$ is a separate space for OOD, 
If there is a finite discrete domain $D_{XY}\in \mathscr{D}_{XY}^{s}$ such that
$
    \inf_{h\in \mathcal{H}}R_{D}^{\rm out}({\bm h})>0,
$ then OOD detection is not learnable in  $\mathscr{D}_{XY}^{s}$ for $\mathcal{H}$.
\end{lemma}
\begin{proof}[Proof of Lemma \ref{T9}]
Suppose that ${\rm supp} D_{X_{\rm O}}=\{\mathbf{x}_1^{\rm out},...,\mathbf{x}_l^{\rm out}\}$, then it is clear that $D_{XY}$ has OOD convex decomposition $\delta_{\mathbf{x}_1^{\rm out}},...,\delta_{\mathbf{x}_l^{\rm out}}$, where $\delta_{\mathbf{x}}$ is the dirac measure whose support set is $\{\mathbf{x}\}$.

\noindent Since $\mathcal{H}$ is the separate space for OOD (\textit{i.e.}, Assumption \ref{ass1} holds), then $\forall j=1,...,l$, 
\begin{equation*}
\inf_{h\in \mathcal{H}} R_{\delta_{\mathbf{x}_j^{\rm out}}}(h)=0,
\end{equation*}
where 
\begin{equation*}
    R_{\delta_{\mathbf{x}_j^{\rm out}}}(h) = \int_{\mathcal{X}} \ell(h(\mathbf{x}),K+1){\rm d}\delta_{\mathbf{x}_j^{\rm out}}(\mathbf{x}).
\end{equation*}
\\
\noindent This implies that: if $\bigcap_{j=1}^l \argmin_{h\in \mathcal{H}} R_{\delta_{\mathbf{x}_j^{\rm out}}}(h)\neq \emptyset$, then for $\forall h'\in \bigcap_{j=1}^l \argmin_{h\in \mathcal{H}} R_{\delta_{\mathbf{x}_j^{\rm out}}}(h)$,
\begin{equation*}
   h'(\mathbf{x}_i^{\rm out}) = K+1, ~\forall i=1,...,l.
\end{equation*}
Therefore, if $ \bigcap_{j=1}^l \argmin_{h\in \mathcal{H}} R_{\delta_{\mathbf{x}_j^{\rm out}}}(h) \bigcap \argmin_{h\in \mathcal{H}} R_D^{\rm in}(h)\neq \emptyset$, \\$~~~~~~~~~~~~~~~~~~$then for any $h^*\in \bigcap_{j=1}^l \argmin_{h\in \mathcal{H}} R_{\delta_{\mathbf{x}_j^{\rm out}}}(h) \bigcap \argmin_{h\in \mathcal{H}} R_D^{\rm in}(h)$, we have that
\begin{equation*}
   h^*(\mathbf{x}_i^{\rm out}) = K+1,~ \forall i=1,...,l.
\end{equation*}
\\
\noindent \textbf{Proof by Contradiction}: {assume} OOD detection is learnable in  $\mathscr{D}_{XY}^s$ for $\mathcal{H}$, then  Lemmas \ref{C1andC2} and \ref{lemma10} imply that
\begin{equation*}
     \bigcap_{j=1}^l \argmin_{h\in \mathcal{H}} R_{\delta_{\mathbf{x}_j^{\rm out}}}(h) \bigcap \argmin_{h\in \mathcal{H}} R_D^{\rm in}(h)= \argmin_{h\in \mathcal{H}} R_D(h) \neq \emptyset.
    \end{equation*}
    Therefore, for any $h^*\in  \argmin_{h\in \mathcal{H}} R_D(h)$, we have that 
    \begin{equation*}
   h^*(\mathbf{x}_i^{\rm out}) = K+1,~\forall i=1,...,l,
\end{equation*}
which implies that  for any $h^*\in  \argmin_{h\in \mathcal{H}} R_D(h)$, we have
$
   R_D^{\rm out}(h^*) = 0,
$
which implies that $\inf_{h\in \mathcal{H}} R_D^{\rm out}(h)=0$.
%\noindent  Additionally, since we \textbf{assume} that OOD detection is learnable in  $\mathscr{D}_{XY}^s$ for $\mathcal{H}$, then  Lemma \ref{C1andC2} and Lemma \ref{lemma10} also imply that
%\begin{equation}\label{newrevision}
%     \argmin_{h\in \mathcal{H}} R_D(h) =  \argmin_{h\in \mathcal{H}} R_{D}^{\rm out}(h) \bigcap \argmin_{h\in \mathcal{H}} R_D^{\rm in}(h),
 %   \end{equation}
 %   because $D_{X_{\rm O}Y_{\rm O}}$ is also an OOD convex decomposition for $D_{XY}$. 
    
%Eq. \eqref{newrevision} implies that $ \argmin_{h\in \mathcal{H}} R_D(h) \subset \argmin_{h\in \mathcal{H}} R_{D}^{\rm out}(h).$
    
\noindent It is clear that $\inf_{h\in \mathcal{H}} R_D^{\rm out}(h)=0$ is \textbf{inconsistent} with the condition $\inf_{h\in \mathcal{H}} R_D^{\rm out}(h)>0$. Therefore, 
OOD detection is not learnable in  $\mathscr{D}_{XY}^s$ for $\mathcal{H}$.
\end{proof}
\vspace{0.5cm}

\begin{lemma}\label{LemmaforTheorem5}
If Assumption \ref{ass1} holds, 
  ${\rm VCdim}(\phi\circ \mathcal{H})=v<+\infty$ and $\sup_{{ h}\in \mathcal{H}} |\{\mathbf{x}\in \mathcal{X}: { h}(\mathbf{x})\in \mathcal{Y}\}| > m$ such that $v<m$, then
  OOD detection is not learnable in $\mathscr{D}_{XY}^{s}$ for $\mathcal{H}$, where $\phi$ maps ID's labels to ${1}$ and maps OOD's labels to $2$.
\end{lemma}

\begin{proof}[Proof of Lemma \ref{LemmaforTheorem5}]
Due to $\sup_{{ h}\in \mathcal{H}} |\{\mathbf{x}\in \mathcal{X}: {\bm h}(\mathbf{x})\in \mathcal{Y}\}| > m$, we can obtain a set
\begin{equation*}
   C=\{\mathbf{x}_1,...,\mathbf{x}_m,\mathbf{x}_{m+1}\},
\end{equation*}
which satisfies that there exists $\tilde{h}\in \mathcal{H}$ such that $\tilde{h}(\mathbf{x}_i)\in \mathcal{Y}$ for any $i=1,...,m,m+1$.

\noindent Let $\mathcal{H}_C^{\phi}=\{(\phi\circ h(\mathbf{x}_1),...,\phi\circ h(\mathbf{x}_m),\phi\circ h(\mathbf{x}_{m+1}):h\in \mathcal{H}\}$. It is clear that 
\begin{equation*}
(1,1,...,1)=(\phi\circ \tilde{h}(\mathbf{x}_1),...,\phi\circ \tilde{h}(\mathbf{x}_m),\phi\circ \tilde{h}(\mathbf{x}_{m+1}))\in\mathcal{H}_C^{\phi}, 
\end{equation*}
where $(1,1,...,1)$ means all elements are $1$.

Let $\mathcal{H}_{m+1}^{\phi}=\{(\phi\circ h(\mathbf{x}_1),...,\phi\circ h(\mathbf{x}_m),\phi\circ h(\mathbf{x}_{m+1}):h~\textnormal{is~any~hypothesis~function~from~}\mathcal{X}~\textnormal{to}~\mathcal{Y}_{\rm all}\}$.

Clearly, $\mathcal{H}_C^{\phi}\subset \mathcal{H}_{m+1}^{\phi}$ and $|\mathcal{H}_{m+1}^{\phi}|=2^{m+1}$. Sauer-Shelah-Perles Lemma (Lemma 6.10 in \citep{shalev2014understanding}) implies that 
\begin{equation*}
    |\mathcal{H}^{\phi}_C|\leq \sum_{i=0}^v \tbinom{m+1}{i}.
\end{equation*}
Since $
    \sum_{i=0}^v \tbinom{m+1}{i}<2^{m+1}-1
$ (because $v<m$), we obtain that $|\mathcal{H}^{\phi}_C|\leq2^{m+1}-2$. Therefore, $\mathcal{H}^{\phi}_C\cup \{(2,2...,2)\}$ is a proper subset of $\mathcal{H}_{m+1}^{\phi}$, where $(2,2,...,2)$ means that all elements are $2$.  Note that $(1,1...,1)$ (all elements are 1) also belongs to $\mathcal{H}_C^{\phi}$. Hence, $\mathcal{H}^{\phi}_C\cup \{(2,2...,2)\}\cup \{(1,1...,1)\}$ is a proper subset of $\mathcal{H}_{m+1}^{\phi}$, which implies that we can obtain a hypothesis function $h'$ satisfying that:
\begin{equation*}
\begin{split}
&1) (\phi\circ h'(\mathbf{x}_1),...,\phi\circ h'(\mathbf{x}_m),\phi\circ h'(\mathbf{x}_{m+1}))\notin \mathcal{H}^{\phi}_C;
\\
&2) \textnormal{~ There~ exist}~  \mathbf{x}_j,\mathbf{x}_p\in C \textnormal{~ such ~that}~ \phi\circ h'(\mathbf{x}_j)=2~ {\rm and}~ \phi\circ h'(\mathbf{x}_p)=1.
\end{split}
\end{equation*}

Let $C_{\rm I}=C\cap \{\mathbf{x}\in \mathcal{X}:\phi\circ h'(\mathbf{x})=1\}$ and $C_{\rm O}=C\cap \{\mathbf{x}\in \mathcal{X}:\phi\circ h'(\mathbf{x})=2\}$;

\noindent Then, we construct a special domain $D_{XY}$:
\begin{equation*}
    D_{XY} = 0.5*D_{X_{\rm I}}*D_{Y_{\rm I}|X_{\rm I}}+0.5* D_{X_{\rm O}}*D_{Y_{\rm O}|X_{\rm O}},~\textnormal{where}
\end{equation*}
\begin{equation*}
    D_{X_{\rm I}}= \frac{1}{|C_{\rm I}|}\sum_{\mathbf{x}\in C_{\rm I}} \delta_{\mathbf{x}}~~~\textnormal{and}~~~D_{Y_{\rm I}|X_{\rm I}}(y|\mathbf{x})=1,~\textnormal{if}~~\tilde{h}(\mathbf{x})=y~~\textnormal{and}~~\mathbf{x}\in C_{\rm I};
\end{equation*}
and
\begin{equation*}
    D_{X_{\rm O}}= \frac{1}{|C_{\rm O}|}\sum_{\mathbf{x}\in C_{\rm O}} \delta_{\mathbf{x}}~~~\textnormal{and}~~~D_{Y_{\rm O}|X_{\rm O}}(K+1|\mathbf{x})=1,~\textnormal{if}~~\mathbf{x}\in C_{\rm O}.
\end{equation*}
%whose margin distribution $D_X$ is $\frac{1}{m+1} \sum_{i=1}^{m+1} \delta_{\mathbf{x}_i}$ and conditional distribution $D_{Y|X}$ satisfies that for any $\mathbf{x}\in \mathcal{X}$, if $\tilde{h}(\mathbf{x})=y$ and $\phi\circ h'(\mathbf{x})=1$, then $D_{Y|X}(y|\mathbf{x})=1$; and if $\phi\circ h'(\mathbf{x})=2$, then $D_{Y|X}(K+1|\mathbf{x})=1$. Since $\phi\circ h'(\mathbf{x}_p)=1$, $\pi^{\rm out}=D_{Y=K+1}<1$. So $D_{XY}$ is a domain for OOD detection.

\noindent Since $D_{XY}$ is a finite discrete distribution and $(\phi\circ h'(\mathbf{x}_1),...,\phi\circ h'(\mathbf{x}_m),\phi\circ h'(\mathbf{x}_{m+1}))\notin \mathcal{H}^{\phi}_C$, it is clear that $\argmin_{h\in \mathcal{H}} R_D(h)\neq \emptyset$ and $\inf_{h\in \mathcal{H}} R_D(h)>0$. 

Additionally, $R_D^{\rm in}(\tilde{h})=0$. Therefore, $\inf_{h\in \mathcal{H}}R_D^{\rm in}(h)=0$.

\noindent \textbf{Proof by Contradiction}: {suppose} that  OOD detection is learnable in  $\mathscr{D}_{XY}^s$ for $\mathcal{H}$, then  Lemma \ref{C1andC2} implies that
\begin{equation*}
    \inf_{h\in \mathcal{H}} R_D(h) = 0.5* \inf_{h\in \mathcal{H}} R_D^{\rm in}(h)+ 0.5*\inf_{h\in \mathcal{H}} R_D^{\rm out}(h).
    \end{equation*}
 Therefore, if  OOD detection is learnable in  $\mathscr{D}_{XY}^s$ for $\mathcal{H}$, then $\inf_{h\in \mathcal{H}} R_D^{\rm out}(h)>0$.

\noindent
 Until now, we have constructed a domain $D_{XY}$ (defined over $\mathcal{X}\times \mathcal{Y}_{\rm all}$) with finite support and satisfying that $\inf_{h\in \mathcal{H}} R_D^{\rm out}(h)>0$. Note that $\mathcal{H}$ is the separate space for OOD data (Assumption \ref{ass1} holds).  Using Lemma \ref{T9}, we know that OOD detection is not learnable in  $\mathscr{D}_{XY}^s$ for $\mathcal{H}$, which is \textbf{inconsistent} with our assumption that   OOD detection is learnable in  $\mathscr{D}_{XY}^s$ for $\mathcal{H}$. Therefore,  OOD detection is not learnable in  $\mathscr{D}_{XY}^s$ for $\mathcal{H}$. We have completed the proof.
\end{proof}
\vspace{0.5cm}
\thmImpSeptwo*
\begin{proof}[Proof of Theorem \ref{T12}]
Let ${\rm VCdim}(\phi\circ \mathcal{H})=v$. Since $\sup_{{ h}\in \mathcal{H}} |\{\mathbf{x}\in \mathcal{X}: {\bm h}(\mathbf{x})\in \mathcal{Y}\}|=+\infty$, it is clear that $\sup_{{ h}\in \mathcal{H}} |\{\mathbf{x}\in \mathcal{X}: {\bm h}(\mathbf{x})\in \mathcal{Y}\}|>v$. Using Lemma \ref{LemmaforTheorem5}, we complete this proof.
\end{proof}
\section{Proofs of Theorem \ref{T13} and Theorem \ref{T17}}\label{SI}
\subsection{Proof of Theorem \ref{T13}}
Firstly, we need two lemmas, which are motivated by Lemma {19.2} and Lemma {19.3} in \citep{shalev2014understanding}.
\begin{lemma}\label{L14}
Let $C_1$,...,$C_{r}$ be a cover of space $\mathcal{X}$, \textit{i.e.}, $\sum_{i=1}^r C_i=\mathcal{X}$. Let $S_{X}=\{\mathbf{x}^1,...,\mathbf{x}^n\}$ be a sequence of $n$ data  drawn from $D_{X_{\rm I}}$, i.i.d. Then 
\begin{equation*}
    \mathbb{E}_{S_X \sim D^n_{X_{\rm I}}} \Big(\sum_{i:C_i \cap S_X =\emptyset} D_{X_{\rm I}}(C_i) \Big) \leq \frac{r}{en}.
\end{equation*}
\end{lemma}
\begin{proof}[Proof of Lemma \ref{L14}]
\begin{equation*}
 \mathbb{E}_{S_X\sim D^n_{X_{\rm I}}} \Big(\sum_{i:C_i \cap S_X =\emptyset} D_{X_{\rm I}}(C_i) \Big) = \sum_{i=1}^r\Big( D_{X_{\rm I}}(C_i) \cdot \mathbb{E}_{S_X\sim D^n_{X_{\rm I}}} \big (\mathbf{1}_{C_i \cap S_X =\emptyset}\big ) \Big),
 \end{equation*}
 where $\mathbf{1}$ is the characteristic function.
 
 \noindent For each $i$,
 \begin{equation*}
 \begin{split}
     \mathbb{E}_{S_X\sim D^n_{X_{\rm I}}} \big (\mathbf{1}_{C_i \cap S_X =\emptyset}\big ) & = \int_{\mathcal{X}^n} \mathbf{1}_{C_i \cap S_X =\emptyset} {\rm d} D^n_{X_{\rm I}}(S_X)\\ &  = \big ( \int_{\mathcal{X}} \mathbf{1}_{C_i \cap \{\mathbf{x}\} =\emptyset} {\rm d} D_{X_{\rm I}}(\mathbf{x}) \big )^n \\ & =\big ( 1- D_{X_{\rm I}}(C_i)\big )^n \leq e^{-nD_{X_{\rm I}}(C_i)}.
     \end{split}
 \end{equation*}
 Therefore, 
 \begin{equation*}
 \begin{split}
     \mathbb{E}_{S_X\sim D^n_{X_{\rm I}}} \Big(\sum_{i:C_i \cap S =\emptyset} D_{X_{\rm I}}(C_i) \Big)  &\leq \sum_{i=1}^r D_{X_{\rm I}}(C_i)e^{-nD_{X_{\rm I}}(C_i)} \\ & \leq r\max_{i\in \{1,...,r\}} D_{X_{\rm I}}(C_i)e^{-nD_{X_{\rm I}}(C_i)}\leq \frac{r}{ne},
     \end{split}
 \end{equation*}
 here we have used inequality: $ \max_{i\in \{1,...,r\}} a_i e^{-na_i}\leq 1/{(ne)}$. The proof has been completed.
\end{proof}
\vspace{0.5cm}
\begin{lemma}\label{L15} Let $K=1$. When $\mathcal{X}\subset \mathbb{R}^d$ is a bounded set, there exists a monotonically decreasing sequence $\epsilon_{\rm cons}(m)$ satisfying that  $\epsilon_{\rm cons}(m) \rightarrow 0$, as $m\rightarrow 0$, such that
\begin{equation*}
    \mathbb{E}_{\mathbf{x}\sim  D_{X_{\rm I}},S\sim D^n_{X_{\rm I}Y_{\rm I}}}{\rm dist}( \mathbf{x},\pi_1(\mathbf{x},S))<\epsilon_{\rm cons}(n),
\end{equation*}
where ${\rm dist}$ is the Euclidean distance, $\pi_1(\mathbf{x},S)= \argmin_{\tilde{\mathbf{x}}\in S_X} {\rm dist}(\mathbf{x},\tilde{\mathbf{x}})$, here $S_X$ is the feature part of $S$, \textit{i.e.}, $S_X=\{\mathbf{x}^1,...,\mathbf{x}^n\}$, if $S=\{(\mathbf{x}^1,y^1),...,(\mathbf{x}^n,y^n)\}$.

\end{lemma}
\begin{proof}[Proof of Lemma \ref{L15}]
Since $\mathcal{X}$ is bounded, without loss of generality, we set  $\mathcal{X}\subset [0,1)^d$. Fix  $\epsilon=1/T$, for some integer $T$. Let $r = T^d$ and $C_1,C_2,...,C_r$ be a cover of $\mathcal{X}$: for every $(a_1,...,a_T)\in [T]^d:=[1,...,T]^d$, there exists a $C_i=\{\mathbf{x}=(x_1,...,x_d):\forall j\in \{1,...,d\}, x_j\in [(a_j-1)/T,a_j/T)\}$. 

\noindent If $\mathbf{x}, \mathbf{x}'$ belong to some $C_i$, then ${\rm dist}(\mathbf{x}, \mathbf{x}') \leq \sqrt{d}\epsilon$; otherwise, ${\rm dist}(\mathbf{x}, \mathbf{x}') \leq \sqrt{d}$. Therefore,
\begin{equation*}
\begin{split}
      &\mathbb{E}_{\mathbf{x}\sim  D_{X_{\rm I}},S\sim D^n_{X_{\rm I}Y_{\rm I}}}{\rm dist}( \mathbf{x},\pi_1(\mathbf{x},S)) 
      \\ \leq & \mathbb{E}_{S\sim D^n_{X_{\rm I}Y_{\rm I}}} \Big ( \sqrt{d}\epsilon \sum_{i:C_i \cap S_X  \neq \emptyset} D_{X_{\rm I}}(C_i) + \sqrt{d}\sum_{i:C_i \cap S_X  = \emptyset} D_{X_{\rm I}}(C_i) \Big )  \\ \leq &\mathbb{E}_{S_X \sim D^n_{X_{\rm I}}} \Big ( \sqrt{d}\epsilon \sum_{i:C_i \cap S_X  \neq \emptyset} D_{X_{\rm I}}(C_i) + \sqrt{d}\sum_{i:C_i \cap S_X  = \emptyset} D_{X_{\rm I}}(C_i) \Big )  .
      \end{split}
\end{equation*}
Note that $C_1,...,C_r$ are disjoint. 

Therefore, $\sum_{i:C_i \cap S_X  \neq \emptyset} D_{X_{\rm I}}(C_i)\leq D_{X_{\rm I}}( \sum_{i:C_i \cap S_X  \neq \emptyset} C_i) \leq 1$. Using Lemma \ref{L14}, we obtain
\begin{equation*}
\begin{split}
      &\mathbb{E}_{\mathbf{x}\sim  D_{X_{\rm I}},S\sim D^n_{X_{\rm I}Y_{\rm I}}}{\rm dist}( \mathbf{x},\pi_1(\mathbf{x},S)) 
   \leq \sqrt{d}\epsilon+\frac{r\sqrt{d}}{ne}=\sqrt{d}\epsilon+\frac{\sqrt{d}}{ne{\epsilon}^d}.
      \end{split}
\end{equation*}
If we set $\epsilon= 2n^{-1/(d+1)}$, then
\begin{equation*}
\begin{split}
      &\mathbb{E}_{\mathbf{x}\sim  D_{X_{\rm I}},S\sim D^n_{X_{\rm I}Y_{\rm I}}}{\rm dist}( \mathbf{x},\pi_1(\mathbf{x},S)) 
   \leq \frac{2\sqrt{d}}{n^{1/(d+1)}}+\frac{\sqrt{d}}{2^den^{1/(d+1)}}.
      \end{split}
\end{equation*}
If we set $\epsilon_{\rm cons}(n)=\frac{2\sqrt{d}}{n^{1/(d+1)}}+\frac{\sqrt{d}}{2^den^{1/(d+1)}}$, we complete this proof.
\end{proof}
\vspace{0.5cm}
\thmPosbSep*

\begin{proof}[Proof of Theorem \ref{T13}]
\textbf{First}, we prove that  if the hypothesis space $\mathcal{H}$ is a separate space for OOD (\textit{i.e.}, Assumption \ref{ass1} holds), the constant function $h^{\rm in}:=1 \in \mathcal{H}$, then that OOD detection is learnable in  $\mathscr{D}_{XY}^s$ for $\mathcal{H}$ implies  $\mathcal{H}_{\rm all}-\{h^{\rm out}\} \subset \mathcal{H}$.

\noindent \textbf{Proof by Contradiction}: suppose that there exists $h'\in \mathcal{H}_{\rm all}$ such that $h'\neq h^{\rm out}$ and $h'\notin \mathcal{H}$.

\noindent Let $\mathcal{X}=\{\mathbf{x}_1,...,\mathbf{x}_m\}$, $C_{\rm I}=\{\mathbf{x}\in \mathcal{X}:h'(\mathbf{x})\in \mathcal{Y}\}$ and $C_{\rm O}=\{\mathbf{x}\in \mathcal{X}:h'(\mathbf{x})=K+1\}$.

Because $h'\neq h^{\rm out}$, we know that $C_{\rm I}\neq \emptyset$.

We construct a special domain $D_{XY}\in \mathscr{D}_{XY}^s$: if $C_{\rm O}=\emptyset$, then $ D_{XY} = D_{X_{\rm I}}*D_{Y_{\rm I}|X_{\rm I}}$; otherwise,
\begin{equation*}
    D_{XY} = 0.5*D_{X_{\rm I}}*D_{Y_{\rm I}|X_{\rm I}}+0.5*D_{X_{\rm O}}*D_{Y_{\rm O}|X_{\rm O}},~~\textnormal{where}
\end{equation*}
\begin{equation*}
    D_{X_{\rm I}} = \frac{1}{|C_{\rm I}|} \sum_{\mathbf{x}\in C_{\rm I}}\delta_{\mathbf{x}}~~\textnormal{and}~~D_{Y_{\rm I}|X_{\rm I}}(y|\mathbf{x})=1,~~\textnormal{if}~h'(\mathbf{x})=y~\textnormal{and}~\mathbf{x}\in C_{\rm I},
\end{equation*}
and
\begin{equation*}
    D_{X_{\rm O}} = \frac{1}{|C_{\rm O}|} \sum_{\mathbf{x}\in C_{\rm O}}\delta_{\mathbf{x}}~~\textnormal{and}~~D_{Y_{\rm O}|X_{\rm O}}(K+1|\mathbf{x})=1,~~\textnormal{if}~\mathbf{x}\in C_{\rm O}.
\end{equation*}
\noindent Since $h'\notin \mathcal{H}$ and $|\mathcal{X}|<+\infty$, then $\argmin_{h\in \mathcal{H}} R_D(h)\neq \emptyset$, and $\inf_{h\in \mathcal{H}} R_D(h)>0$. Additionally, $R_D^{\rm in}(h^{\rm in})=0$ (here $h^{\rm in}=1$), hence, $\inf_{h\in \mathcal{H}} R_D^{\rm in}(h)=0$.

\noindent Since OOD detection  is learnable in  $\mathscr{D}_{XY}^s$ for $\mathcal{H}$, Lemma \ref{C1andC2} implies that
\begin{equation*}
    \inf_{h\in \mathcal{H}} R_D(h) = (1-\pi^{\rm out}) \inf_{h\in \mathcal{H}} R_D^{\rm in}(h)+ \pi^{\rm out}\inf_{h\in \mathcal{H}} R_D^{\rm out}(h),
    \end{equation*}
    where $\pi^{\rm out}=D_{Y}(Y=K+1)=1$ or $0.5$.
Since $\inf_{h\in \mathcal{H}} R_D^{\rm in}(h)=0$ and $\inf_{h\in \mathcal{H}} R_D(h)>0$, we obtain that $\inf_{h\in \mathcal{H}} R_D^{\rm out}(h)>0$.

\noindent Until now, we have constructed a special domain $D_{XY}\in \mathscr{D}_{XY}^s$ satisfying that $\inf_{h\in \mathcal{H}} R_D^{\rm out}(h)>0$. Using Lemma \ref{T9}, we know that OOD detection in  $\mathscr{D}_{XY}^s$  is not learnable  for $\mathcal{H}$, which is \textbf{inconsistent} with the condition that OOD detection is learnable in  $\mathscr{D}_{XY}^s$ for $\mathcal{H}$. Therefore, the assumption (there exists $h'\in \mathcal{H}_{\rm all}$ such that $h'\neq h^{\rm out}$ and $h\notin \mathcal{H}$) doesn't hold, which implies that $\mathcal{H}_{\rm all}-\{h^{\rm out}\} \subset \mathcal{H}$.

\noindent \textbf{Second}, we prove that if $\mathcal{H}_{\rm all}-\{h^{\rm out}\} \subset \mathcal{H}$, then OOD detection is learnable in  $\mathscr{D}_{XY}^s$ for $\mathcal{H}$.

\noindent To prove this result, we need to design a special  algorithm. Let $d_0=\min_{\mathbf{x},\mathbf{x}'\in \mathcal{X} ~{\rm and}~ {\mathbf{x}\neq\mathbf{x}'}}{\rm dist}(\mathbf{x},\mathbf{x}')$, where ${\rm dist}$ is the Euclidean distance. It is clear that $d_0>0$. Let 
\\
$~~~~~~~~~~~~~~~~~~~~~~~~~~~~~~~~~~~~~~~~~~~~~~~${\centering{
$ \mathbf{A}(S)(\mathbf{x})=\left \{
\begin{aligned}
~~~~~1,&~~~~{\rm if}~~ {\rm dist}( \mathbf{x},\pi_1(\mathbf{x},S))<0.5*d_0;\\
~~~~~2,&~~~~{\rm if}~~ {\rm dist}( \mathbf{x},\pi_1(\mathbf{x},S))\geq 0.5*d_0,\\
\end{aligned}
 \right.
$}}
\\
{ where}~$\pi_1(\mathbf{x},S)= \argmin_{\tilde{\mathbf{x}}\in S_{X}} {\rm dist}(\mathbf{x},\tilde{\mathbf{x}})$,
 here $S_X$ is the feature part of $S$, \textit{i.e.}, $S_X=\{\mathbf{x}^1,...,\mathbf{x}^n\}$, if $S=\{(\mathbf{x}^1,y^1),...,(\mathbf{x}^n,y^n)\}$.

For any $\mathbf{x}\in {\rm supp} D_{X_{\rm I}}$, it is easy to check that for almost all $S\sim D^n_{X_{\rm I}Y_{\rm I}}$,
\begin{equation*}
    {\rm dist}( \mathbf{x},\pi_1(\mathbf{x},S))>0.5*d_0,
\end{equation*}
which implies that
\begin{equation*}
     \mathbf{A}({S})(\mathbf{x}) =2,
\end{equation*}
hence,
\begin{equation}\label{E1}
    \mathbb{E}_{S\sim D^n_{X_{\rm I}Y_{\rm I}}} R_{D}^{\rm out}(\mathbf{A}(S))=0.
\end{equation}

Using Lemma \ref{L15}, for any $\mathbf{x}\in {\rm supp} D_{X_{\rm I}}$, we have
\begin{equation*}
    \mathbb{E}_{\mathbf{x}\sim  D_{X_{\rm I}},S\sim D^n_{X_{\rm I}Y_{\rm I}}}{\rm dist}( \mathbf{x},\pi_1(\mathbf{x},S))<\epsilon_{\rm cons}(n),
\end{equation*}
where $\epsilon_{\rm cons}(n) \rightarrow 0$, as $n\rightarrow 0$ and $\epsilon_{\rm cons}(n)$ is a monotonically decreasing
sequence.

Hence, we have that 
\begin{equation*}
   D_{X_{\rm I}}\times  D_{X_{\rm I}Y_{\rm I}}^n(\{(\mathbf{x},S):{\rm dist}( \mathbf{x},\pi_1(\mathbf{x},S))\geq 0.5*d_0 \})\leq 2\epsilon_{\rm cons}(n)/d_0,
\end{equation*}
where $ D_{X_{\rm I}}\times  D_{X_{\rm I}Y_{\rm I}}^n$ is the product measure of $D_{X_{\rm I}}$ and $D_{X_{\rm I}Y_{\rm I}}^n $ \cite{cohn2013measure}.
Therefore,
\begin{equation*}
    D_{X_{\rm I}}\times  D_{X_{\rm I}Y_{\rm I}}^n(\{(\mathbf{x},S):\mathbf{A}(S)(\mathbf{x})=1\})> 1- 2\epsilon_{\rm cons}(n)/d_0,
\end{equation*}
which implies that
\begin{equation}\label{E2}
  \mathbb{E}_{S\sim D^n_{X_{\rm I}Y_{\rm I}}} R_{D}^{\rm in} (\mathbf{A}(S)) \leq 2B\epsilon_{\rm cons}(n)/d_0,
\end{equation}
where $B=\max\{\ell(1,2),\ell(2,1)\}$. Using Eq. \eqref{E1} and Eq. \eqref{E2}, we have proved that 
\begin{equation}
    \mathbb{E}_{S\sim D_{X_{\rm I}Y_{\rm I}}^n} R_D(\mathbf{A}(S))\leq 0+ 2B\epsilon_{\rm cons}(m)/d_0 \leq \inf_{h\in \mathcal{H}}R_D(h)+2B\epsilon_{\rm cons}(m)/d_0.
\end{equation}
It is easy to check that $\mathbf{A}(S)\in \mathcal{H}_{\rm all}-\{h^{\rm out}\}$. Therefore, we have constructed a consistent  algorithm $\mathbf{A}$ for $\mathcal{H}$. We have completed this proof.
\end{proof}
\vspace{-0.2cm}

%\subsection{Proof of Theorem \ref{T16}}

%\begin{theorem}
%\label{T16}
%Let $K=1$ and $\mathcal{X}$ be a bounded set. If  $\mathcal{H}_{\rm all}-\{h^{\rm out}\} \subset \mathcal{H}$, then OOD detection is learnable in  $\mathscr{D}_{XY}^{\tau}$ for $\mathcal{H}$, where $\mathcal{H}_{\rm all}$ and $h^{\rm out}$ are defined in Theorem \ref{T13}.
%\end{theorem}
%\end{restatable}
%\begin{proof}[Proof of Theorem \ref{T16}]
%The proof is similar to the proof of the second part of Theorem \ref{T13}. The unique difference is that we set $d_0=\tau/2$ ($d_0$ is in the proof of Theorem \ref{T13}). We omit the proof.
%\end{proof}
\subsection{Proof of Theorem \ref{T17}}

\thmPosbMultione*
\begin{proof}[Proof of Theorem \ref{T17}]
Since $|\mathcal{X}|<+\infty$, we know that $|\mathcal{H}|<+\infty$, which implies that $\mathcal{H}^{\rm in}$ is agnostic PAC learnable for supervised learning in classification. Therefore, there exist an algorithm $\mathbf{A}^{\rm in}: \cup_{n=1}^{+\infty}(\mathcal{X}\times\mathcal{Y})^n\rightarrow \mathcal{H}^{\rm in}$ and a monotonically decreasing
sequence $\epsilon(n)$, such that $\epsilon(n)\rightarrow 0$, as $n\rightarrow +\infty$, and for any $D_{XY}\in \mathscr{D}_{XY}^s$,
\begin{equation*}
    \mathbb{E}_{S\sim D^n_{X_{\rm I}Y_{\rm I}}} R_D^{\rm in}(\mathbf{A}^{\rm in}(S)) \leq \inf_{h\in \mathcal{H}^{\rm in}} R_{D}^{\rm in}(h)+\epsilon(n).
\end{equation*}

Since $|\mathcal{X}|<+\infty$ and  $\mathcal{H}^{\rm b}$ almost contains all binary classifiers, then using Theorem \ref{T13} and Theorem \ref{T1}, we obtain that there exist an algorithm $\mathbf{A}^{\rm b}: \cup_{n=1}^{+\infty}(\mathcal{X}\times \{1,2\})^n\rightarrow \mathcal{H}^{\rm b}$ and a monotonically decreasing sequence $\epsilon'(n)$, such that
$\epsilon'(n)\rightarrow 0$, as $n\rightarrow +\infty$, and for any $D_{XY}\in \mathscr{D}_{XY}^s$,
\begin{equation*}
\begin{split}
    & \mathbb{E}_{S\sim D^n_{X_{\rm I}Y_{\rm I}}} R_{\phi({D})}^{\rm in}(\mathbf{A}^{\rm b}(\phi(S))) \leq \inf_{h\in \mathcal{H}^{\rm b}} R_{\phi(D)}^{\rm in}(h)+\epsilon'(n),
    \\ &
    \mathbb{E}_{S\sim D^n_{X_{\rm I}Y_{\rm I}}} R_{\phi({D})}^{\rm out}(\mathbf{A}^{\rm b}(\phi(S))) \leq \inf_{h\in \mathcal{H}^{\rm b}} R_{\phi(D)}^{\rm out}(h)+\epsilon'(n),
    \end{split}
\end{equation*}
where $\phi$ maps ID's labels to $1$ and OOD's label to $2$,
\begin{equation}
    R_{\phi({D})}^{\rm in}(\mathbf{A}^{\rm b}(\phi(S)))  = \int_{\mathcal{X}\times \mathcal{Y}} \ell(\mathbf{A}^{\rm b}(\phi(S))(\mathbf{x}), \phi(y)) {\rm d} D_{X_{\rm I}Y_{\rm I}}(\mathbf{x},y),
\end{equation}
\begin{equation}
    R_{\phi({D})}^{\rm in}(h)  = \int_{\mathcal{X}\times \mathcal{Y}} \ell(h(\mathbf{x}), \phi(y)) {\rm d} D_{X_{\rm I}Y_{\rm I}}(\mathbf{x},y),
\end{equation}
\begin{equation}
    R_{\phi({D})}^{\rm out}(\mathbf{A}^{\rm b}(\phi(S)))  = \int_{\mathcal{X}\times \{K+1\}} \ell(\mathbf{A}^{\rm b}(\phi(S))(\mathbf{x}), \phi(y)) {\rm d} D_{X_{\rm O}Y_{\rm O}}(\mathbf{x},y),
\end{equation}
and
\begin{equation}
    R_{\phi({D})}^{\rm out}(h)  = \int_{\mathcal{X}\times \{K+1\}} \ell(h(\mathbf{x}), \phi(y)) {\rm d} D_{X_{\rm O}Y_{\rm O}}(\mathbf{x},y),
\end{equation}
here $\phi(S)=\{(\mathbf{x}^1,\phi({y}^1)),...,(\mathbf{x}^n,\phi({y}^n))\}$, if $S=\{(\mathbf{x}^1,{y}^1),...,(\mathbf{x}^n,{y}^n)\}$.
\\

Note that $\mathcal{H}^b$ almost contains all classifiers, and $\mathscr{D}^s_{XY}$ is the separate space. Hence,  
\begin{equation*}
\begin{split}
    & \mathbb{E}_{S\sim D^n_{X_{\rm I}Y_{\rm I}}} R_{\phi({D})}^{\rm in}(\mathbf{A}^{\rm b}(\phi(S))) \leq \epsilon'(n),
    ~~~
    \mathbb{E}_{S\sim D^n_{X_{\rm I}Y_{\rm I}}} R_{\phi({D})}^{\rm out}(\mathbf{A}^{\rm b}(\phi(S))) \leq \epsilon'(n).
    \end{split}
\end{equation*}

\textbf{Next}, we construct an algorithm $\mathbf{A}$ using $\mathbf{A}^{\rm in}$ and $\mathbf{A}^{\rm out}$.
\\

~~~~~~~~~~~~~~~~~~~~~~~~~~~~~~~~~~~~~~~~~~~~~~~{\centering{
$ \mathbf{A}(S)(\mathbf{x})=\left \{
\begin{aligned}
K+1,&~~~~{\rm if}~~ \mathbf{A}^{\rm b}(\phi(S))(\mathbf{x})=2;\\
\mathbf{A}^{\rm in}(S)(\mathbf{x}),&~~~~{\rm if}~~ \mathbf{A}^{\rm b}(\phi(S))(\mathbf{x})=1.\\
\end{aligned}
 \right.
$}}
\\

\noindent Since $\inf_{h\in \mathcal{H}} R_{\phi(D)}^{\rm in}(\phi\circ h)=0$, $\inf_{h\in \mathcal{H}} R_{D}^{\rm out}(h)=0$, then by Condition \ref{C3}, it is easy to check that
\begin{equation*}
\inf_{h\in \mathcal{H}^{\rm in}} R_{D}^{\rm in}(h)=\inf_{h\in \mathcal{H}} R_{D}^{\rm in}(h).
\end{equation*}

Additionally, the risk $R_D^{\rm in}(\mathbf{A}(S))$ is from two parts: 1) ID data are detected as OOD data; 2) ID data are detected as ID data, but are classified as incorrect ID classes. Therefore, we have the inequality: 
\begin{equation}\label{IDlearnable}
\begin{split}
    \mathbb{E}_{S\sim D^n_{X_{\rm I}Y_{\rm I}}} R_D^{\rm in}(\mathbf{A}(S))& \leq    \mathbb{E}_{S\sim D^n_{X_{\rm I}Y_{\rm I}}} R_D^{\rm in}(\mathbf{A}^{\rm in}(S)) +  c\mathbb{E}_{S\sim D^n_{X_{\rm I}Y_{\rm I}}} R_{\phi({D})}^{\rm in}(\mathbf{A}^{\rm b}(\phi(S))) \\ & \leq \inf_{h\in \mathcal{H}^{\rm in}} R_{D}^{\rm in}(h)+\epsilon(n) + c\epsilon'(n)  =  \inf_{h\in \mathcal{H}} R_{D}^{\rm in}(h)+\epsilon(n) + c\epsilon'(n), 
    \end{split}
\end{equation}
where $c= \max_{y_1,y_2\in \mathcal{Y}} \ell(y_1,y_2)/\min\{\ell(1,2),\ell(2,1)\}$.
\\

\noindent Note that the risk $R_D^{\rm out}(\mathbf{A}(S))$ is from the case that  OOD data are detected as ID data. Therefore,
\begin{equation}\label{OODlearnable}
\begin{split}
    \mathbb{E}_{S\sim D^n_{X_{\rm I}Y_{\rm I}}} R_D^{\rm out}(\mathbf{A}(S))& \leq      c\mathbb{E}_{S\sim D^n_{X_{|rm I}Y_{\rm I}}} R_{\phi({D})}^{\rm out}(\mathbf{A}^{\rm b}(\phi(S))) \\ & \leq   c\epsilon'(n)\leq \inf_{h\in \mathcal{H}} R_D^{\rm out}(h)+ c\epsilon'(n). 
    \end{split}
\end{equation}
Note that $ (1-\alpha)\inf_{h\in \mathcal{H}} R_{D}^{\rm in}(h)+\alpha \inf_{h\in \mathcal{H}} R_{D}^{\rm out}(h)\leq \inf_{h\in \mathcal{H}} R_D^{\alpha}(h)$. Then, using Eq. \eqref{IDlearnable} and Eq. \eqref{OODlearnable}, we obtain that for any $\alpha\in [0,1]$,
\begin{equation*}
\begin{split}
    \mathbb{E}_{S\sim D^n_{X_{\rm I}Y_{\rm I}}} R_D^{\alpha}(\mathbf{A}(S))& \leq     \inf_{h\in \mathcal{H}} R_D^{\alpha}(h)+\epsilon(n)+ c\epsilon'(n). 
    \end{split}
\end{equation*}
According to Theorem \ref{T1} (the second result), we complete the proof.
\end{proof}

\section{Proofs of Theorems \ref{T-SET} and \ref{T-SET2}}
\subsection{Proof of Theorem \ref{T-SET}}
\begin{lemma}\label{lemma7-new}
Given a prior-unknown space $\mathscr{D}_{XY}$ and a hypothesis space $\mathcal{H}$, if Condition \ref{Con2} holds, then for any equivalence class $[D_{XY}']$ with respect to $\mathscr{D}_{XY}$, OOD detection is learnable in the equivalence class $[D_{XY}']$ for $\mathcal{H}$. Furthermore, the learning rate can attain $O(1/n)$.
\end{lemma}
\begin{proof}
 Let $\mathscr{F}$ be a set consisting of all infinite sequences, whose coordinates are hypothesis functions, \textit{i.e.},
 \begin{equation*}
 \mathscr{F}=\{{\bm h}=(h_1,...,h_n,...): \forall h_n\in \mathcal{H}, n=1,....,+\infty\}.
 \end{equation*}

For each ${\bm h}\in \mathscr{F}$, there is a corresponding algorithm $\mathbf{A}_{\bm h}$: $\mathbf{A}_{\bm h}(S)=h_n,~{\rm if}~|S|=n$. $\mathscr{F}$ generates an algorithm class $\mathscr{A}=\{\mathbf{A}_{\bm h}: \forall {\bm h}\in \mathscr{F}\}$. We select a consistent algorithm from the algorithm class $\mathscr{A}$. 

We construct a special infinite sequence $\tilde{{\bm h}}=(\tilde{h}_1,...,\tilde{h}_n,...)\in \mathscr{F}$. For each positive integer $n$, we select $\tilde{h}_n$ from
\begin{equation*}
\bigcap_{\forall D_{XY}\in [D_{XY}']}\{ h' \in \mathcal{H}: R_D^{\rm out}(h') \leq \inf_{h\in \mathcal{H}} R_D^{\rm out}(h)+2/n\}\bigcap  \{ h' \in \mathcal{H}: R_D^{\rm in}(h') \leq \inf_{h\in \mathcal{H}} R_D^{\rm in}(h)+2/n\}.
\end{equation*}
The existence of $\tilde{h}_n$ is based on Condition \ref{Con2}. It is easy to check that for any $D_{XY}\in [D_{XY}']$,
 \begin{equation*}
 \begin{split}
  & \mathbb{E}_{S\sim D^n_{X_{\rm I}Y_{\rm I}}}  R_D^{\rm in}(\mathbf{A}_{\tilde{{\bm h}}}(S))\leq  \inf_{h\in \mathcal{H}} R_D^{\rm in}(h)+2/n.
   \\
   & \mathbb{E}_{S\sim D^n_{X_{\rm I}Y_{\rm I}}}  R_D^{\rm out}(\mathbf{A}_{\tilde{{\bm h}}}(S))\leq  \inf_{h\in \mathcal{H}} R_D^{\rm out}(h)+2/n.
    \end{split}
 \end{equation*}
 Since $(1-\alpha)\inf_{h\in \mathcal{H}} R_D^{\rm in}(h)+\alpha \inf_{h\in \mathcal{H}} R_D^{\rm out}(h)\leq \inf_{h\in \mathcal{H}} R_D^{\rm \alpha}(h)$, we obtain that for any $\alpha\in [0,1]$, 
 \begin{equation*}
 \begin{split}
  & \mathbb{E}_{S\sim D^n_{X_{\rm I}Y_{\rm I}}}  R_D^{\alpha}(\mathbf{A}_{\tilde{{\bm h}}}(S))\leq  \inf_{h\in \mathcal{H}} R_D^{\alpha}(h)+2/n.
    \end{split}
 \end{equation*}
Using Theorem \ref{T1} (the second result), we have completed this proof.
\end{proof}
\vspace{0.5cm}
\thmPosbtennew*
\begin{proof}[Proof of Theorem \ref{T-SET}]
$~~~$

\textbf{First}, we prove that if OOD detection is learnable in $\mathscr{D}_{XY}^F$ for $\mathcal{H}$, then Condition \ref{Con2} holds.

Since $\mathscr{D}^{F}_{XY}$ is the prior-unknown space, by Theorem \ref{T1}, there exist an algorithm $\mathbf{A}: \cup_{n=1}^{+\infty}(\mathcal{X}\times\mathcal{Y})^n\rightarrow \mathcal{H}$ and a monotonically decreasing sequence $\epsilon_{\rm cons}(n)$, such that $\epsilon_{\rm cons}(n)\rightarrow 0$, as $n\rightarrow +\infty$, and for any $D_{XY}\in \mathscr{D}_{XY}^F$, 
\begin{equation*}
\begin{split}
     &\mathbb{E}_{S\sim D^n_{X_{\rm I}Y_{\rm I}}}\big[ R^{\rm in}_D(\mathbf{A}(S))- \inf_{h\in \mathcal{H}}R^{\rm in}_D(h)\big]\leq \epsilon_{\rm cons}(n),
     \\ &\mathbb{E}_{S\sim D^n_{X_{\rm I}Y_{\rm I}}}\big[ R^{\rm out}_D(\mathbf{A}(S))- \inf_{h\in \mathcal{H}}R^{\rm out}_D(h)\big]\leq \epsilon_{\rm cons}(n).
     \end{split}
\end{equation*}
Then, for any $\epsilon>0$, we can find $n_{\epsilon}$ such that $\epsilon\geq \epsilon_{\rm cons}(n_{\epsilon})$, therefore, if $n={n_{\epsilon}}$, we have
\begin{equation*}
\begin{split}
     &\mathbb{E}_{S\sim D^{n_{\epsilon}}_{X_{\rm I}Y_{\rm I}}}\big[ R^{\rm in}_D(\mathbf{A}(S))- \inf_{h\in \mathcal{H}}R^{\rm in}_D(h)\big]\leq \epsilon,
     \\ &\mathbb{E}_{S\sim D^{n_{\epsilon}}_{X_{\rm I}Y_{\rm I}}}\big[ R^{\rm out}_D(\mathbf{A}(S))- \inf_{h\in \mathcal{H}}R^{\rm out}_D(h)\big]\leq \epsilon, 
     \end{split}
\end{equation*}
which implies that there exists $S_{\epsilon}\sim D^{n_{\epsilon}}_{X_{\rm I}Y_{\rm I}}$ such that
\begin{equation*}
\begin{split}
     & R^{\rm in}_D(\mathbf{A}(S_{\epsilon}))- \inf_{h\in \mathcal{H}}R^{\rm in}_D(h)\leq \epsilon,
     \\ & R^{\rm out}_D(\mathbf{A}(S_{\epsilon}))- \inf_{h\in \mathcal{H}}R^{\rm out}_D(h)\leq \epsilon. 
     \end{split}
\end{equation*}
Therefore, for any equivalence class $[D_{XY}']$ with respect to $\mathscr{D}_{XY}^F$ and any $\epsilon>0$, there exists a hypothesis function $\mathbf{A}(S_{\epsilon})\in \mathcal{H}$ such that for any domain $D_{XY}\in [D_{XY}']$,
 \begin{equation*}
 \mathbf{A}(S_{\epsilon})\in \{ h' \in \mathcal{H}: R_D^{\rm out}(h') \leq \inf_{h\in \mathcal{H}} R_D^{\rm out}(h)+\epsilon\} \cap   \{ h' \in \mathcal{H}: R_D^{\rm in}(h') \leq \inf_{h\in \mathcal{H}} R_D^{\rm in}(h)+\epsilon\},
 \end{equation*}
 which implies that Condition \ref{Con2} holds.
 
 \textbf{Second}, we prove Condition \ref{Con2} implies the learnability of OOD detection in $\mathscr{D}_{XY}^F$ for $\mathcal{H}.$
 
For convenience, we assume that all equivalence classes are $[D^1_{XY}],...,[D^m_{XY}]$. By Lemma \ref{lemma7-new},  for every  equivalence class $[D_{XY}^i]$, we can find a corresponding algorithm $\mathbf{A}_{D^i}$ such that OOD detection is learnable in $[D_{XY}^i]$ for $\mathcal{H}$. Additionally, we also set the learning rate for $
\mathbf{A}_{D^i}$ is $\epsilon^i(n)$. By Lemma \ref{lemma7-new}, we know that $\epsilon^i(n)$ can attain $O(1/n)$.

Let ${\mathcal{Z}}$ be $\mathcal{X}\times\mathcal{Y}$. Then, we consider a bounded universal kernel $K(\cdot,\cdot)$ defined over $\mathcal{Z}\times \mathcal{Z}$. Consider the \emph{maximum mean discrepancy} (MMD) \cite{DBLP:journals/jmlr/GrettonBRSS12}, which is a metric between distributions: for any distributions $P$ and $Q$ defined over ${\mathcal{Z}}$, we use ${\rm MMD}_K(Q,P)$ to represent the distance.

Let $\mathscr{F}$ be a set consisting of all finite sequences, whose coordinates are labeled data, \textit{i.e.},
 \begin{equation*}
 \mathscr{F}=\{\mathbf{S}=(S_1,...,S_i,...,S_m): \forall i=1,...,m~\textnormal{and}~\forall~\textnormal{labeled~data}~ S_i\}.
 \end{equation*}
 
 Then, we define an algorithm space as follows:
 \begin{equation*}
     \mathscr{A}=\{\mathbf{A}_{\mathbf{S}}\footnote{In this paper, we regard an algorithm as a mapping from $\cup_{n=1}^{+\infty}(\mathcal{X}\times\mathcal{Y})^n$ to $\mathcal{H}$. So we can design an algorithm like this.}:\forall~\mathbf{S}\in  \mathscr{F}\},
 \end{equation*}
 where
 \begin{equation*}
      \mathbf{A}_{\mathbf{S}}(S) = \mathbf{A}_{D^i}(S),~\textnormal{if}~i=\argmin_{i\in\{1,...m\}}  {\rm MMD}_K(P_{S_i},P_S),
 \end{equation*}
  here 
 \begin{equation*}
     P_S = \frac{1}{n}\sum_{(\mathbf{x},y)\in S} \delta_{(\mathbf{x},y)},~~~P_{S_i} = \frac{1}{n}\sum_{(\mathbf{x},y)\in S_i}, \delta_{(\mathbf{x},y)}
 \end{equation*}
 and $\delta_{(\mathbf{x},y)}$ is the Dirac measure.
 Next, we prove that we can find an algorithm $\mathbf{A}$ from the algorithm space $\mathscr{A}$  such that $\mathbf{A}$ is the consistent algorithm.
 
  Since the number of different equivalence classes is finite, we know that there exists a constant $c>0$ such that for any different equivalence classes $[D_{XY}^i]$ and $[D_{XY}^j]$ ($i\neq j$),
 \begin{equation*}
     {\rm MMD}_K(D^i_{X_{\rm I}Y_{\rm I}},D^j_{X_{\rm I}Y_{\rm I}}) >c.
 \end{equation*}
 
  Additionally,  according to \cite{DBLP:journals/jmlr/GrettonBRSS12} and the property of $\mathscr{D}_{XY}^F$ (the number of different equivalence classes is finite), there exists a monotonically decreasing $\epsilon(n)\rightarrow 0$, as $n\rightarrow +\infty$ such that for any $D_{XY}\in \mathscr{D}$,
 \begin{equation}\label{MMDcon}
     \mathbb{E}_{S\sim D^n_{X_{\rm I}Y_{\rm I}}} {\rm MMD}_K(D_{X_{\rm I}Y_{\rm I}},P_S) \leq \epsilon(n),~\text{where}~\epsilon(n)=O(\frac{1}{\sqrt{n^{1-\theta}}}).
 \end{equation}
 Therefore, for every  equivalence class $[D_{XY}^i]$, we can find data points $S_{D^i}$ such that
 \begin{equation*}
      {\rm MMD}_K(D^i_{X_{\rm I}Y_{\rm I}},P_{S_{D^i}}) < \frac{c}{100}.
 \end{equation*}
 \\
 Let $\mathbf{S}'=\{S_{D^1},...,S_{D^i},...,S_{D^m}\}$. Then, we prove that $\mathbf{A}_{\mathbf{S}'}$ is a consistent algorithm.
 By Eq. \eqref{MMDcon}, it is easy to check that for any $i\in\{1,...,m\}$ and any $0<\delta<1$,
 \begin{equation*}
     \mathbb{P}_{S\sim D^{i,n}_{X_{\rm I}Y_{\rm I}}}\big [{\rm MMD}_K(D^i_{X_{\rm I}Y_{\rm I}},P_{S})\leq \frac{\epsilon(n)}{\delta} \big ] >1- \delta,
 \end{equation*}
 which implies that
 \begin{equation*}
     \mathbb{P}_{S\sim D^{i,n}_{X_{\rm I}Y_{\rm I}}}\big [{\rm MMD}_K(P_{S_{D^i}},P_{S})\leq \frac{\epsilon(n)}{\delta}+\frac{c}{100} \big ] >1- \delta.
 \end{equation*}
 Therefore, (here we set $\delta=200\epsilon(n)/c$)
 \begin{equation*}
      \mathbb{P}_{S\sim D^{i,n}_{X_{\rm I}Y_{\rm I}}}\big [ \mathbf{A}_{\mathbf{S}'}(S) \neq \mathbf{A}_{D^i}(S) \big ] \leq \frac{200\epsilon(n)}{c}.
 \end{equation*}
 Because $\mathbf{A}_{D^i}$  is a consistent algorithm for $[D_{XY}^i]$, we conclude that for all $\alpha\in [0,1]$,
 \begin{equation*}
    %\begin{equation*}
\begin{split}
     \mathbb{E}_{S\sim D^{i,n}_{X_{\rm I}Y_{\rm I}}}\big[ R^{\alpha}_D( \mathbf{A}_{\mathbf{S}'}(S) )- \inf_{h\in \mathcal{H}}R^{\alpha}_D(h)\big]\leq \epsilon^i(n)+\frac{200B\epsilon(n)}{c},
     \end{split}
%\end{equation*}
 \end{equation*}
where $\epsilon^i(n)=O(1/n)$ is the learning rate of $
\mathbf{A}_{D^i}$ and $B$ is the upper bound of the loss $\ell$.

 Let $\epsilon^{\rm max}(n)=\max \{\epsilon^1(n),...,\epsilon^m(n)\}+\frac{200B\epsilon(n)}{c}$. 
 
 Then, we obtain that 
  for any $D_{XY}\in \mathscr{D}_{XY}^F$ and all $\alpha\in [0,1]$,
 \begin{equation*}
    %\begin{equation*}
\begin{split}
     \mathbb{E}_{S\sim D^n_{X_{\rm I}Y_{\rm I}}}\big[ R^{\alpha}_D( \mathbf{A}_{\mathbf{S}'}(S) )- \inf_{h\in \mathcal{H}}R^{\alpha}_D(h)\big]\leq \epsilon^{\rm max}(n)=O(\frac{1}{\sqrt{n^{1-\theta}}}).
     \end{split}
%\end{equation*}
 \end{equation*}
According to Theorem \ref{T1} (the second result), $\mathbf{A}_{\mathbf{S}'}$ is the consistent algorithm. This proof is completed.
\end{proof}
%\subsection{Proof of Theorem \ref{T17.1}}
%\begin{restatable}{theorem}{thmPosbMultitwo}
%\begin{theorem}
%\label{T17.1}
%Let $\mathcal{H}$ be $\mathcal{H}^{\rm in} \bullet \mathcal{H}^{\rm b}$. If  $\mathcal{X}$ is a bounded set, $\mathcal{H}^b$ contains all binary classifiers,  Condition \ref{C3} and Assumption \ref{ass2} hold, then OOD detection is learnable in $\mathscr{D}_{XY}^{\tau}$ for $\mathcal{H}$.
%\end{theorem}
%\begin{proof}[Proof of Theorem \ref{T17.1}]
%The proof is similar to the proof of Theorem \ref{T17}. We omit it.
%\end{proof}
\subsection{Proof of Theorem \ref{T-SET2}}
\thmPosbtennewtwo*
\begin{proof}[Proof of Theorem \ref{T-SET2}]
\textbf{First}, we consider the case that the loss $\ell$ is the zero-one loss.

Since $\mu(\mathcal{X})<+\infty$, without loss of generality, we assume that $\mu(\mathcal{X})=1$.  We also assume that $f_{\rm I}$ is $D_{X_{\rm I}}$'s density function and $f_{\rm O}$ is $D_{X_{\rm O}}$'s density function. Let $f$ be the density function for $0.5*D_{X_{\rm I}} + 0.5*D_{X_{\rm O}}$. It is easy to check that $f=0.5*f_{\rm I}+0.5*f_{\rm O}$.
Additionally, due to Realizability Assumption, it is obvious that for any samples $S=\{(\mathbf{x}_1,y_1),...,(\mathbf{x}_n,y_n)\}\sim D^n_{X_{\rm I}Y_{\rm I}}$, i.i.d., we have that there exists $h^*\in \mathcal{H}$ such that
\begin{equation*}
    \frac{1}{n} \sum_{i=1}^n \ell(h^*(\mathbf{x}_i),y_i) = 0.
\end{equation*}

 Given $m$ data points $S_m = \{\mathbf{x}_1',...,\mathbf{x}_m'\}\subset \mathcal{X}^m$. We consider the following learning rule:
\begin{equation*}
    \min_{h\in \mathcal{H}} \frac{1}{m} \sum_{j=1}^m \ell(h(\mathbf{x}_j'),K+1),~
    \textnormal{ subject~to~} \frac{1}{n} \sum_{i=1}^n \ell(h(\mathbf{x}_i),y_i) = 0.
\end{equation*}

We denote the algorithm, which solves the above rule, as $\mathbf{A}_{S_m}\footnote{In this paper, we regard an algorithm as a mapping from $\cup_{n=1}^{+\infty}(\mathcal{X}\times\mathcal{Y})^n$ to $\mathcal{H}$. So we can design an algorithm like this.}$. For different data points $S_m$, we have different algorithm $\mathbf{A}_{S_m}$. Let $\mathcal{S}$ be the infinite sequence set that consists of all infinite sequences, whose coordinates are data points, \textit{i.e.},
\begin{equation}
    \mathcal{S}:= \{\mathbf{S}:=(S_1,S_2,...,S_m,...): S_m ~\textnormal{are~any~}m~\textnormal{ data~points},~m=1,...,+\infty\}.
\end{equation}

\noindent Using $\mathcal{S}$, we construct an algorithm space as follows:
\begin{equation*}
    \mathscr{A}:= \{ \mathbf{A}_{\mathbf{S}}: \forall~\mathbf{S}\in \mathcal{S}\},~\textnormal{where}~ \mathbf{A}_{\mathbf{S}}(S)=\mathbf{A}_{S_n}(S), ~{\rm if}~ |S|=n.
\end{equation*}

\noindent Next, we prove that there exists an algorithm $\mathbf{A}_{\mathbf{S}}\in \mathscr{A}$, which is a
consistent algorithm. \noindent Given data points $S_n\sim \mu^n$, i.i.d., using the  Natarajan
dimension theory and Empirical risk minimization principle \cite{shalev2014understanding}, it is easy to obtain that there exists a uniform constant $C_{\theta}$ such that (we mainly use the uniform bounds to obtain the following bounds)
\begin{equation*}
    \mathbb{E}_{S\sim D^n_{X_{\rm I}Y_{\rm I}}} \sup_{h\in \mathcal{H}_S}R_{D}^{\rm in}(h) \leq \inf_{h\in \mathcal{H}} R_D^{\rm in}(h)+\frac{C_{\theta}}{\sqrt{n^{1-\theta}}},
\end{equation*}
and because of $\mathcal{H}_S \subset \mathcal{H}$,
\begin{equation}\label{5.6q1}
    \mathbb{E}_{S_n \sim \mu^n} \sup_{S\in (\mathcal{X}\times \mathcal{Y})^n}[R_{\mu}(\mathbf{A}_{S_n}(S),K+1) - \inf_{h\in \mathcal{H}_S} R_{\mu}(h,K+1)] \leq \frac{C_{\theta}}{\sqrt{n^{1-\theta}}},
\end{equation}
where 
\begin{equation*}
    \mathcal{H}_S = \{h\in \mathcal{H}: \sum_{i=1}^n \ell(h(\mathbf{x}_i),y_i) = 0\},~\textnormal{here}~S=\{(\mathbf{x}_1,y_1),...,(\mathbf{x}_n,y_n)\},
    \end{equation*}
    and 
\begin{equation*}
   R_{\mu}(h,K+1) = \mathbb{E}_{\mathbf{x}\sim \mu} \ell(h(\mathbf{x}),K+1)=\int_{\mathcal{X}} \ell(h(\mathbf{x}),K+1) {\rm d}\mu(\mathbf{x}).
\end{equation*}

We set $\mathscr{D}_{\rm I} = \{D_{X_{\rm I}Y_{\rm I}}:\text{there exists } D_{X_{\rm O}Y_{\rm O}} \text{such that}~ (1-\alpha)D_{X_{\rm I}Y_{\rm I}}+\alpha D_{X_{\rm O}Y_{\rm O}} \in \mathscr{D}^{\mu,b}_{XY}\}$. Then by Eq. \eqref{5.6q1}, we have
\begin{equation}\label{5.6q11}
     \mathbb{E}_{S_n \sim \mu^n} \sup_{D_{X_{\rm I}Y_{\rm I}}\in \mathscr{D}_{\rm I}} \mathbb{E}_{S\sim D^n_{X_{\rm I}Y_{\rm I}}}[R_{\mu}(\mathbf{A}_{S_n}(S),K+1) - \inf_{h\in \mathcal{H}_S} R_{\mu}(h,K+1)] \leq \frac{C_{\theta}}{\sqrt{n^{1-\theta}}}.
\end{equation}

Due to Realizability Assumption, we obtain that $\inf_{h\in \mathcal{H}} R_D^{\rm in}(h)=0$. Therefore,
\begin{equation}\label{5.6q3}
    \mathbb{E}_{S\sim D^n_{X_{\rm I}Y_{\rm I}}} \sup_{h\in \mathcal{H}_S}R_{D}^{\rm in}(h) \leq \frac{C_{\theta}}{\sqrt{n^{1-\theta}}},
\end{equation}
which implies that (in following inequalities, $g$ is the groundtruth labeling function, \textit{i.e.}, $R_D(g)=0$)
\begin{equation*}
\begin{split}
    \frac{C_{\theta}}{\sqrt{n}} \geq  \mathbb{E}_{S\sim D^n_{X_{\rm I}Y_{\rm I}}} \sup_{h\in \mathcal{H}_S}R_{D}^{\rm in}(h)=&  \mathbb{E}_{S\sim D^n_{X_{\rm I}Y_{\rm I}}} \sup_{h\in \mathcal{H}_S}\int_{g<K+1} \ell(h(\mathbf{x}),g(\mathbf{x}))f_{\rm I}(\mathbf{x}) {\rm d} \mu(\mathbf{x}) \\ \geq & \frac{2}{b}\mathbb{E}_{S\sim D^n_{X_{\rm I}Y_{\rm I}}} \sup_{h\in \mathcal{H}_S}\int_{g<K+1}\ell(h(\mathbf{x}),g(\mathbf{x})){\rm d} \mu(\mathbf{x}).
    \end{split}
\end{equation*}
 This implies that (here we have used the property of zero-one loss)
\begin{equation*}
    \mathbb{E}_{S\sim D^n_{X_{\rm I}Y_{\rm I}}} \inf_{h\in \mathcal{H}_S}\int_{g<K+1}\ell(h(\mathbf{x}),K+1){\rm d} \mu(\mathbf{x}) \geq   \mu({\mathbf{x}\in \mathcal{X}:g(\mathbf{x})<K+1})-\frac{C_{\theta}b}{2\sqrt{n^{1-\theta}}}.
\end{equation*}
Therefore,
\begin{equation}\label{zhen1}
     \mathbb{E}_{S\sim D^n_{X_{\rm I}Y_{\rm I}}} \inf_{h\in \mathcal{H}_S} R_{\mu}(h,K+1) \geq \mu({\mathbf{x}\in \mathcal{X}:g(\mathbf{x})<K+1})-\frac{C_{\theta}b}{2\sqrt{n^{1-\theta}}}.
\end{equation}

Additionally, $R_{\mu}(g,K+1)=\mu({\mathbf{x}\in \mathcal{X}:g(\mathbf{x})<K+1})$ and $g\in \mathcal{H}_S$, which implies that 
\begin{equation}\label{zhen2}
\inf_{h\in \mathcal{H}_S} R_{\mu}(h,K+1)\leq \mu({\mathbf{x}\in \mathcal{X}:g(\mathbf{x})<K+1}).
\end{equation}

Combining inequalities \eqref{zhen1} and \eqref{zhen2}, we obtain that
\begin{equation}\label{5.6q2}
   \big | \mathbb{E}_{S\sim D^n_{X_{\rm I}Y_{\rm I}}} \inf_{h\in \mathcal{H}_S} R_{\mu}(h,K+1) - \mu({\mathbf{x}\in \mathcal{X}:g(\mathbf{x})<K+1}) \big |\leq \frac{C_{\theta}b}{2\sqrt{n^{1-\theta}}}.
\end{equation}

Using inequalities \eqref{5.6q11} and \eqref{5.6q2}, we obtain that
\begin{equation}\label{5.6q5}
 \mathbb{E}_{S_n \sim \mu^n}   \sup_{D_{X_{\rm I}Y_{\rm I}}\in \mathscr{D}_{\rm I} } \big [ \mathbb{E}_{S\sim D^n_{X_{\rm I}Y_{\rm I}}} R_{\mu}(\mathbf{A}_{S_n}(S),K+1) -  \mu({\mathbf{x}\in \mathcal{X}:g(\mathbf{x})<K+1}) \big ] \leq \frac{C_{\theta}(b+1)}{\sqrt{n^{1-\theta}}}.
\end{equation}

Using inequality \eqref{5.6q3}, we have
\begin{equation}\label{5.6q4}
    \mathbb{E}_{S_n \sim \mu^n} \sup_{D_{X_{\rm I}Y_{\rm I}}\in \mathscr{D}_{\rm I} } \mathbb{E}_{S\sim D^n_{X_{\rm I}Y_{\rm I}}} R_{D}^{\rm in}(\mathbf{A}_{S_n}(S)) \leq \frac{C_{\theta}}{\sqrt{n^{1-\theta}}},
\end{equation}
which implies that (here we use the property of zero-one loss)
\begin{equation}\label{5.6q6}
\begin{split}
  \mathbb{E}_{S_n \sim \mu^n}\sup_{D_{X_{\rm I}Y_{\rm I}}\in \mathscr{D}_{\rm I}}\mathbb{E}_{S\sim D^n_{X_{\rm I}Y_{\rm I}}} \big[& -\int_{g<K+1} \ell(\mathbf{A}_{S_n}(S)(\mathbf{x}),K+1){\rm d}\mu(\mathbf{x})\\&+\mu({\mathbf{x}\in \mathcal{X}:g(\mathbf{x})<K+1})\big ] \leq \frac{2bC_{\theta}}{\sqrt{n^{1-\theta}}}.
  \end{split}
\end{equation}

Combining inequalities \eqref{5.6q5} and \eqref{5.6q6}, we have
\begin{equation*}
    \mathbb{E}_{S_n \sim \mu^n}\sup_{D_{X_{\rm I}Y_{\rm I}}\in \mathscr{D}_{\rm I}}\mathbb{E}_{S\sim D^n_{X_{\rm I}Y_{\rm I}}} \int_{g=K+1} \ell(\mathbf{A}_{S_n}(S)(\mathbf{x}),K+1){\rm d}\mu(\mathbf{x}) \leq \frac{2bC_{\theta}}{\sqrt{n^{1-\theta}}}+\frac{C_{\theta}(b+1)}{\sqrt{n^{1-\theta}}}.
\end{equation*}
Therefore, there exist data points $S_n'$ such that
\begin{equation}\label{5.6q7}
\begin{split}
&\sup_{D_{X_{\rm I}Y_{\rm I}}\in \mathscr{D}_{\rm I}} \mathbb{E}_{S\sim D^n_{X_{\rm I}Y_{\rm I}}} R_D^{\rm out}(\mathbf{A}_{S_n'})\\ =& \sup_{D_{X_{\rm I}Y_{\rm I}}\in \mathscr{D}_{\rm I}}\mathbb{E}_{S\sim D^n_{X_{\rm I}Y_{\rm I}}} \int_{g=K+1} \ell(\mathbf{A}_{S_n'}(S)(\mathbf{x}),K+1)f_{\rm O}(\mathbf{x}){\rm d}\mu(\mathbf{x}) \\ \leq  &   2b\sup_{D_{X_{\rm I}Y_{\rm I}}\in \mathscr{D}_{\rm I}}\mathbb{E}_{S\sim D^n_{X_{\rm I}Y_{\rm I}}} \int_{g=K+1} \ell(\mathbf{A}_{S_n'}(S)(\mathbf{x}),K+1){\rm d}\mu(\mathbf{x}) \leq \frac{4b^2C_{\theta}}{\sqrt{n^{1-\theta}}}+\frac{2C_{\theta}(b^2+b)}{\sqrt{n^{1-\theta}}}.
     \end{split}
\end{equation}

Combining inequalities \eqref{5.6q3} and \eqref{5.6q7}, we obtain that for any $n$, there exists data points $S_n'$ such that
\begin{equation*}
    \mathbb{E}_{S\sim D^n_{X_{\rm I}Y_{\rm I}}} R_D^{\alpha}(\mathbf{A}_{S_n'}) \leq \max \big \{\frac{4b^2C_{\theta}}{\sqrt{n^{1-\theta}}}+\frac{2C_{\theta}(b^2+b)}{\sqrt{n^{1-\theta}}},\frac{C_{\theta}}{\sqrt{n^{1-\theta}}} \big \}.
\end{equation*}

We set data point sequences $\mathbf{S}'=(S_1',S_2',...,S_n',...)$. Then, $\mathbf{A}_{\mathbf{S}'}\in \mathscr{A}$ is the universally consistent algorithm, \textit{i.e.}, for any $\alpha\in [0,1]$
\begin{equation*}
     \mathbb{E}_{S\sim D^n_{X_{\rm I}Y_{\rm I}}} R_D^{\alpha}(\mathbf{A}_{\mathbf{S}'}) \leq \max \big \{\frac{4b^2C_{\theta}}{\sqrt{n^{1-\theta}}}+\frac{2C_{\theta}(b^2+b)}{\sqrt{n^{1-\theta}}},\frac{C_{\theta}}{\sqrt{n^{1-\theta}}} \big\}.
\end{equation*}
We have completed this proof when $\ell$ is the zero-one loss.
\\

\textbf{Second}, we prove the case that $\ell$ is not the zero-one loss. We use the notation $\ell_{0-1}$ as the zero-one loss. According the definition of loss introduced in Section \ref{S3}, we know that there exists a constant $M>0$ such that for any $y_1,y_2\in \mathcal{Y}_{\rm all}$,
\begin{equation*}
    \frac{1}{M}\ell_{0-1}(y_1,y_2)\leq \ell(y_1,y_2) \leq M \ell_{0-1}(y_1,y_2).
\end{equation*}
Hence,
\begin{equation*}
     \frac{1}{M} R_D^{\alpha,\ell_{0-1}}(h) \leq R_D^{\alpha,\ell}(h) \leq M R_D^{\alpha,\ell_{0-1}}(h),
\end{equation*}
where
$R_D^{\alpha,\ell_{0-1}}$ is the $\alpha$-risk with zero-one loss, and $R_D^{\alpha,\ell}$ is the $\alpha$-risk for loss $\ell$.
\\
Above inequality tells us that Realizability Assumption holds with zero-one loss if and only if Realizability Assumption holds with the loss $\ell$. Therefore, we use the result proven in first step. We can find a consistent algorithm $\mathbf{A}$ such that for any $\alpha\in [0,1]$,
\begin{equation*}
     \mathbb{E}_{S\sim D^n_{X_{\rm I}Y_{\rm I}}} R_D^{\alpha, \ell_{0-1}}(\mathbf{A}) \leq O(\frac{1}{\sqrt{n^{1-\theta}}}),
\end{equation*}
which implies that for any $\alpha\in [0,1]$,
\begin{equation*}
     \frac{1}{M}\mathbb{E}_{S\sim D^n_{X_{\rm I}Y_{\rm I}}} R_D^{\alpha, \ell}(\mathbf{A}) \leq O(\frac{1}{\sqrt{n^{1-\theta}}}).
\end{equation*}
We have completed this proof.
\end{proof}
%\newpage

\section{Proof of Proposition \ref{Pr1} and Proof of Proposition \ref{P2}}\label{SK}
%\subsection{Recalling the Important Notations for FCNN-based Hypothesis Space and Score-based Hypothesis Space}
To better understand the contents in Appendices \ref{SK}-\ref{SM},
we introduce the important notations for FCNN-based hypothesis space and score-based hypothesis space detaily.
\\

 \textbf{FCNN-based Hypothesis Space.} Given a sequence $\mathbf{q}=(l_1,l_2,...,l_g)$, where $l_i$ and $g$ are positive integers and $g>2$, we use $g$ to represent the depth of neural network and use $l_i$ to represent the width of the $i$-th layer. After the activation function $\sigma$ is selected, we can obtain the architecture of FCNN according to the sequence $\mathbf{q}$. Given any weights  $\mathbf{w}_i \in  \mathbb{R}^{l_{i}\times l_{i-1}}$ and bias $\mathbf{b}_i \in \mathbb{R}^{l_{i}\times 1}$, the output of the $i$-layer can be written as follows: for any $\mathbf{x}\in \mathbb{R}^{l_1}$,
\begin{equation*}
    \mathbf{f}_i(\mathbf{x}) = \sigma(\mathbf{w}_i\mathbf{f}_{i-1}(\mathbf{x})+\mathbf{b}_i), ~ \forall i=2,...,g-1,
\end{equation*}
where $\mathbf{f}_{i-1}(\mathbf{x})$ is the $i$-th layer output and $\mathbf{f}_1(\mathbf{x})=\mathbf{x}$. Then, the output of FCNN is
$
    \mathbf{f}_{\mathbf{w},\mathbf{b}}(\mathbf{x})= \mathbf{w}_{g}\mathbf{f}_{{g-1}}(\mathbf{x})+\mathbf{b}_g,
$
where $\mathbf{w}=\{\mathbf{w}_2,...,\mathbf{w}_g\}$ and $\mathbf{b}=\{\mathbf{b}_2,...,\mathbf{b}_g\}$.

An FCNN-based scoring function space is defined as:
\begin{equation*}
    \mathcal{F}_{\mathbf{q}}^{\sigma}:=\{\mathbf{f}_{\mathbf{w},\mathbf{b}}:\forall \mathbf{w}_i\in \mathbb{R}^{l_{i}\times l_{i-1}},~\forall \mathbf{b}_i\in \mathbb{R}^{l_{i}\times 1},~i=2,...,g\}.
\end{equation*}

Additionally, given two sequences $\mathbf{q}=(l_1,...,l_g)$ and $\mathbf{q}'=(l_1',...,l_{g'}')$, we use the notation
$
    \mathbf{q}\lesssim\mathbf{q}'
$
to represent the following equations and inequalities:
\begin{equation*}
\begin{split}
g \leq g', ~~~l_1&=l_1',~~~ l_g=l_{g'}', \\
l_i&\leq l'_i, ~~~\forall i=1,...,g-1, \\
l_{g-1}&\leq l'_i, ~~~\forall i=g,...,g'-1.
\end{split}
\end{equation*}
%\begin{Proposition}
%\label{prop1}
% If $\mathbf{q}\lesssim\mathbf{q}'$, then $ \mathcal{F}_{\mathbf{q}}^{\sigma}\subset \mathcal{F}_{\mathbf{q}'}^{\sigma}$.
%\end{Proposition}
%Proposition~\ref{prop1} reveals that $\mathbf{q}$ can be used to help describe how rich an FCNN-based scoring function space is.
 %Once $\mathbf{q}\lesssim\mathbf{q}'$, we will know that $\mathcal{F}_{\mathbf{q}'}^{\sigma}$ contains $\mathcal{F}_{\mathbf{q}}^{\sigma}$. This proposition is used in the proofs of Theorems~\ref{T24} and \ref{T27}.

% \subsection{FCNN-based Hypothesis Space}
Given a sequence $\mathbf{q}=(l_1,...l_g)$ satisfying that $l_1=d$ and $l_g=K+1$, the FCNN-based scoring function space $\mathcal{F}_{\mathbf{q}}^{\sigma}$ can induce an FCNN-based hypothesis space. Before defining the FCNN-based hypothesis space, we define the induced hypothesis function. For any $\mathbf{f}_{\mathbf{w},\mathbf{b}}\in \mathcal{F}_{\mathbf{q}}^{\sigma}$, the induced hypothesis function  is:
\begin{equation*}
    h_{\mathbf{w},\mathbf{b}}(\mathbf{x}):=\argmax_{k\in\{1,...,K+1\}} {f}^k_{\mathbf{w},\mathbf{b}}(\mathbf{x}),~\forall \mathbf{x}\in \mathcal{X},
\end{equation*}
where ${f}^k_{\mathbf{w},\mathbf{b}}(\mathbf{x})$ is the $k$-th coordinate of $\mathbf{f}_{\mathbf{w},\mathbf{b}}(\mathbf{x})$. Then, we define the  FCNN-based hypothesis space as follows:
\begin{equation*}
  \mathcal{H}_{\mathbf{q}}^{\sigma}:= \{h_{\mathbf{w},\mathbf{b}}:\forall \mathbf{w}_i\in \mathbb{R}^{l_{i}\times l_{i-1}},~\forall \mathbf{b}_i\in \mathbb{R}^{l_{i}\times 1},~i=2,...,g\}.
\end{equation*}
\\
\textbf{Score-based Hypothesis Space.}
 Many OOD algorithms detect OOD data using a score-based strategy. That is, given a threshold $\lambda$, a scoring function space $\mathcal{F}_l\subset \{\mathbf{f}:\mathcal{X}\rightarrow \mathbb{R}^l\}$ and a scoring function $E: \mathcal{F}_l\rightarrow \mathbb{R}$, then $\mathbf{x}$ is regarded as ID, if $E(\mathbf{f}(\mathbf{x}))\geq \lambda$; otherwise, $\mathbf{x}$ is regarded as OOD. 

Using $E$, $\lambda$ and $\mathbf{f}\in \mathcal{F}_{\mathbf{q}}^{\sigma}$, we can generate a binary classifier $h^{\lambda}_{\mathbf{f},E}$:
\begin{equation*}
    h^{\lambda}_{\mathbf{f},E}(\mathbf{x}) := \left \{
    \begin{aligned}
   &~~~~1,~~{\rm if}~E(\mathbf{f}(\mathbf{x}))\geq \lambda;\\ 
   &~~~~  2,~~{\rm if}~E(\mathbf{f}(\mathbf{x}))< \lambda,
    \end{aligned}
    \right.
\end{equation*}
where $1$ represents ID data, and $2$ represents OOD data. Hence, a binary classification hypothesis space $\mathcal{H}^b$, which consists of all $h^{\lambda}_{\mathbf{f},E}$, is generated. We define the score-based hypothesis space $
    \mathcal{H}^{{\sigma},\lambda}_{{\mathbf{q}},E} := \{h^{\lambda}_{\mathbf{f},E}:\forall \mathbf{f}\in \mathcal{F}_{\mathbf{q}}^{\sigma}\}
$.

Next, we introduce two important propositions.
% \begin{Proposition}
% \end{Proposition}

%We note that, in many works \cite{}, the bias $\mathbf{b}_g=\mathbf{0}$. It is easy to check that when $\mathbf{b}_g=\mathbf{0}$ and $g>2$, Proposition \ref{P1} holds.
% Then, we introduce the fully-convolutional network (FCN)

% \subsection{Score-based Hypothesis Space}

\begin{Proposition}\label{Pr1}
Given a sequence $\mathbf{q}=(l_1,...l_g)$  satisfying that $l_1=d$ and $l_g=K+1$ (note that $d$ is the dimension of input data and $K+1$ is the dimension of output), then the constant functions $h_{1}$, $h_{2}$,...,$h_{K+1}$ belong to $\mathcal{H}_{\mathbf{q}}^{\sigma}$, where $h_{i}(\mathbf{x})=i$, for any $\mathbf{x}\in \mathcal{X}$. Therefore, Assumption \ref{ass1} holds for $\mathcal{H}_{\mathbf{q}}^{\sigma}$.
\end{Proposition}

\begin{proof}[Proof of Proposition \ref{Pr1}]
Note that the output of FCNN can be written as
\begin{equation*}
    \mathbf{f}_{\mathbf{w},\mathbf{b}}(\mathbf{x})= \mathbf{w}_{g}\mathbf{f}_{g-1}(\mathbf{x})+\mathbf{b}_g,
\end{equation*}
where $\mathbf{w}_{g}\in \mathbb{R}^{(K+1)\times l_{g-1}}$, $\mathbf{b}_g\in  \mathbb{R}^{(K+1)\times 1}$ and $\mathbf{f}_{{g-1}}(\mathbf{x})$ is the output of the $l_{g-1}$-th layer. If we set $\mathbf{w}_{g}=\mathbf{0}$, and set $\mathbf{b}_g=\mathbf{y}_i$, where $\mathbf{y}_i$ is the one-hot vector corresponding to label $i$. Then $\mathbf{f}_{\mathbf{w},\mathbf{b}}(\mathbf{x}) = \mathbf{y}_i$, for any $\mathbf{x}\in \mathcal{X}$. Therefore, $h_i(\mathbf{x})\in \mathcal{H}_{\mathbf{q}}^{\sigma}$, for any $i=1,...,K,K+1.$
\end{proof}

Note that in some works \cite{Safran2017Depth-Width}, $\mathbf{b}_g$ is fixed to $\mathbf{0}$. In fact, it is easy to check that when $g>2$ and activation function $\sigma$ is not a constant, Proposition \ref{P1} still holds, even if $\mathbf{b}_g=\mathbf{0}$.

\begin{Proposition}\label{P2}
For any sequence $\mathbf{q}=(l_1,...,l_g)$ satisfying that $l_1=d$ and $l_g=l$ $($note that $d$ is the dimension of input data and $l$ is the dimension of output$)$, if $\{\mathbf{v}\in \mathbb{R}^l:E(\mathbf{v}) \geq \lambda\}\neq \emptyset$ and  $\{\mathbf{v}\in \mathbb{R}^l:E(\mathbf{v}) < \lambda\}\neq \emptyset$, then the functions $h_{1}$ and $h_{2}$ belong to $\mathcal{H}^{{\sigma},\lambda}_{{\mathbf{q}},E}$, where $h_{1}(\mathbf{x})=1$ and $h_{2}(\mathbf{x})=2$, for any $\mathbf{x}\in \mathcal{X}$, where $1$ represents the ID labels, and $2$ represents the OOD labels. Therefore, Assumption \ref{ass1} holds.
\end{Proposition}

\begin{proof}[Proof of Proposition \ref{P2}]
Since $\{\mathbf{v}\in \mathbb{R}^l:E(\mathbf{v})\geq \lambda\}\neq \emptyset $ and $\{\mathbf{v}\in \mathbb{R}^l:E(\mathbf{v})<\lambda\}\neq \emptyset $, we can find $\mathbf{v}_1\in \{\mathbf{v}\in \mathbb{R}^l:E(\mathbf{v})\geq \lambda\}$ and $\mathbf{v}_2\in \{\mathbf{v}\in \mathbb{R}^l:E(\mathbf{v})< \lambda\}$.

\noindent For any  $\mathbf{f}_{\mathbf{w},\mathbf{b}}\in \mathcal{F}^{\sigma}_{\mathbf{q}}$, we have \begin{equation*}
    \mathbf{f}_{\mathbf{w},\mathbf{b}}(\mathbf{x})= \mathbf{w}_{g}\mathbf{f}_{{g-1}}(\mathbf{x})+\mathbf{b}_g,
\end{equation*}
where $\mathbf{w}_{g}\in \mathbb{R}^{l\times l_{g-1}}$, $\mathbf{b}_g\in  \mathbb{R}^{l\times 1}$ and $\mathbf{f}_{{g-1}}(\mathbf{x})$ is the output of the $l_{g-1}$-th layer. 

\noindent If we set $\mathbf{w}_g=\mathbf{0}_{l\times l_{g-1}}$ and $\mathbf{b}_g=\mathbf{v}_1$, then $\mathbf{f}_{\mathbf{w},\mathbf{b}}(\mathbf{x})=\mathbf{v}_1$ for any $\mathbf{x} \in \mathcal{X}$, where $\mathbf{0}_{l\times l_{g-1}}$ is $l\times l_{g-1}$ zero matrix. Hence, $h_1$ can be induced by $\mathbf{f}_{\mathbf{w},\mathbf{b}}$. Therefore, $h_1\in \mathcal{H}_{\mathbf{q},E}^{\sigma,\lambda}$.

\noindent Similarly, if we set $\mathbf{w}_g=\mathbf{0}_{l\times l_{g-1}}$ and $\mathbf{b}_g=\mathbf{v}_2$, then $\mathbf{f}_{\mathbf{w},\mathbf{b}}(\mathbf{x})=\mathbf{v}_2$ for any $\mathbf{x} \in \mathcal{X}$, where $\mathbf{0}_{l\times l_{g-1}}$ is $l\times l_{g-1}$ zero matrix. Hence, $h_2$ can be induced by $\mathbf{f}_{\mathbf{w},\mathbf{b}}$. Therefore, $h_2\in \mathcal{H}_{\mathbf{q},E}^{\sigma,\lambda}$.
\end{proof}
It is easy to check that when $g>2$ and activation function $\sigma$ is not a constant, Proposition \ref{P2} still holds, even if $\mathbf{b}_g=\mathbf{0}$.

\section{Proof of Theorem \ref{T24}}\label{SL}

Before proving Theorem \ref{T24}, we need several lemmas.

\begin{lemma}\label{L22}
Let $\sigma$ be ReLU function: $\max\{x,0\}$. Given $\mathbf{q}=(l_1,...,l_g)$ and $\mathbf{q}'=(l_1',...,l_{g}')$ such that $l_g=l_{g}'$ and $l_1=l_{1}'$, and $l_i\leq l'_i$ $(i=1,...,g-1)$, then $\mathcal{F}_{\mathbf{q}}^{\sigma}\subset \mathcal{F}_{\mathbf{q}'}^{\sigma}$ and $\mathcal{H}_{\mathbf{q}}^{\sigma}\subset \mathcal{H}_{\mathbf{q}'}^{\sigma}$.
\end{lemma}
\begin{proof}[Proof of Lemma \ref{L22}]
Given any weights  $\mathbf{w}_i \in  \mathbb{R}^{l_{i}\times l_{i-1}}$ and bias $\mathbf{b}_i \in \mathbb{R}^{l_{i}\times 1}$, the $i$-layer output of FCNN with architecture $\mathbf{q}$ can be written as
\begin{equation*}
    \mathbf{f}_i(\mathbf{x}) = \sigma(\mathbf{w}_i\mathbf{f}_{i-1}(\mathbf{x})+\mathbf{b}_i),~~\forall \mathbf{x}\in \mathbb{R}^{l_1}, \forall i=2,...,g-1,
\end{equation*}
where $\mathbf{f}_{i-1}(\mathbf{x})$ is the $i$-th layer output and $\mathbf{f}_1(\mathbf{x})=\mathbf{x}$. Then, the output of last layer is
\begin{equation*}
    \mathbf{f}_{\mathbf{w},\mathbf{b}}(\mathbf{x})= \mathbf{w}_{g}\mathbf{f}_{{g-1}}(\mathbf{x})+\mathbf{b}_g.
\end{equation*}
We will show that $\mathbf{f}_{\mathbf{w},\mathbf{b}}
\in \mathcal{F}_{\mathbf{q}'}^{\sigma}$. We construct $\mathbf{f}_{\mathbf{w}',\mathbf{b}'}$ as follows: for every $\mathbf{w}'_i\in \mathbb{R}^{l_{i}'\times l_{i-1}'}$, if $l_i'-l_i>0$ and $l_{i-1}'-l_{i-1}>0$, we set
\begin{equation*}
    \mathbf{w}'_i = \left[\begin{matrix}
 &\mathbf{w}_i&\mathbf{0}_{l_i\times (l_{i-1}'-l_{i-1})}\\
& \mathbf{0}_{(l_i'-l_i)\times l_{i-1}'}& \mathbf{0}_{(l_i'-l_i)\times (l_{i-1}'-l_{i-1})}
  \end{matrix}\right],~~~ \mathbf{b}'_i = \left[\begin{matrix}
 &\mathbf{b}_i\\
& \mathbf{0}_{(l_i'-l_i)\times 1}
  \end{matrix}\right]
\end{equation*}
where $\mathbf{0}_{pq}$ means the $p\times q$ zero matrix.
If $l_i'-l_i=0$ and $l_{i-1}'-l_{i-1}>0$, we set
\begin{equation*}
    \mathbf{w}'_i = \left[\begin{matrix}
 &\mathbf{w}_i&\mathbf{0}_{l_i\times (l_{i-1}'-l_{i-1})}
  \end{matrix}\right],~~~\mathbf{b}'_i = \mathbf{b}_i.
\end{equation*}
If $l_{i-1}'-l_{i-1}=0$ and $l_i'-l_i>0$, we set
\begin{equation*}
    \mathbf{w}'_i = \left[\begin{matrix}
 &\mathbf{w}_i\\
&\mathbf{0}_{(l_i'-l_i)\times l_{i-1}'}
  \end{matrix}\right], ~~~ \mathbf{b}'_i = \left[\begin{matrix}
 &\mathbf{b}_i\\
& \mathbf{0}_{(l_i'-l_i)\times 1}
\end{matrix}\right].
\end{equation*}
If $l_{i-1}'-l_{i-1}=0$ and $l_i'-l_i=0$, we set
\begin{equation*}
    \mathbf{w}'_i = \mathbf{w}_i, ~~~ \mathbf{b}'_i = \mathbf{b}_i.
\end{equation*}
It is easy to check that if $l_i'-l_i>0$
\begin{equation*}
    \mathbf{f}_i'= \left[\begin{matrix}
 &\mathbf{f}_i\\
& \mathbf{0}_{(l_i'-l_i)\times 1}
  \end{matrix}\right].
\end{equation*}

\noindent If $l_i'-l_i=0$,
\begin{equation*}
    \mathbf{f}_i'=\mathbf{f}_i.
\end{equation*}

\noindent Since $l_g'-l_g=0$,
\begin{equation*}
    \mathbf{f}_g'=\mathbf{f}_g,~ \textit{i.e.}, ~  \mathbf{f}_{\mathbf{w}',\mathbf{b}'}=\mathbf{f}_{\mathbf{w},\mathbf{b}}.
\end{equation*}
Therefore, $f_{\mathbf{w},\mathbf{b}}\in \mathcal{F}_{\mathbf{q}'}^{\sigma}$, which implies that $\mathcal{F}_{\mathbf{q}}^{\sigma}\subset \mathcal{F}_{\mathbf{q}'}^{\sigma}$. Therefore, $\mathcal{H}_{\mathbf{q}}^{\sigma}\subset \mathcal{H}_{\mathbf{q}'}^{\sigma}$.
\end{proof}
\vspace{0.5cm}
\begin{lemma}\label{L7-contain}
Let $\sigma$ be the ReLU function: $\sigma(x)=\max\{x,0\}$. Then, $\mathbf{q}\lesssim \mathbf{q}'$ implies that $\mathcal{F}_{\mathbf{q}}^{\sigma}\subset \mathcal{F}_{\mathbf{q}'}^{\sigma}$, $\mathcal{H}_{\mathbf{q}}^{\sigma}\subset \mathcal{H}_{\mathbf{q}'}^{\sigma}$, where $\mathbf{q}=(l_1,...,l_g)$ and $\mathbf{q}'=(l_1',...,l_{g'}')$.
\end{lemma}
\begin{proof}[Proof of Lemma \ref{L7-contain}] 
Given $l''=(l_1'',...,l_{g''}'')$ satisfying that $g\leq g''$, $l_i''=l_i$ for $i=1,...,g-1$, $l_i''=l_{g-1}$ for $i=g,...,g''-1$, and $l_{g''}''=l_{g}$, we first prove that $\mathcal{F}_{\mathbf{q}}^{\sigma}\subset \mathcal{F}_{\mathbf{q}''}^{\sigma}$ and $\mathcal{H}_{\mathbf{q}}^{\sigma}\subset \mathcal{H}_{\mathbf{q}''}^{\sigma}$.

\noindent Given any weights  $\mathbf{w}_i \in  \mathbb{R}^{l_{i}\times l_{i-1}}$ and bias $\mathbf{b}_i \in \mathbb{R}^{l_{i}\times 1}$, the $i$-th layer output of FCNN with architecture $\mathbf{q}$ can be written as
\begin{equation*}
    \mathbf{f}_i(\mathbf{x}) = \sigma(\mathbf{w}_i\mathbf{f}_{i-1}(\mathbf{x})+\mathbf{b}_i),~~\forall \mathbf{x}\in \mathbb{R}^{l_1}, \forall i=2,...,g-1,
\end{equation*}
where $\mathbf{f}_{i-1}(\mathbf{x})$ is the $i$-th layer output and $\mathbf{f}_1(\mathbf{x})=\mathbf{x}$. Then, the output of the last layer is
\begin{equation*}
    \mathbf{f}_{\mathbf{w},\mathbf{b}}(\mathbf{x})= \mathbf{w}_{g}\mathbf{f}_{{g-1}}(\mathbf{x})+\mathbf{b}_g.
\end{equation*}
We will show that $\mathbf{f}_{\mathbf{w},\mathbf{b}}
\in \mathcal{F}_{\mathbf{q}''}^{\sigma}$. We construct $\mathbf{f}_{\mathbf{w}'',\mathbf{b}''}$ as follows: if $i=2,...,g-1$, then $\mathbf{w}''_i=\mathbf{w}$ and $\mathbf{b}_i''=\mathbf{b}_i$; if $i=g,...,g''-1$, then $\mathbf{w}''_i=\mathbf{I}_{l_{g-1}\times l_{g-1}}$ and $\mathbf{b}''_i=\mathbf{0}_{l_{g-1}\times 1}$, where $\mathbf{I}_{l_{g-1}\times l_{g-1}}$ is the $l_{g-1}\times l_{g-1}$ identity matrix, and $\mathbf{0}_{l_{g-1}\times 1}$ is the $l_{g-1}\times 1$ zero matrix; and if $i=g''$, then $\mathbf{w}''_{g''}=\mathbf{w}_{g}$, $\mathbf{b}''_{g''}=\mathbf{b}_{g}$. Then it is easy to check that the output of the $i$-th layer is
\begin{equation*}
    \mathbf{f}''_i=\mathbf{f}_{g-1}, \forall i=g-1,g,...,g''-1.
\end{equation*}
Therefore, $\mathbf{f}_{\mathbf{w}'',\mathbf{b}''}=\mathbf{f}_{\mathbf{w},\mathbf{b}}$, which implies that $\mathcal{F}_{\mathbf{q}}^{\sigma}\subset \mathcal{F}_{\mathbf{q}''}^{\sigma}$. Hence, $\mathcal{H}_{\mathbf{q}}^{\sigma}\subset \mathcal{H}_{\mathbf{q}''}^{\sigma}$.

\noindent When $g'' = g'$, we use Lemma \ref{L22} ($\mathbf{q}''$ and $\mathbf{q}$ satisfy the condition in Lemma \ref{L22}), which implies that $\mathcal{F}_{\mathbf{q}''}^{\sigma}\subset \mathcal{F}_{\mathbf{q}'}^{\sigma}$, $\mathcal{H}_{\mathbf{q}''}^{\sigma}\subset \mathcal{H}_{\mathbf{q}'}^{\sigma}$. Therefore, $\mathcal{F}_{\mathbf{q}}^{\sigma}\subset \mathcal{F}_{\mathbf{q}'}^{\sigma}$, $\mathcal{H}_{\mathbf{q}}^{\sigma}\subset \mathcal{H}_{\mathbf{q}'}^{\sigma}$.
\end{proof}
\vspace{0.5cm}
\begin{lemma}\cite{pinkus1999approximation}\label{L7_UPT}
If the activation function $\sigma$ is not a polynomial, then for any continuous function $f$ defined in $\mathbb{R}^d$, and any compact set $C\subset \mathbb{R}^d$, there exists a fully-connected neural network with architecture $\mathbf{q}$ $(l_1=d,l_g=1)$ such that
\begin{equation*}
 \inf_{f_{\mathbf{w},\mathbf{b}}\in \mathcal{F}_{\mathbf{q}}^{\sigma}} \max_{\mathbf{x}\in C} |f_{\mathbf{w},\mathbf{b}}(\mathbf{x})-f(\mathbf{x})|<\epsilon.
\end{equation*}
\end{lemma}
\begin{proof}[Proof of Lemma \ref{L7_UPT}]
The proof of Lemma \ref{L7_UPT} can be found in Theorem 3.1 in \citep{pinkus1999approximation}.
\end{proof}
\vspace{0.5cm}
\begin{lemma}\label{L8UPT}
If the activation function $\sigma$ is the ReLU function, then for any continuous vector-valued function $\mathbf{f}\in C(\mathbb{R}^d;\mathbb{R}^l)$, and any compact set $C\subset \mathbb{R}^d$, there exists a fully-connected neural network with architecture $\mathbf{q}$ $(l_1=d,l_g=l)$ such that
\begin{equation*}
 \inf_{\mathbf{f}_{\mathbf{w},\mathbf{b}}\in \mathcal{F}_{\mathbf{q}}^{\sigma}} \max_{\mathbf{x}\in C} \|\mathbf{f}_{\mathbf{w},\mathbf{b}}(\mathbf{x})-\mathbf{f}(\mathbf{x})\|_2<\epsilon,
\end{equation*}
where $\|\cdot\|_{2}$ is the $\ell_2$ norm. $($Note that we can also prove the same result, if $\sigma$ is not a polynomial.$)$
\end{lemma}\label{L8_UPT}
\begin{proof}[Proof of Lemma \ref{L8UPT}]
Let $\mathbf{f}=[f_1,...,f_l]^{\top}$, where $f_i$ is the $i$-th coordinate of $\mathbf{f}$. Based on Lemma \ref{L7_UPT}, we obtain $l$ sequences $\mathbf{q}^1$, $\mathbf{q}^2$,...,$\mathbf{q}^l$ such that
\begin{equation*}
\begin{split}
     &\inf_{g_{1}\in \mathcal{F}_{\mathbf{q}^1}^{\sigma}} \max_{\mathbf{x}\in C} |g_{1}(\mathbf{x})-f_1(\mathbf{x})|<\epsilon/\sqrt{l},
   \\ & \inf_{g_{2}\in \mathcal{F}_{\mathbf{q}^2}^{\sigma}} \max_{\mathbf{x}\in C} |g_{2}(\mathbf{x})-f_2(\mathbf{x})|<\epsilon/\sqrt{l},
      \\&~~~~~~~~~~~~~~~~~~~~~~~~~~~~~~~~~...
        \\&~~~~~~~~~~~~~~~~~~~~~~~~~~~~~~~~~...
      \\
      & \inf_{g_{l}\in \mathcal{F}_{\mathbf{q}^l}^{\sigma}} \max_{\mathbf{x}\in C} |g_{l}(\mathbf{x})-f_l(\mathbf{x})|<\epsilon/\sqrt{l}.
     \end{split}
\end{equation*}
 It is easy to find a sequence $\mathbf{q}=(l_1,...,l_g)$ ($l_g=1$) such that $\mathbf{q}^i\lesssim\mathbf{q}$, for all $i=1,...,l$. Using Lemma \ref{L7-contain},  we obtain that $\mathcal{F}_{\mathbf{q}^i}^{\sigma}\subset \mathcal{F}_{\mathbf{q}}^{\sigma}$. Therefore,
 \begin{equation*}
\begin{split}
     &\inf_{g\in \mathcal{F}_{\mathbf{q}}^{\sigma}} \max_{\mathbf{x}\in C} |g(\mathbf{x})-f_1(\mathbf{x})|<\epsilon/\sqrt{l},
   \\ & \inf_{g\in \mathcal{F}_{\mathbf{q}}^{\sigma}} \max_{\mathbf{x}\in C} |g(\mathbf{x})-f_2(\mathbf{x})|<\epsilon/\sqrt{l},
      \\&~~~~~~~~~~~~~~~~~~~~~~~~~~~~~~~~~...
        \\&~~~~~~~~~~~~~~~~~~~~~~~~~~~~~~~~~...
      \\
      & \inf_{g\in \mathcal{F}_{\mathbf{q}}^{\sigma}} \max_{\mathbf{x}\in C} |g(\mathbf{x})-f_l(\mathbf{x})|<\epsilon/\sqrt{l}.
     \end{split}
\end{equation*}
Therefore, for each $i$, we can find $g_{\mathbf{w}^i,\mathbf{b}^i}$ from $\mathcal{F}_{\mathbf{q}}^{\sigma}$ such that
\begin{equation*}
     \max_{\mathbf{x}\in C} |g_{\mathbf{w}^i,\mathbf{b}^i}(\mathbf{x})-f_i(\mathbf{x})|<\epsilon/\sqrt{l},
\end{equation*}
where $\mathbf{w}^i$ represents weights and $\mathbf{b}^i$ represents bias.

\noindent We construct a larger FCNN with $\mathbf{q}'=(l_1',l_2',...,l_g')$ satisfying that $l_1'=d$, $l_i'=l* l_i$, for $i=2,...,g$. We can regard this larger FCNN as a combinations of $l$ FCNNs with architecture $\mathbf{q}$, that is: there are $m$ disjoint sub-FCNNs with architecture $\mathbf{q}$ in the larger FCNN with architecture $\mathbf{q}'$. For $i$-th sub-FCNN, we use weights $\mathbf{w}^i$ and bias $\mathbf{b}^i$. For weights and bias which connect different sub-FCNNs, we set these weights and bias to $\mathbf{0}$. Finally, we can obtain that $\mathbf{g}_{\mathbf{w},\mathbf{b}}=[g_{\mathbf{w}^1,\mathbf{b}^1},g_{\mathbf{w}^2,\mathbf{b}^2},...,g_{\mathbf{w}^l,\mathbf{b}^l}]^{\top}\in \mathcal{F}_{\mathbf{q}'}^{\sigma}$, which implies that
\begin{equation*}
     \max_{\mathbf{x}\in C} \|\mathbf{g}_{\mathbf{w},\mathbf{b}}(\mathbf{x})-\mathbf{f}(\mathbf{x})\|_2<\epsilon.
\end{equation*}
We have completed this proof.
\end{proof}
\noindent Given a sequence $\mathbf{q}=(l_1,...,l_g)$,  we are interested in following function space $\mathcal{F}^{\sigma}_{\mathbf{q},\mathbf{M}}$:
\begin{equation*}
    \mathcal{F}^{\sigma}_{\mathbf{q},\mathbf{M}} := \{ \mathbf{M}\cdot (\sigma \circ \mathbf{f}): \forall~ \mathbf{f}\in \mathcal{F}^{\sigma}_{\mathbf{q}}\},
\end{equation*}
where  $\circ$ means the composition of two functions, $\cdot$ means the product of two matrices, and
\begin{equation*}
\mathbf{M} = \left[\begin{matrix}
 &\mathbf{1}_{1\times (l_g-1)}&0\\
& \mathbf{0}_{1\times (l_g-1)}&1
  \end{matrix}\right],
\end{equation*}
here $\mathbf{1}_{1\times (l_g-1)}$ is the $1\times (l_g-1)$ matrix whose all elements are $1$, and $\mathbf{0}_{1 \times (l_g-1)}$ is the $1\times (l_g-1)$ zero matrix.
Using $ \mathcal{F}^{\sigma}_{\mathbf{q},\mathbf{M}}$, we can construct a binary classification space $\mathcal{H}^{\sigma}_{\mathbf{q},\mathbf{M}}$, which consists of all classifiers satisfying the following condition:
\begin{equation*}
    h(\mathbf{x})=  \argmin_{k=\{1,2\}} f^k_{\mathbf{M}}(\mathbf{x}),
\end{equation*}
where $f^k_{\mathbf{M}}(\mathbf{x})$ is the $k$-th coordinate of $\mathbf{M}\cdot (\sigma \circ \mathbf{f})$.
\vspace{0.5cm}
\begin{lemma}\label{L20}
Suppose that $\sigma$ is the ReLU function: $\max \{x,0\}$. Given a sequence $\mathbf{q}=(l_1,...,l_g)$ satisfying that $l_1=d$ and $l_g=K+1$, then the space $\mathcal{H}^{\sigma}_{\mathbf{q},\mathbf{M}}$ contains $\phi\circ \mathcal{H}^{\sigma}_{\mathbf{q}}$, and $\mathcal{H}^{\sigma}_{\mathbf{q},\mathbf{M}}$ has finite VC dimension $($Vapnik–Chervonenkis dimension$)$, where  $\phi$ maps ID data to $1$ and OOD data to $2$. Furthermore, if given $\mathbf{q}'=(l_1',...,l_g')$ satisfying that $l_g'=K$ and $l_i'=l_i$, for $i=1,...,g-1$, then $\mathcal{H}^{\sigma}_{\mathbf{q}}\subset \mathcal{H}^{\sigma}_{\mathbf{q}'}\bullet \mathcal{H}^{\sigma}_{\mathbf{q},\mathbf{M}}$.
\end{lemma}
\begin{proof}[Proof of Lemma \ref{L20}]
For any $h_{\mathbf{w},\mathbf{b}}\in \mathcal{H}^{\sigma}_{\mathbf{q}}$, then there exists $\mathbf{f}_{\mathbf{w},\mathbf{b}}\in \mathcal{F}^{\sigma}_{\mathbf{q}}$ such that $h_{\mathbf{w},\mathbf{b}}$ is induced by $\mathbf{f}_{\mathbf{w},\mathbf{b}}$. We can write $\mathbf{f}_{\mathbf{w},\mathbf{b}}$ as follows:
\begin{equation*}
    \mathbf{f}_{\mathbf{w},\mathbf{b}}(\mathbf{x})= \mathbf{w}_{g}\mathbf{f}_{{g-1}}(\mathbf{x})+\mathbf{b}_g,
\end{equation*}
where $\mathbf{w}_{g}\in \mathbb{R}^{(K+1)\times l_{g-1}}$, $\mathbf{b}_g\in  \mathbb{R}^{(K+1)\times 1}$ and $\mathbf{f}_{{g-1}}(\mathbf{x})$ is the output of the $l_{g-1}$-th layer.

\noindent Suppose that 
\begin{equation*}
\mathbf{w}_{g} = \left[\begin{matrix}
 &\mathbf{v}_1\\
&  \mathbf{v}_2\\
&  ...\\
&  \mathbf{v}_{K}\\
& \mathbf{v}_{K+1}
  \end{matrix}\right], ~~~
\mathbf{b}_{g} = \left[\begin{matrix}
 &{b}_1\\
&{b}_2\\
&...\\
&{b}_{K}\\
&{b}_{K+1}
  \end{matrix}\right],
 \end{equation*}
where $\mathbf{v}_i\in \mathbb{R}^{1\times l_{g-1}}$ and $b_i \in \mathbb{R}$.

\noindent We set
\begin{equation*}
     \mathbf{f}_{\mathbf{w}',\mathbf{b}'}(\mathbf{x})= \mathbf{w}'_{g}\mathbf{f}_{{g-1}}(\mathbf{x})+\mathbf{b}_g',
\end{equation*}
where 
\begin{equation*}
\mathbf{w}'_{g} = \left[\begin{matrix}
 &\mathbf{v}_1\\
&  \mathbf{v}_2\\
&  ...\\
& ~~~ \mathbf{v}_{K}
  \end{matrix}\right], ~~~
\mathbf{b}_{g}' = \left[\begin{matrix}
 &{b}_1\\
&{b}_2\\
&...\\
&{b}_{K}
  \end{matrix}\right],
 \end{equation*}
It is obvious that $ \mathbf{f}_{\mathbf{w}',\mathbf{b}'} \in \mathcal{F}_{\mathbf{q}'}^{\sigma}$. Using  $\mathbf{f}_{\mathbf{w}',\mathbf{b}'} \in \mathcal{F}_{\mathbf{q}'}^{\sigma}$, we construct a classifier ${h}_{\mathbf{w}',\mathbf{b}'}\in \mathcal{H}^{\sigma}_{\mathbf{q}'}$:
\begin{equation*}
    {h}_{\mathbf{w}',\mathbf{b}'}= \argmax_{k\in \{1,...,K\}}{f}_{\mathbf{w}',\mathbf{b}'}^k,
\end{equation*}
where ${f}_{\mathbf{w}',\mathbf{b}'}^k$ is the $k$-th coordinate of $\mathbf{f}_{\mathbf{w}',\mathbf{b}'}$.

\noindent Additionally,  we consider
\begin{equation*}
    \mathbf{f}_{\mathbf{w},\mathbf{b},\mathbf{B}}= \mathbf{M} \cdot \sigma(\mathbf{B}\cdot \mathbf{f}_{\mathbf{w},\mathbf{b}})\in \mathcal{F}_{\mathbf{q},\mathbf{M}}^{\sigma},
\end{equation*}
where 
\begin{equation*}
\mathbf{B} = \left[\begin{matrix}
 &\mathbf{I}_{(l_g-1)\times (l_g-1) }&- \mathbf{1}_{(l_g-1)\times 1}\\
& \mathbf{0}_{1\times (l_g-1)}& 0
  \end{matrix}\right],
\end{equation*}
here $\mathbf{I}_{(l_g-1)\times (l_g-1) }$ is the $(l_g-1)\times (l_g-1)$ identity matrix, $\mathbf{0}_{1\times (l_g-1)}$ is the $1\times (l_g-1)$ zero matrix, and $\mathbf{1}_{(l_g-1)\times 1}$ is the $(l_g-1)\times 1$ matrix, whose all elements are $1$.

\noindent Then, we define that for any $\mathbf{x}\in \mathcal{X}$,
\begin{equation*}
    h_{\mathbf{w},\mathbf{b},\mathbf{B}}(\mathbf{x}):= \argmax_{k\in\{1,2\}} {f}_{\mathbf{w},\mathbf{b},\mathbf{B}}^k(\mathbf{x}),
\end{equation*}
where ${f}_{\mathbf{w},\mathbf{b},\mathbf{B}}^k(\mathbf{x})$ is the $k$-th coordinate of $ \mathbf{f}_{\mathbf{w},\mathbf{b},\mathbf{B}}(\mathbf{x})$. Furthermore, we can check that $h_{\mathbf{w},\mathbf{b},\mathbf{B}}$ can be written as follows: for any $\mathbf{x}\in \mathcal{X}$,
\begin{equation*}
    h_{\mathbf{w},\mathbf{b},\mathbf{B}}(\mathbf{x})=\left \{ 
    \begin{aligned}
       ~~~1,&~~~&{\rm if}~{f}_{\mathbf{w},\mathbf{b},\mathbf{B}}^1(\mathbf{x})>0;
     \\  ~~~2,&~~~&{\rm if}~{f}_{\mathbf{w},\mathbf{b},\mathbf{B}}^1(\mathbf{x})\leq 0.
    \end{aligned}
    \right.
\end{equation*}

It is easy to check that
\begin{equation*}
  h_{\mathbf{w},\mathbf{b},\mathbf{B}} = \phi \circ h_{\mathbf{w},\mathbf{b}},
\end{equation*}
where $\phi$ maps ID labels to $1$ and OOD labels to $2$.

\noindent Therefore, $h_{\mathbf{w},\mathbf{b}}(\mathbf{x})=K+1$ if and only if $h_{\mathbf{w},\mathbf{b},\mathbf{B}}=2$; and $h_{\mathbf{w},\mathbf{b}}(\mathbf{x})=k$ ($k\neq K+1$) if and only if $h_{\mathbf{w},\mathbf{b},\mathbf{B}}=1$ and $h_{\mathbf{w}',\mathbf{b}'}(\mathbf{x})=k$. This implies that $\mathcal{H}^{\sigma}_{\mathbf{q}}\subset \mathcal{H}^{\sigma}_{\mathbf{q}'}\bullet \mathcal{H}^{\sigma}_{\mathbf{q},\mathbf{M}}$ and $\phi\circ \mathcal{H}^{\sigma}_{\mathbf{q}} \subset \mathcal{H}^{\sigma}_{\mathbf{q},\mathbf{M}}$.

\noindent Let $\tilde{\mathbf{q}}$ be $(l_1,...,l_g,2)$. Then $\mathcal{F}_{\mathbf{q},\mathbf{M}}^{\sigma}\subset \mathcal{F}_{\tilde{\mathbf{q}}}^{\sigma}$. Hence, $\mathcal{H}_{\mathbf{q},\mathbf{M}}^{\sigma}\subset \mathcal{H}_{\tilde{\mathbf{q}}}^{\sigma}$. According to the VC dimension theory \cite{Peter2019Nearly} for feed-forward neural networks, $\mathcal{H}_{\tilde{\mathbf{q}}}^{\sigma}$ has finite VC dimension. Hence, $\mathcal{H}_{\mathbf{q},\mathbf{M}}^{\sigma}$ has finite VC-dimension. We have completed the proof.
\end{proof}
\vspace{0.5cm}
\begin{lemma}\label{L23}
Let $|\mathcal{X}|<+\infty$ and $\sigma$ be the ReLU function: $\max\{x,0\}$. Given $r$ hypothesis functions $h_1,h_2,...,h_r\in \{h:\mathcal{X}\rightarrow \{1,...,l\}\}$, then there exists a sequence $\mathbf{q}=(l_1,...,l_g)$ with $l_1=d$ and $l_g=l$, such that $h_1,...,h_r\in \mathcal{H}_{\mathbf{q}}^{\sigma}$.
\end{lemma}
\begin{proof}[Proof of Lemma \ref{L23}]
For each $h_i$ ($i=1,...,r$), we introduce a corresponding $\mathbf{f}_i$ (defined over $\mathcal{X}$) satisfying that for any $\mathbf{x}\in \mathcal{X}$, $\mathbf{f}_i(\mathbf{x})=\mathbf{y}_k$ if and only if $h_i(\mathbf{x})=k$, where $\mathbf{y}_k\in \mathbb{R}^{l}$ is the one-hot vector corresponding to the label $k$. Clearly, $\mathbf{f}_i$ is a continuous function in $\mathcal{X}$, because $\mathcal{X}$ is a discrete set. Tietze Extension Theorem implies that $\mathbf{f}_i$ can be extended to a  continuous function in $\mathbb{R}^d$.

\noindent Since $\mathcal{X}$ is a compact set, then Lemma \ref{L8UPT} implies that there exist a sequence $\mathbf{q}^i=(l_1^i,...,l_{g^i}^i)$ ($l_1^i=d$ and $l_{g^i}^i=l$) and $\mathbf{f}_{\mathbf{w},\mathbf{b}}\in \mathcal{F}_{\mathbf{q}^i}^{\sigma}$  such that 
\begin{equation*}
 \max_{\mathbf{x}\in \mathcal{X}} \|\mathbf{f}_{\mathbf{w},\mathbf{b}}(\mathbf{x})-\mathbf{f}_i(\mathbf{x})\|_{\ell_2}<\frac{1}{10\cdot l},
\end{equation*}
where $\|\cdot\|_{\ell_2}$ is the $\ell_2$ norm in $\mathbb{R}^{l}$.
Therefore, for any $\mathbf{x}\in \mathcal{X}$, it easy to check that
\begin{equation*}
    \argmax_{k\in\{1,...,l\}}{f}^k_{\mathbf{w},\mathbf{b}}(\mathbf{x})=h_i(\mathbf{x}),
\end{equation*}
where ${f}^k_{\mathbf{w},\mathbf{b}}(\mathbf{x})$ is the $k$-th coordinate of $\mathbf{f}_{\mathbf{w},\mathbf{b}}(\mathbf{x})$. Therefore,
$h_i(\mathbf{x})\in \mathcal{H}_{\mathbf{q}^i}^{\sigma}$.

\noindent Let $\mathbf{q}$ be $(l_1,...,l_g)$ $(l_1=d$ and $l_g=l)$ satisfying that $\mathbf{q}^i\lesssim\mathbf{q}$. Using Lemma \ref{L7-contain}, we obtain that $\mathcal{H}_{\mathbf{q}^i}^{\sigma}\subset \mathcal{H}_{\mathbf{q}}^{\sigma}$, for each $i=1,...,r$. Therefore, $h_1,...,h_r \in \mathcal{H}_{\mathbf{q}}^{\sigma}$.
\end{proof}
\vspace{0.5cm}
\begin{lemma}\label{L26}
Let the activation function $\sigma$ be the ReLU function. Suppose that $|\mathcal{X}|<+\infty$. If $\{\mathbf{v}\in \mathbb{R}^l:E(\mathbf{v})\geq \lambda\}$ and $\{\mathbf{v}\in \mathbb{R}^l:E(\mathbf{v})<\lambda\}$ both contain nonempty open sets of $\mathbb{R}^l$ $($here, open set is a topological terminology$)$. There exists a sequence $\mathbf{q}=(l_1,...,l_g)$ $(l_1=d$ and $l_g=l)$ such that $\mathcal{H}^{\sigma,\lambda}_{\mathbf{q},E}$ consists of all binary classifiers. 
\end{lemma}
\begin{proof}[Proof of Lemma \ref{L26}]
Since $\{\mathbf{v}\in \mathbb{R}^l:E(\mathbf{v})\geq \lambda\}$, $\{\mathbf{v}\in \mathbb{R}^l:E(\mathbf{v})<\lambda\}$ both contain nonempty open sets, we can find $\mathbf{v}_1\in \{\mathbf{v}\in \mathbb{R}^l:E(\mathbf{v})\geq \lambda\}$, $\mathbf{v}_2\in \{\mathbf{v}\in \mathbb{R}^l:E(\mathbf{v})< \lambda\}$ and a constant $r>0$ such that $B_r(\mathbf{v}_1)\subset \{\mathbf{v}\in \mathbb{R}^l:E(\mathbf{v})\geq \lambda\}$ and $B_r(\mathbf{v}_2)\subset \{\mathbf{v}\in \mathbb{R}^l:E(\mathbf{v})< \lambda\}$, where $B_r(\mathbf{v}_1)=\{\mathbf{v}:\|\mathbf{v}-\mathbf{v}_1\|_{\ell_2}<r\}$ and $B_r(\mathbf{v}_2)=\{\mathbf{v}:\|\mathbf{v}-\mathbf{v}_2\|_{\ell_2}<r\}$, here $\|\cdot\|_{\ell_2}$ is the $\ell_2$ norm.

\noindent For any binary classifier $h$ over $\mathcal{X}$, we can induce a vector-valued function as follows: for any $\mathbf{x}\in \mathcal{X}$,
\begin{equation*}
    \mathbf{f}(\mathbf{x}) = \left \{
    \begin{aligned}
   &~~~~\mathbf{v}_1,~~{\rm if}~h(\mathbf{x})=1;\\ 
   &~~~~  \mathbf{v}_2,~~{\rm if}~h(\mathbf{x})=2.
    \end{aligned}
    \right.
\end{equation*}

\noindent Since $\mathcal{X}$ is a finite set, then Tietze Extension Theorem implies that $\mathbf{f}$ can be extended to a  continuous function in $\mathbb{R}^d$. Since $\mathcal{X}$ is a compact set, Lemma \ref{L8UPT} implies that there exists a sequence $\mathbf{q}^h=(l_1^h,...,l_{g^h}^h)$ ($l_1^h=d$ and $l_{g^h}^h=l$) and $\mathbf{f}_{\mathbf{w},\mathbf{b}}\in \mathcal{F}_{\mathbf{q}^h}^{\sigma}$  such that 
\begin{equation*}
 \max_{\mathbf{x}\in \mathcal{X}} \|\mathbf{f}_{\mathbf{w},\mathbf{b}}(\mathbf{x})-\mathbf{f}(\mathbf{x})\|_{\ell_2}<\frac{r}{2},
\end{equation*}
where $\|\cdot\|_{\ell_2}$ is the $\ell_2$ norm in $\mathbb{R}^{l}$.
Therefore, for any $\mathbf{x}\in \mathcal{X}$, it is easy to check that $E(\mathbf{f}_{\mathbf{w},\mathbf{b}}(\mathbf{x}))\geq \lambda$ if and only if $h(\mathbf{x})=1$, and $E(\mathbf{f}_{\mathbf{w},\mathbf{b}}(\mathbf{x}))< \lambda$ if and only if $h(\mathbf{x})=2$.

\noindent For each $h$, we have found a sequence $\mathbf{q}^h$ such that $h$ is induced by $\mathbf{f}_{\mathbf{w},\mathbf{b}}\in \mathcal{F}_{\mathbf{q}^h}^{\sigma}$, $E$ and $\lambda$. Since $|\mathcal{X}|<+\infty$, only finite binary classifiers are defined over $\mathcal{X}$.  Using Lemma \ref{L23}, we can find a sequence $\mathbf{q}$ such that $\mathcal{H}^b_{\rm all}=\mathcal{H}_{\mathbf{q},E}^{\sigma,\lambda}$, where  $\mathcal{H}^b_{\rm all}$ consists of all binary classifiers.
\end{proof}
\vspace{0.5cm}

\begin{lemma}
\label{T27}
Suppose the hypothesis space is score-based.
Let $|\mathcal{X}|<+\infty$. If $\{\mathbf{v}\in \mathbb{R}^l:E(\mathbf{v})\geq \lambda\}$ and $\{\mathbf{v}\in \mathbb{R}^l:E(\mathbf{v})<\lambda\}$ both contain nonempty open sets, and Condition \ref{C3} holds, then there exists a sequence $\mathbf{q}=(l_1,...,l_g)$ $(l_1=d$ and $l_g=l)$ such that for any sequence $\mathbf{q}'$ satisfying $\mathbf{q}\lesssim\mathbf{q}'$ and any ID hypothesis space $\mathcal{H}^{\rm in}$, OOD detection is learnable in the separate space $\mathscr{D}_{XY}^{s}$ for $\mathcal{H}^{\rm in}\bullet \mathcal{H}^{\rm b}$, where $\mathcal{H}^{\rm b}=\mathcal{H}^{{\sigma},\lambda}_{{\mathbf{q}'},E}$ and $\mathcal{H}^{\rm in}\bullet \mathcal{H}^{\rm b}$ is defined below Eq. (\ref{Eq.dot}).
\end{lemma}

\begin{proof}[Proof of Lemma \ref{T27}]
Note that we use the ReLU function as the activation function in this lemma. Using Lemma \ref{L7-contain}, Lemma \ref{L26} and Theorem \ref{T17}, we can prove this result.
\end{proof}
\vspace{0.5cm}
\thmAppFCNN*
\begin{proof}[Proof of Theorem \ref{T24}]
Note that we use the ReLU function as the activation function in this theorem.

\textbf{$\bullet$ The Case that $\mathcal{H}$ is FCNN-based.}

\textbf{First}, we prove that if $|\mathcal{X}|=+\infty$, then OOD detection is not learnable in $\mathscr{D}^s_{XY}$ for $\mathcal{H}_{\mathbf{q}}^{\sigma}$, for any sequence $\mathbf{q}=(l_1,...,l_g)$ $(l_1=d$ and $l_g=K+1)$.

\noindent By Lemma \ref{L20}, Theorems 5 and 8 in \cite{bartlett2003vapnik}, we know that ${\rm VCdim}(\phi\circ \mathcal{H}_{\mathbf{q}}^{\sigma})<+\infty$, where $\phi$ maps ID data to $1$ and maps OOD data to $2$. Additionally, Proposition \ref{Pr1} implies that Assumption \ref{ass1} holds and $\sup_{{h}\in \mathcal{H}_\mathbf{q}^{\sigma}} |\{\mathbf{x}\in \mathcal{X}: {h}(\mathbf{x})\in \mathcal{Y}\}|=+\infty$, when $|\mathcal{X}|=+\infty$. Therefore, Theorem \ref{T12} implies that  OOD detection is not learnable in $\mathscr{D}^s_{XY}$ for $\mathcal{H}_{\mathbf{q}}^{\sigma}$, when $|\mathcal{X}|=+\infty$.

\noindent \textbf{Second}, we prove that if $|\mathcal{X}|<+\infty$, there exists a sequence $\mathbf{q}=(l_1,...,l_g)$ $(l_1=d$ and $l_g=K+1)$ such that OOD detection is learnable in $\mathscr{D}^s_{XY}$ for $\mathcal{H}_{\mathbf{q}}^{\sigma}$.

\noindent Since $|\mathcal{X}|<+\infty$, it is clear that $|\mathcal{H}_{\rm all}|<+\infty$, where $\mathcal{H}_{\rm all}$ consists of all hypothesis functions from $\mathcal{X}$ to $\mathcal{Y}_{\rm all}$. According to Lemma \ref{L23}, there exists a sequence $\mathbf{q}$ such that $\mathcal{H}_{\rm all}\subset \mathcal{H}_{\mathbf{q}}^{\sigma} $. Additionally, Lemma \ref{L20} implies that there exist $\mathcal{H}^{\rm in}$ and $\mathcal{H}^{\rm b}$ such that $\mathcal{H}_{\mathbf{q}}^{\sigma}\subset \mathcal{H}^{\rm in}\bullet \mathcal{H}^{\rm b}$. Since $\mathcal{H}_{\rm all}$ consists all hypothesis space, $\mathcal{H}_{\rm all}=\mathcal{H}_{\mathbf{q}}^{\sigma}= \mathcal{H}^{\rm in}\bullet \mathcal{H}^{\rm b}$. Therefore, $\mathcal{H}^{\rm b}$ contains all binary classifiers from $\mathcal{X}$ to $\{1,2\}$. Theorem \ref{T17} implies that OOD detection is learnable in $\mathscr{D}^s_{XY}$ for $\mathcal{H}_{\mathbf{q}}^{\sigma}$. 

\noindent \textbf{Third}, we prove that if $|\mathcal{X}|<+\infty$, then there exists a sequence $\mathbf{q}=(l_1,...,l_g)$ $(l_1=d$ and $l_g=K+1)$ such that for any sequence $\mathbf{q}'=(l_1',...,l_{g'}')$ satisfying that $\mathbf{q}\lesssim \mathbf{q}'$, OOD detection is learnable in $\mathscr{D}^s_{XY}$ for $\mathcal{H}_{\mathbf{q}'}^{\sigma}$.

\noindent We can use the sequence $\mathbf{q}$ constructed in the second step of the proof. Therefore, $\mathcal{H}_{\mathbf{q}}^{\sigma}=\mathcal{H}_{\rm all}$.  Lemma \ref{L7-contain} implies that  $\mathcal{H}_{\mathbf{q}}^{\sigma}\subset \mathcal{H}_{\mathbf{q}'}^{\sigma}$. Therefore, $\mathcal{H}_{\mathbf{q}'}^{\sigma}=\mathcal{H}_{\rm all}=\mathcal{H}_{\mathbf{q}}^{\sigma}$. The proving process (second step of the proof) has shown that if $|\mathcal{X}|<+\infty$, Condition \ref{C3} holds and hypothesis space $\mathcal{H}$ consists of all hypothesis functions, then OOD detection is learnable in $\mathscr{D}^s_{XY}$ for $\mathcal{H}$. Therefore, OOD detection is learnable in $\mathscr{D}^s_{XY}$ for $\mathcal{H}_{\mathbf{q}'}^{\sigma}$. We complete the proof when the hypothesis space $\mathcal{H}$ is FCNN-based.

\textbf{$\bullet$ The Case that $\mathcal{H}$ is score-based}

\textbf{Fourth}, we prove that if $|\mathcal{X}|=+\infty$, then OOD detection is not learnable in $\mathscr{D}^s_{XY}$ for $\mathcal{H}^{\rm in}\bullet \mathcal{H}^{\rm b}$, where $\mathcal{H}^{\rm b}=\mathcal{H}_{\mathbf{q},E}^{\sigma,\lambda}$ for any sequence $\mathbf{q}=(l_1,...,l_g)$ $(l_1=d$, $l_g=l$), where $E$ is in Eqs. \eqref{score1} or \eqref{score3}.

\noindent By Theorems 5 and 8 in \cite{bartlett2003vapnik}, we know that ${\rm VCdim}( \mathcal{H}_{\mathbf{q},E}^{\sigma,\lambda})<+\infty$. Additionally, Proposition \ref{P2} implies that Assumption \ref{ass1} holds and $\sup_{{h}\in \mathcal{H}_\mathbf{q}^{\sigma}} |\{\mathbf{x}\in \mathcal{X}: {h}(\mathbf{x})\in \mathcal{Y}\}|=+\infty$, when $|\mathcal{X}|=+\infty$. Hence, Theorem \ref{T12} implies that  OOD detection is not learnable in $\mathscr{D}^s_{XY}$ for $\mathcal{H}_{\mathbf{q}}^{\sigma}$, when $|\mathcal{X}|=+\infty$.

\noindent \textbf{Fifth}, we prove that if $|\mathcal{X}|<+\infty$, there exists a sequence $\mathbf{q}=(l_1,...,l_g)$ $(l_1=d$ and $l_g=l)$ such that OOD detection is learnable in $\mathscr{D}^s_{XY}$ for for $\mathcal{H}^{\rm in}\bullet \mathcal{H}^{\rm b}$, where $\mathcal{H}^{\rm b}=\mathcal{H}_{\mathbf{q},E}^{\sigma,\lambda}$ for any sequence $\mathbf{q}=(l_1,...,l_g)$ $(l_1=d$, $l_g=l$), where $E$ is in Eq. \eqref{score1} or Eq. \eqref{score3}.

Based on Lemma \ref{T27}, we only need to show that $\{\mathbf{v}\in \mathbb{R}^l:E(\mathbf{v})\geq \lambda\}$ and $\{\mathbf{v}\in \mathbb{R}^l:E(\mathbf{v})<\lambda\}$ both contain nonempty open sets for different scoring functions $E$.

\noindent Since $\max_{k\in\{1,...,l\}}  \frac{\exp{(v^k)}}{\sum_{c=1}^l \exp{(v^c)}}$,  $\max_{k\in\{1,...,l\}}  \frac{\exp{(v^k/T)}}{\sum_{c=1}^K \exp{(v^c/T)}}$ and $T\log \sum_{c=1}^l  \exp{(v^c/T)}$ are continuous functions, whose ranges contain $(\frac{1}{l},1)$, $(\frac{1}{l},1)$, $(0,+\infty)$ and $(0,+\infty)$, respectively. 

Based on the property of continuous function ($E^{-1}(A)$ is an open set, if $A$ is an open set), we obtain that $\{\mathbf{v}\in \mathbb{R}^l:E(\mathbf{v})\geq \lambda\}$ and $\{\mathbf{v}\in \mathbb{R}^l:E(\mathbf{v})<\lambda\}$ both contain nonempty open sets. Using Lemma \ref{T27}, we complete the fifth step.

\noindent \textbf{Sixth}, we prove that if $|\mathcal{X}|<+\infty$, then there exists a sequence $\mathbf{q}=(l_1,...,l_g)$ $(l_1=d$ and $l_g=l)$ such that for any sequence $\mathbf{q}'=(l_1',...,l_{g'}')$ satisfying that $\mathbf{q}\lesssim \mathbf{q}'$, OOD detection is learnable in $\mathscr{D}^s_{XY}$ for for $\mathcal{H}^{\rm in}\bullet \mathcal{H}^{\rm b}$, where $\mathcal{H}^{\rm b}=\mathcal{H}_{\mathbf{q}',E}^{\sigma,\lambda}$, where $E$ is in Eq. \eqref{score1} or Eq. \eqref{score3}.

\noindent In the fifth step, we have proven that Eq. \eqref{score1} and Eq. \eqref{score3} meet the condition in Lemma \ref{T27}. Therefore, Lemma \ref{T27} implies this result. We complete the proof when the hypothesis space $\mathcal{H}$ is score-based.
\end{proof}

\section{Proofs of Theorem \ref{T24.3} and Theorem \ref{overlapcase}}\label{SM}
\subsection{Proof of Theorem \ref{T24.3}}
\thmAppFCNNtwo*
\begin{proof}[Proof of Theorem \ref{T24.3}]
$~$

1) By Lemma \ref{C1andC2}, we conclude that Learnability in $\mathscr{D}_{XY}^{\mu,b}$ for $\mathcal{H}\Rightarrow$ Condition \ref{C1}.

2) By Proposition \ref{Pr1} and Proposition \ref{P2}, we know that when $K=1$, there exist $h_1, h_2\in \mathcal{H}$, where $h_1=1$ and $h_2=2$, here $1$ represents ID, and $2$ represent OOD. Therefore, we know that when $K=1$,
$\inf_{h\in \mathcal{H}}R_D^{\rm in}(h)=0$ and $\inf_{h\in \mathcal{H}}R_D^{\rm out}(h)=0$, for any $D_{XY}\in \mathscr{D}_{XY}^{\mu,b}$.

By Condition \ref{C1}, we obtain that $\inf_{h\in \mathcal{H}} R_D(h)=0$, for any $D_{XY}\in \mathscr{D}_{XY}^{\mu,b}$. Because each domain $D_{XY}$ in $\mathscr{D}_{XY}^{\mu,b}$ is attainable, we conclude that Realizability Assumption holds. 

We have proven that Condition \ref{C1}$\Rightarrow$ Realizability Assumption.

3) By Theorems 5 and 8 in \cite{bartlett2003vapnik}, we know that ${\rm VCdim}(\phi\circ \mathcal{H}_{\mathbf{q}}^{\sigma})<+\infty$ and ${\rm VCdim}( \mathcal{H}_{\mathbf{q},E}^{\sigma,\lambda})<+\infty$. Then, using Theorem \ref{T-SET2}, we conclude that Realizability Assumption$\Rightarrow$ Learnability in $\mathscr{D}_{XY}^{\mu,b}$ for $\mathcal{H}$.

4) According to the results in 1), 2) and 3), we have proven that

$~~~~~~~~~~~~~~~~~~$Learnability in $\mathscr{D}_{XY}^{\mu,b}$ for $\mathcal{H}\Leftrightarrow$Condition \ref{C1}$\Leftrightarrow$ Realizability Assumption.

5) By Lemma \ref{Lemma1}, we conclude that Condition \ref{Con2}$\Rightarrow$Condition \ref{C1}.

6) \textbf{Here we prove that Learnability in $\mathscr{D}_{XY}^{\mu,b}$ for $\mathcal{H}\Rightarrow$Condition \ref{Con2}.} Since $\mathscr{D}^{\mu,b}_{XY}$ is the prior-unknown space, by Theorem \ref{T1}, there exist an algorithm $\mathbf{A}: \cup_{n=1}^{+\infty}(\mathcal{X}\times\mathcal{Y})^n\rightarrow \mathcal{H}$ and a monotonically decreasing sequence $\epsilon_{\rm cons}(n)$, such that $\epsilon_{\rm cons}(n)\rightarrow 0$, as $n\rightarrow +\infty$, and for any $D_{XY}\in \mathscr{D}_{XY}^{\mu,b}$, 
\begin{equation*}
\begin{split}
     &\mathbb{E}_{S\sim D^n_{X_{\rm I}Y_{\rm I}}}\big[ R^{\rm in}_D(\mathbf{A}(S))- \inf_{h\in \mathcal{H}}R^{\rm in}_D(h)\big]\leq \epsilon_{\rm cons}(n),
     \\ &\mathbb{E}_{S\sim D^n_{X_{\rm I}Y_{\rm I}}}\big[ R^{\rm out}_D(\mathbf{A}(S))- \inf_{h\in \mathcal{H}}R^{\rm out}_D(h)\big]\leq \epsilon_{\rm cons}(n).
     \end{split}
\end{equation*}
Then, for any $\epsilon>0$, we can find $n_{\epsilon}$ such that $\epsilon\geq \epsilon_{\rm cons}(n_{\epsilon})$, therefore, if $n={n_{\epsilon}}$, we have
\begin{equation*}
\begin{split}
     &\mathbb{E}_{S\sim D^{n_{\epsilon}}_{X_{\rm I}Y_{\rm I}}}\big[ R^{\rm in}_D(\mathbf{A}(S))- \inf_{h\in \mathcal{H}}R^{\rm in}_D(h)\big]\leq \epsilon,
     \\ &\mathbb{E}_{S\sim D^{n_{\epsilon}}_{X_{\rm I}Y_{\rm I}}}\big[ R^{\rm out}_D(\mathbf{A}(S))- \inf_{h\in \mathcal{H}}R^{\rm out}_D(h)\big]\leq \epsilon, 
     \end{split}
\end{equation*}
which implies that there exists $S_{\epsilon}\sim D^{n_{\epsilon}}_{X_{\rm I}Y_{\rm I}}$ such that
\begin{equation*}
\begin{split}
     & R^{\rm in}_D(\mathbf{A}(S_{\epsilon}))- \inf_{h\in \mathcal{H}}R^{\rm in}_D(h)\leq \epsilon,
     \\ & R^{\rm out}_D(\mathbf{A}(S_{\epsilon}))- \inf_{h\in \mathcal{H}}R^{\rm out}_D(h)\leq \epsilon. 
     \end{split}
\end{equation*}
Therefore, for any equivalence class $[D_{XY}']$ with respect to $\mathscr{D}_{XY}^{\mu,b}$ and any $\epsilon>0$, there exists a hypothesis function $\mathbf{A}(S_{\epsilon})\in \mathcal{H}$ such that for any domain $D_{XY}\in [D_{XY}']$,
 \begin{equation*}
 \mathbf{A}(S_{\epsilon})\in \{ h' \in \mathcal{H}: R_D^{\rm out}(h') \leq \inf_{h\in \mathcal{H}} R_D^{\rm out}(h)+\epsilon\} \cap   \{ h' \in \mathcal{H}: R_D^{\rm in}(h') \leq \inf_{h\in \mathcal{H}} R_D^{\rm in}(h)+\epsilon\},
 \end{equation*}
 which implies that Condition \ref{Con2} holds. Therefore, Learnability in $\mathscr{D}_{XY}^{\mu,b}$ for $\mathcal{H}\Rightarrow$Condition \ref{Con2}.

7) Note that in 4), 5) and 6), we have proven that

Learnability in $\mathscr{D}_{XY}^{\mu,b}$ for $\mathcal{H}\Rightarrow$Condition \ref{Con2}$\Rightarrow$Condition \ref{C1}, and Learnability in $\mathscr{D}_{XY}^{\mu,b}$ for $\mathcal{H}\Leftrightarrow$Condition \ref{C1}, thus, we conclude that Learnability in $\mathscr{D}_{XY}^{\mu,b}$ for $\mathcal{H}\Leftrightarrow$Condition \ref{Con2}$\Leftrightarrow$Condition \ref{C1}.

8) Combining 4) and 7), we have completed the proof.

%conclude that Learnability in $\mathscr{D}_{XY}^{\mu,b}$ for $\mathcal{H}\Rightarrow$ Condition \ref{C1}.
\end{proof}
%\subsection{Proof of Corollary  \ref{Cor2}}

%\corTwo*

%\begin{proof}[Proof of Theorem \ref{Cor2}]
%According to Theorem \ref{T27}, we only need to show that $\{\mathbf{v}\in \mathbb{R}^K:E(\mathbf{v})\geq \lambda\}$ and $\{\mathbf{v}\in \mathbb{R}^K:E(\mathbf{v})<\lambda\}$ both contain nonempty open sets for different scoring functions $E$. Here, open set is a topological terminology.
%\\

%\noindent Since $\max_{k\in\{1,...,K\}}  \frac{\exp{(v^k)}}{\sum_{c=1}^K \exp{(v^c)}}$,  $\max_{k\in\{1,...,K\}}  \frac{\exp{(v^k/T)}}{\sum_{c=1}^K \exp{(v^c/T)}}$, $T\log \sum_{c=1}^K  \exp{(v^c/T)}$ and $\sum_{c=1}^K \log   (1+\exp{(v^c)})$ are continuous functions, whose ranges contain $(\frac{1}{K},1)$, $(\frac{1}{K},1)$, $(0,+\infty)$ and $(0,+\infty)$, respectively. According the property of continuous function (the inverse image of an open set is an open set, \textit{i.e.}, $E^{-1}(A)$ is an open set, if $A$ is an open set), we obtain that $\{\mathbf{v}\in \mathbb{R}^K:E(\mathbf{v})\geq \lambda\}$ and $\{\mathbf{v}\in \mathbb{R}^K:E(\mathbf{v})<\lambda\}$ both contain nonempty open sets. Using Theorem \ref{T27}, we complete this proof.
%\end{proof}
%\newpage

\subsection{Proof of Theorem \ref{overlapcase}}

\thmImpOverlapFCNN*

\begin{proof}[Proof of Theorem \ref{overlapcase}]
Using Proposition \ref{Pr1} and Proposition \ref{P2}, we obtain that $\inf_{h\in \mathcal{H}} R_D^{\rm in}(h)=0$ and $\inf_{h\in \mathcal{H}} R_D^{\rm out}(h)=0$. Then, Theorem \ref{T5} implies this result.
\end{proof}
%\subsection{Overlap in Score-based Hypothesis Space}\label{SN.2}
%\begin{theorem}\label{T14app}
%Let $K=1$.  Given a priori-unknown space $\mathscr{D}_{XY}$ and any sequence $\mathbf{q}=(l_1,...,l_g)$ $(l_1=d$ and $l_g=K)$, if $\{\mathbf{v}\in \mathbb{R}^K:E(\mathbf{v}) \geq \lambda\}\neq \emptyset$ and  $\{\mathbf{v}\in \mathbb{R}^K:E(\mathbf{v}) < \lambda\}\neq \emptyset$, and there is an OOD domain $D_{XY}\in \mathscr{D}_{XY}$, which has overlap between ID and OOD, then OOD detection is not learnable in $\mathscr{D}_{XY}$ for  $\mathcal{H}_{\mathbf{q},E}^{\sigma,\lambda}$.
%\end{theorem}
%\begin{proof}
%Using Proposition \ref{P2}, we obtain that $\inf_{h\in \mathcal{H}_{\mathbf{q},E}^{\sigma,\lambda}} R_D^{\rm in}(h)=0$ and $\inf_{h\in \mathcal{H}_{\mathbf{q},E}^{\sigma,\lambda}} R_D^{\rm out}(h)=0$. Then, Theorem \ref{T5} implies this result.
%\end{proof}

{Note that if we replace the activation function $\sigma$ (ReLU function) in Theorem  \ref{overlapcase}  with any other activation functions, Theorem \ref{overlapcase} still hold.}

\end{document}